\documentclass[final,5p,times,authoryear]{elsarticle}
\PassOptionsToPackage{colorlinks=true, linkcolor=red, citecolor=red, urlcolor=blue}{hyperref}
\usepackage{amssymb}
\usepackage{booktabs}
\usepackage{amsmath}
\usepackage{multirow}
\usepackage{amsmath} 
\usepackage{color}
\usepackage{appendix}
\usepackage{hyperref}
\usepackage{cleveref}
\usepackage{enumitem}
\RequirePackage{subcaption}
\usepackage{microtype}
\usepackage[table]{xcolor}
\usepackage[ruled,linesnumbered]{algorithm2e}
\newtheorem{proposition}{Proposition}

\journal{}

\begin{document}

\begin{frontmatter}

%% Title, authors and addresses

%% use the tnoteref command within \title for footnotes;
%% use the tnotetext command for theassociated footnote;
%% use the fnref command within \author or \affiliation for footnotes;
%% use the fntext command for theassociated footnote;
%% use the corref command within \author for corresponding author footnotes;
%% use the cortext command for theassociated footnote;
%% use the ead command for the email address,
%% and the form \ead[url] for the home page:
%% \title{Title\tnoteref{label1}}
%% \tnotetext[label1]{}
%% \author{Name\corref{cor1}\fnref{label2}}
%% \ead{email address}
%% \ead[url]{home page}
%% \fntext[label2]{}
%% \cortext[cor1]{}
%% \affiliation{organization={},
%%            addressline={}, 
%%            city={},
%%            postcode={}, 
%%            state={},
%%            country={}}
%% \fntext[label3]{}

\title{Rethinking the impact of noisy labels in graph classification: A
utility and privacy perspective}

%% use optional labels to link authors explicitly to addresses:
%% \author[label1,label2]{}
%% \affiliation[label1]{organization={},
%%             addressline={},
%%             city={},
%%             postcode={},
%%             state={},
%%             country={}}
%%
%% \affiliation[label2]{organization={},
%%             addressline={},
%%             city={},
%%             postcode={},
%%             state={},
%%             country={}}

% \author{}

% \affiliation{organization={},%Department and Organization
%             addressline={}, 
%             city={},
%             postcode={}, 
%             state={},
%             country={}}

\author[1,2,3]{De Li}
\ead{lide@stu.gxnu.edu.cn}

% Second author
\author[1,2,3]{Xianxian Li\corref{mycorrespondingauthor}}
\cortext[mycorrespondingauthor]{Corresponding author}
\ead{lixx@gxnu.edu.cn}

% Third author
\author[3]{Zeming Gan}
\ead{ganzm@stu.gxnu.edu.cn}
\author[3]{Qiyu Li}
\ead{qyL029@stu.gxnu.edu.cn}
\author[1,2,3]{Bin Qu}
\ead{binqu@ieee.org}
\author[1,2,3]{Jinyan Wang\corref{mycorrespondingauthor}}
% \cortext[mycorrespondingauthor]{Corresponding author}
\ead{wangjy612@gxnu.edu.cn}

% \credit{Data curation, Writing - Original draft preparation}
% Address/affiliation
\affiliation[1]{organization={Key Lab of Education Blockchain and Intelligent Technology, Ministry of Education, Guangxi Normal University}, 
    city={Guilin},
    postcode={541004}, 
    country={China}}
% Address/affiliation
\affiliation[2]{organization={Guangxi Key Lab of Multi-source Information Mining $\&$ Security, Guangxi Normal University, Guilin, 541004},
country={China}}
\affiliation[3]{organization={School  of Computer Science and Engineering, Guangxi Normal University, Guilin, 541004},
country={China}}

%% Abstract
\begin{abstract}
Graph neural networks (GNNs) based on message-passing mechanisms have achieved advanced results in graph classification tasks. However, their generalization performance degrades when noisy labels are present in the training data. Most existing noisy labeling approaches focus on the visual domain or graph node classification tasks and analyze the impact of noisy labels only from a utility perspective. Unlike existing work, in this paper, we measure the effects of noise labels on graph classification from data privacy and model utility perspectives. We find that noise labels degrade the model's generalization performance and enhance the ability of membership inference attacks on graph data privacy. To this end, we propose the robust graph neural network (RGLC) approach with noisy labeled graph classification. Specifically, we first accurately filter the noisy samples by high-confidence samples and the first feature principal component vector of each class. Then, the robust principal component vectors and the model output under data augmentation are utilized to achieve noise label correction guided by dual spatial information. Finally, supervised graph contrastive learning is introduced to enhance the embedding quality of the model and protect the privacy of the training graph data. The utility and privacy of the proposed method are validated by comparing twelve different methods on eight real graph classification datasets. Compared with the state-of-the-art methods, the RGLC method achieves at most and at least $7.8 \%$ and $0.8 \%$ performance gain at $30 \%$ noisy labeling rate, respectively, and reduces the accuracy of privacy attacks to below $60 \%$.
\end{abstract}

%% Keywords

\end{frontmatter}

%% Add \usepackage{lineno} before \begin{document} and uncomment 
%% following line to enable line numbers
%% \linenumbers

%% main text
%%

%% Use \section commands to start a section
\section{Introduction}
\label{sec1}
Graphs are a ubiquitous structure found in various data analysis scenarios. They appear in a variety of practical scenarios, such as recommendation systems \citep{zhou2024node}, transportation networks \citep{rahmani2023graph}, social networks \citep{schweimer2022generating}, and biochemical graphs \citep{fontanesi2023exploiting}. Graph neural networks continue the powerful feature extraction capabilities of neural networks and take advantage of the natural structural characteristics of graph data to achieve remarkable performance in tasks such as node and graph classification \citep{luo2024towards,DBLP:conf/ijcai/JuLQWCD0Z22,xie2023label}, link prediction \citep{wu2022mtgcn}, and graph structure learning \citep{zhang2024learning}.

\begin{figure*}
    \centering
    \begin{subfigure}{0.24\textwidth} % 调整子图的宽度
        \centering
        \includegraphics[width=\textwidth]{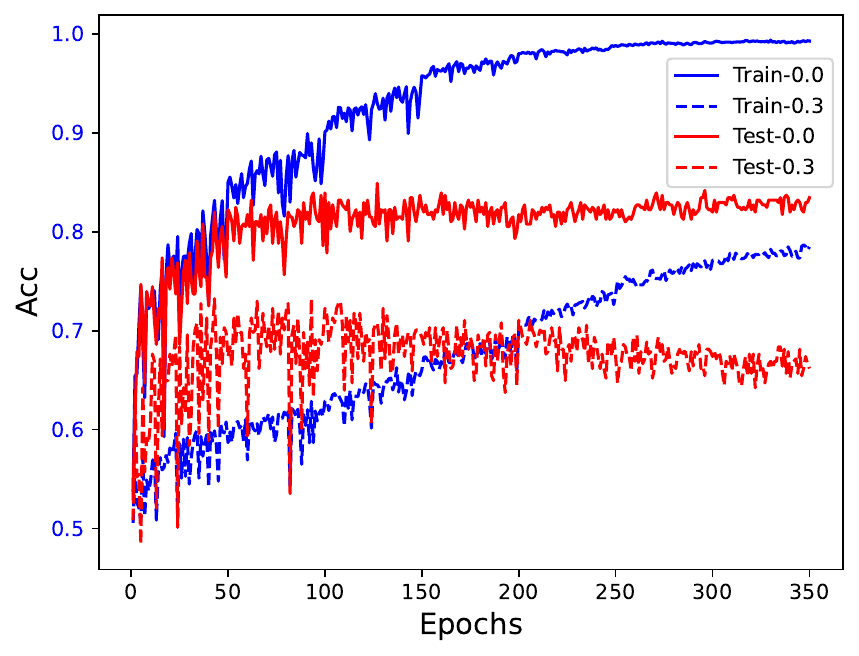}
        \caption{NCI1}
        \label{fig:sub1}
    \end{subfigure}
    \hspace{0.00\textwidth} % 添加水平间距
    \begin{subfigure}{0.24\textwidth}
        \centering
        \includegraphics[width=\textwidth]{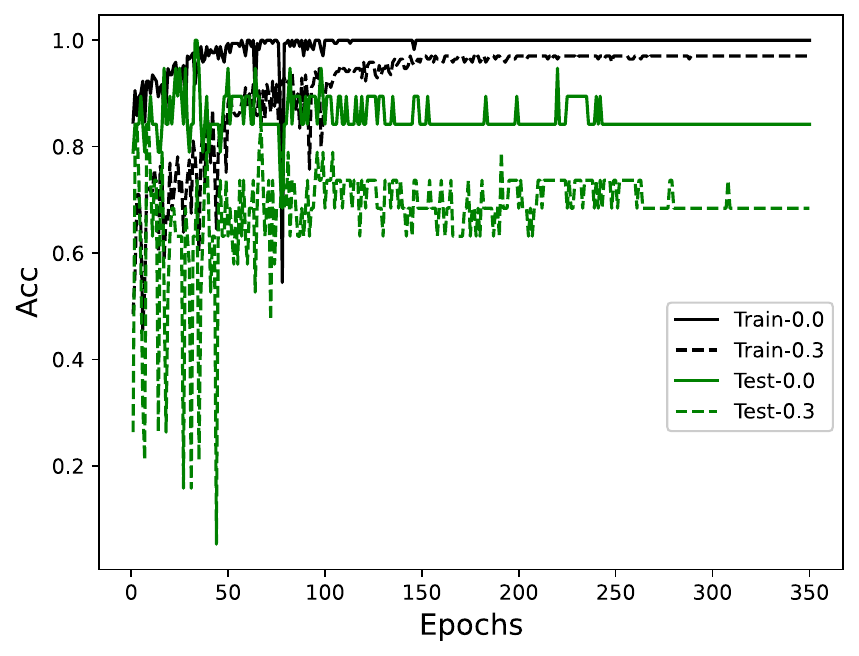}
        \caption{MUTAG}
        \label{fig:sub2}
    \end{subfigure}
    \hspace{0.00\textwidth} % 添加水平间距
    \begin{subfigure}{0.24\textwidth}
        \centering
        \includegraphics[width=\textwidth]{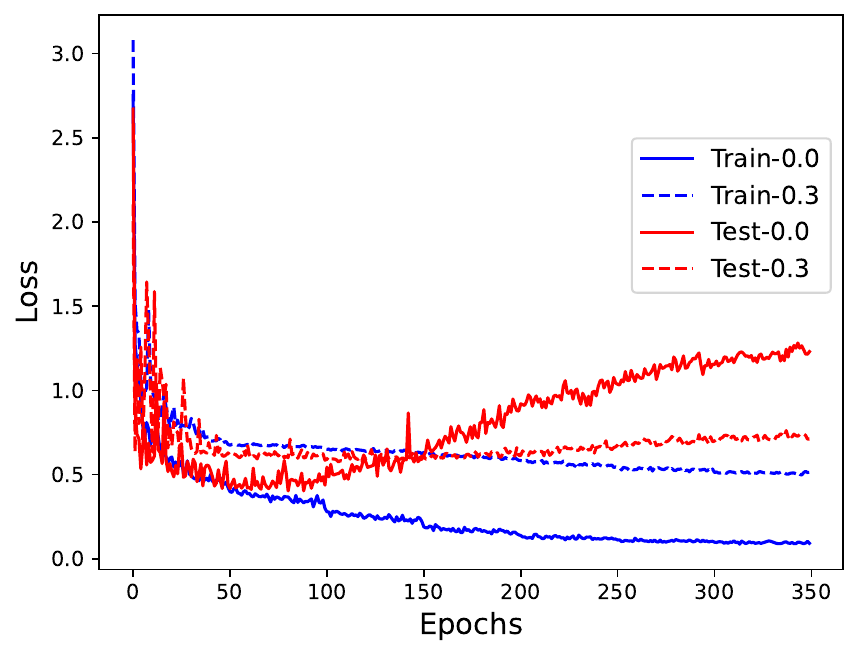}
        \caption{NCI1}
        \label{fig:sub3}
    \end{subfigure}
    \hspace{0.00\textwidth} % 添加水平间距
    \begin{subfigure}{0.235\textwidth}
        \centering
        \includegraphics[width=\textwidth]{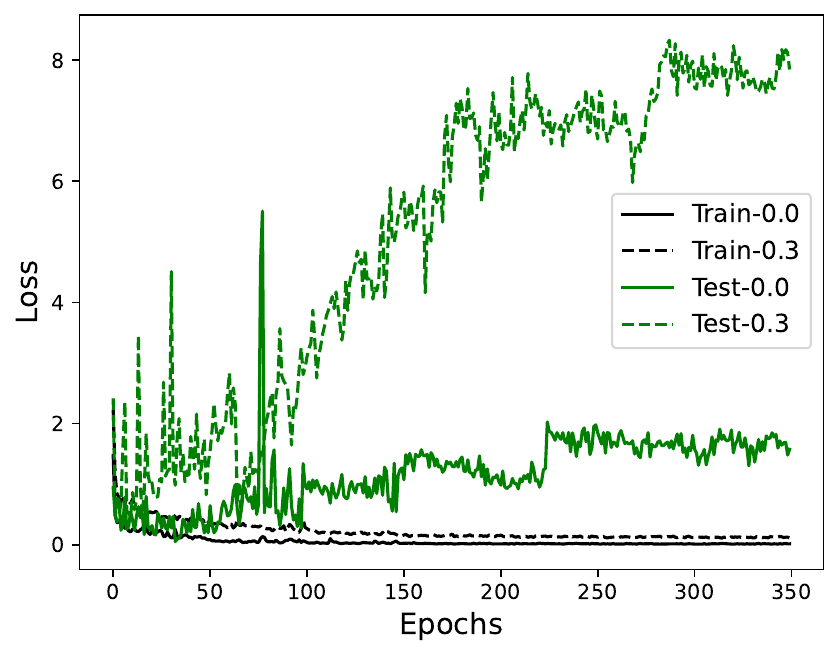}
        \caption{MUTAG}
        \label{fig:sub4}
    \end{subfigure}
    \caption{The accuracy and loss performance of GIN on the MUTAG and NCI1 datasets with different levels of noise damage in the dataset. (a): Accuracy results of GIN at different training and testing cycles on the NCI1 dataset when the clean label and noise label rate is 0.3; (b): GIN's training and testing accuracy on the MUTAG dataset at different noise label rates; (c) and (d): Training loss and test loss of GIN under clean labels and noisy labels on MUTAG and NCI1 datasets, respectively.}
    \label{fig:main_figure}
\end{figure*}

In recent years, graph classification based on GNNs has become one of the important issues in graph machine learning research. It iteratively updates the representation of nodes by aggregating domain information and further obtains the representation of the entire graph through readout operations to perform downstream classification tasks \citep{DBLP:conf/iclr/XuHLJ19}. However, when the training dataset of a graph neural network is damaged (e.g., noisy labels), its generalization performance will gradually deteriorate as the noise rate increases, leading to its further application in some key areas. Fig. 1 shows the accuracy and loss performance of the GNNs model under different noise label rates in the graph classification datasets MUTAG and NCI1. It can be seen from Fig. 1 (a) and (b) that as the proportion of noise in the training set becomes larger, the prediction accuracy of GNNs gradually decreases. In addition, in this study, in addition to the intuitive understanding that noisy labels will reduce the generalization performance of the model, we also found that noisy labels will enhance the success rate of membership inference attacks on graph classification task data, leading to data privacy leakage. The graph neural network community is called upon to increase research on the robustness of graph neural network graph classification tasks from the perspective of member privacy of the training dataset. Multiple studies in the past have shown that the overfitting phenomenon of the model will further increase the risk of leaking member data privacy \citep{salem2019ml,DBLP:conf/uss/LiuWH000CF022,liu2022membership,hu2023defending}. As can be seen in Fig. 1(b) and (c), on the smaller graph dataset MUTAG, the difference between the training loss and the testing loss of the GIN model under noisy label shows a larger difference and a larger overfitting phenomenon compared to the clean label case, which makes the risk of a membership inference attack higher. However, for the larger graph dataset NCI1, when the noise rate is 0.3, the loss gap between training and testing loss is smaller than that in the case of clean label, making it impossible to pass conventional measurements (e.g., cross-entropy or prediction confidence) \citep{song2019privacy} to determine the risk of member privacy leakage. To this end, with the help of \cite{zhang2022inference} proposed method to measure subgraph membership inference attacks based on graph embedding space information in white-box attack scenarios, we find that the noise label leads to a further increase in semantic variability in the embedding space between the member data (e.g., training dataset) and the non-member data (e.g., test dataset), i.e., a further increase in the capability of the subgraph membership inference attacks, which is specifically analyzed and validated in this paper in Subsection \ref{sub1}. Therefore, studying the classification task of graph data with noisy labels can not only improve the generalization of the model in noisy label scenarios but also reduce the leakage of data privacy information.

The OMG method \citep{yin2023omg} is a direct study of the noisy label problem in the field of graph classification. This method uses supervised graph contrastive learning and label correction by nearest neighbor embedding similarity to address the noisy label problem in graph classification. However, due to the instability of nearest neighbor label correction, incorrect label information will guide the mining of positive and negative samples, resulting in suboptimal performance. In addition, there are currently some studies on noise label methods for graph neural network node classification tasks \citep{yuan2023learning,yuan2023alex,xia2024gnn,dai2021nrgnn,qian2023robust,lu2024noise,chen2023erase,LI2024106113}. However, due to the differences between graph and node classification tasks, most of the above methods are unsuitable for graph classification tasks. Most of the existing research on noise label problems mainly focuses on the visual field, which is primarily divided into sample selection \citep{han2018co,ye2022robust,karim2022unicon}, label correction \citep{DBLP:conf/iclr/LiSH20,chen2023co,li2023disc} and regularization methods \citep{DBLP:conf/nips/EnglessonA21,yi2023class,zhou2021learning}. However, due to the complex structural relationships and sparse training data volume of graph data, directly applying noisy label learning methods from the vision domain to graph classification can make the training unstable and lead to sub-optimal generalization performance. To address the impact of noise labels on graph classification, we propose a Robust Graph Neural Network Approach with Noise Labeled Graph Classification (RGLC) in this paper, which achieves effective learning of noise labels through a two-stage noise sample filtering mechanism, a two-view information-guided noise label correction mechanism, and supervised graph contrastive learning. Specifically, inspired by out-of-distribution
detection \citep{DBLP:conf/iclr/LiangLS18} and low-rank graph learning \citep{wang2024low}, we explore the separability of clean and noisy graph data in the graph embedding space, construct Gram matrices for graph embeddings based on the category information of the graph data, and then obtain the first principal component vector (corresponding to the largest eigenvalue) of the Gram matrix for each category using matrix decomposition techniques. Using the property that the ``proximity'' of the clean samples to the principal components is greater than that of the noise samples \footnote{\cite{wang2024low} found that the low-frequency characteristics of label learning with noise, that is, the feature is projected in the embedded space, and the low-rank part (the larger feature principal component vector) covers more data information of the clean label, and only less noisy information is learned.}, we calculate the square of the inner product of each graph data and the principal components of its category to obtain the variance contribution of each sample to the largest principal component of the eigenvalues and then use Gaussian Mixture Models (GMMs) to fit the computed contribution values to realize the screening of the noisy graph data.However, matrix decomposition requires a large amount of computation, especially for graph datasets with a large amount of data. At the same time, high noise label rate will cause interference to the quality of principal component vector. Therefore, we utilize the small loss criterion \footnote{In deep learning with noise labels, samples with small loss values are usually considered clean samples.} to set a higher probability threshold to obtain high-confidence samples before performing graph embedding space separation of noisy samples, i.e., it reduces the computational complexity of the subsequent matrix decomposition and obtains more robust principal component vectors. The effects of the first principal component vector on different quantities and different amounts of noise are given in Appendix \ref{app2}. In addition, to further utilize the supervisory information of noisy data, we propose a robust noise label correction mechanism under dual view, which is different from existing self-correction techniques that use models to predict confidence. On the one hand, class information is constructed in the embedded space by calculating the proximity between the graph embedding of noise data and the first feature principal component vector corresponding to each class. On the other hand, class information is constructed in the output space by the prediction probability of the original graph data and the enhanced graph data. The classification information of the embedded space and the output space are further integrated to realize the label correction of the noisy data. Finally, we use the corrected label information to construct supervised graph contrastive learning to improve the model's characterization ability and protect graph data members' privacy information. The main contributions of this paper are summarized as follows.

 \begin{itemize}
  \item We found that noisy labels reduce the generalization performance of graph neural networks in graph classification tasks and further amplify the privacy leakage of graph or subgraph members. Therefore, we re-examine the impact of noisy labels on graph classification tasks from the perspective of model performance and data privacy.
  \item We propose a two-stage noise sample selection mechanism for accurate noise separation. To further utilize the information of noisy data, we realize the label correction of noisy graph data by constructing class information in the embedded space and the output space.
  \item Relying on precise filtering of noisy graph data and label correction, we introduce supervised graph contrastive learning based on category information to enhance the model's representational capacity. This provides stable predictions for downstream classification tasks and reduces the privacy risks to graph data posed by membership inference attacks.
  \item Many experimental results on real datasets show that the proposed RGLC method can not only improve the impact of noise labels on graph classification performance but also reduce the risk of privacy leakage of members of the training data.
\end{itemize}

The remainder of this paper is organized as follows. Section \ref{se2} summarizes related work, including graph classification, noisy label learning, and membership inference attacks, and gives the differences and connections between our method and existing work. Section \ref{se3} describes the notation definitions, research objectives, and related model structure. Section \ref{se4} first discusses the relationship between noisy labels and privacy leakage, then details our RGLC method. In Section \ref{se5}, we conduct extensive experiments to validate and analyze the efficacy and privacy-preserving capabilities of our RGLC method. Section \ref{se6} concludes the paper and suggests future improvements.

\section{Related Work}
\label{se2}
\subsection{Graph Classification}

Graph classification aims to predict labels for the entire graph, such as solubility or toxicity of molecules. Graph kernels \citep{DBLP:journals/jmlr/ShervashidzeSLMB11,DBLP:conf/nips/TogninalliGLRB19} and message passing-based graph neural networks \citep{DBLP:conf/iclr/XuHLJ19,DBLP:conf/icde/DiYZC21}
are two major methods for achieving graph classification. The former decouples the graph into multiple substructures and uses a series of kernel functions to calculate the similarity of the graph to complete label prediction. The latter mainly propagates and aggregates messages on the graph, where each node receives information from all neighbors. Iteratively performs aggregation and updates node information, and finally all node representations are aggregated into a graph-level representation through readout functions. In recent years, the graph neural network method based on message passing has become a mainstream method for graph classification due to its excellent performance and scalability. In order to reduce the need for a large number of training labels, the TGNN \citep{DBLP:conf/ijcai/JuLQWCD0Z22} and DualGraph \citep{DBLP:conf/icde/LuoJQCDHZ22} methods respectively propose two graph classification methods from the perspective of semi-supervised learning, while GraphCL \citep{you2020graph} and SimGRACE \citep{xia2022simgrace} propose two graph representation learning methods from the perspective of unsupervised learning. However, most existing graph classification methods based on graph neural networks do not consider the impact of noise labels on the model's generalization performance. Although unsupervised graph contrastive methods are unaffected by labels when learning graph embeddings, noisy labels in data for downstream tasks can still reduce the model's generalizability.

\subsection{Learning with noisy labels}
\textbf{Learning with noisy labels in deep learning (LNL).} 
The methods for LNL primarily divide into sample selection \citep{han2018co,ye2022robust,karim2022unicon}, label correction \citep{DBLP:conf/iclr/LiSH20,chen2023co,li2023disc} and regularization methods \citep{DBLP:conf/nips/EnglessonA21,yi2023class,zhou2021learning}. Most sample selection methods leverage the small loss criterion based on the deep model's memorization effect to select clean samples, among which co-teaching \citep{han2018co}, represented by constructing a dual-branch network to select clean samples for each branch, achieves robust learning with noisy labels. Label correction methods aim to further utilize the effective information from the selected noisy samples by attempting to learn or generate pseudo labels to replace the original noisy sample labels, where most existing methods adopt semi-supervised learning techniques for pseudo labeling noisy samples \citep{berthelot2019mixmatch}. Regularization-based methods focus on designing robust loss functions or data augmentation regularization techniques to enhance the model's robustness to noisy labels using all available data. Moreover, unsupervised contrastive learning, inherently independent of noisy labels, can extract undamaged data representations, making learning under contrastive learning with noisy labels a new research trend in recent years \citep{zhang2023rankmatch,ortego2021multi,li2022selective}.

\textbf{Learning noisy labels in graph neural networks.} 
Learning with noisy labels in graph neural networks aims to enhance data performance with inaccurately labeled graphs. NRGNN \citep{dai2021nrgnn} improves node information propagation by connecting labeled and unlabeled nodes with high correlation and utilizes the model's predictive capabilities to mine confident pseudo-labels. Similarly, RTGNN \citep{qian2023robust} enhances node information propagation and employs a dual-branch network to detect and correct noisy samples. On the other hand, GNN Cleaner \citep{xia2024gnn} and ERASE \citep{chen2023erase} leverage the structural characteristics of graphs and use label propagation algorithms to cleanse noisy nodes. Three other methods, CGNN, CR-GNN, and MCLC, are informed by graph contrastive learning to refine noisy label learning. CGNN \citep{yuan2023learning} utilizes unsupervised graph contrastive learning for feature representation and neighbor label correction. CR-GNN \citep{LI2024106113} leverages a dual-channel feature from unsupervised graph contrast to identify prediction-consistent nodes as confident ones for learning amidst noise. MCLC \citep{lu2024noise} integrates unsupervised and semi-supervised contrastive learning \citep{mo2022simple} with self-learning for robust training. However, all of the above methods are designed for the graph neural network node classification task, and cannot be directly migrated to the graph classification task due to the difference between the graph classification and node classification tasks and also lacks the correction of the noise label, which loses valuable supervisory information. \cite{yin2023omg} developed a learning method for graph classification tasks with noisy labels, employing a combination of label-guided supervised graph contrastive learning and neighbor-aware label correction techniques. Although this method led to improvements, its effectiveness was compromised due to inaccuracies in label information and the dependency of neighbor-aware label correction on the quality of graph embeddings from supervised contrastive learning. Compromised embeddings can negatively impact the label correction process, resulting in suboptimal outcomes.

\subsection{Membership inference attack}
Membership inference attacks in machine learning (ML) models occur when an attacker attempts to ascertain if a specific data sample was part of the training dataset for a given ML model \citep{shokri2017membership,DBLP:conf/uss/LiuWH000CF022}. More specifically, for a given one candidate data sample, a trained machine learning model $\mathcal{F}$ and the adversary's external knowledge $\Omega$, the membership inference attack a can be defined as the following function:

\begin{equation}
\mathcal{A}: \mathcal{X}, \mathcal{F}, \Omega \rightarrow\{0,1\} .
\label{eq1}
\end{equation}
where 0 indicates that $\mathcal{X}$ is not a member of the $\mathcal{F}$ training set, and 1 indicates that it is a member. $\mathcal{A}$ symbolizes the attack mechanism. The unintended revelation of data sample membership can lead to significant privacy breaches. For instance, if $\mathcal{X}$ denotes an individual's health records or private data, such attacks could reveal whether this information contributed to training a disease-specific model, thus posing a privacy risk. Membership inference attacks are critical for assessing privacy exposure in statistical data analysis techniques. These attacks are categorized into black-box or white-box types depending on the attacker's capabilities.  Black-box attacks limit the attacker to model outputs through strategies like shadow models or metric-based methods (e.g., cross-entropy, confidence scores). In contrast,  white-box attacks provide access to more in-depth information, including the model's optimal parameters and intermediate embeddings. In this paper, to measure the impact of noise labels on data privacy of graph classification, we set two scenarios: graph-level member inference attack and subgraph-level member inference attack. The graph-level member inference attack is consistent with the method of member inference attack in conventional deep learning \citep{DBLP:conf/uss/LiuWH000CF022}, that is, to judge whether the queried graph data is in the training graph data set. Here, we adopt the black-box scenario proposed by \cite{song2019privacy} to measure the degree of privacy disclosure based on cross-entropy and output accuracy. As for the inference attack scenario of subgraph members, we measure the disclosure of graph data privacy according to \citep{zhang2022inference} in the white box scenario. We use graph embedding information to query whether the subgraph is in the original training graph dataset.

\subsection{Differences and connections with existing methods}
\textit{Differences:} Different from the existing methods of learning with noise labels, which only consider the impact of noise labels on the utility of the model, in this paper, we thoroughly analyze the effects of noise labels on graph classification from the two perspectives of data privacy and utility, providing insights from the perspective of data privacy for the learning of noise labels. In addition, most of the existing methods for learning labels with noise are for data in Euclidean space. Due to the complex structure relationship and the uneven quantity of graph data, the existing deep learning can not achieve good generalization performance when dealing with graph data in non-Euclidean space. However, most of the current noise labeling methods for graph data focus on the semi-supervised node classification task, which is different from the node classification task, so the existing noise methods for node classification cannot be directly transferred to graph classification. In our experiment, RTGNN \citep{qian2023robust} was applied to graph classification with noise labels, and only a tiny performance gain was obtained. Currently, the OMG method \citep{yin2023omg} is the only research on classifying noisy label graphs. The modified label with nearest neighbor information proposed by this method and the graph comparison learning supervised by label information will suffer from too many incorrect labels when the noise label rate is large, which leads to the failure of the nearest neighbor correction method. Unlike the OMG method, we use high-confidence sample selection with small loss and class-first principal component vector robustness to noise to select noise samples accurately. Then, to effectively use the supervisory information of the noise samples, we propose a stable perspective based on the embedded space and the model output space to correct the noise label. Finally, supervised graph Contrastive learning is used to improve the embedding robustness of the model in a noisy environment.

\textit{Connections:} Among the existing deep learning methods with noisy labels, compared with robust loss function, sample selection and label correction techniques can provide more helpful information for the model to improve robustness in noisy label scenarios. Therefore, we adopt two core techniques in this paper: sample selection and label correction. In addition, this paper only discusses the privacy leakage of graph classification tasks under noise label scenarios and does not need to propose a new privacy attack method. Therefore, the existing member inference attack methods \citep{song2019privacy,zhang2022inference} are used to measure the data privacy leakage of graph classification caused by noise labels.

\begin{figure*}[!h]
    \centering
    \begin{subfigure}{0.25\textwidth} % 调整子图的宽度
        \centering
        \includegraphics[width=\textwidth]{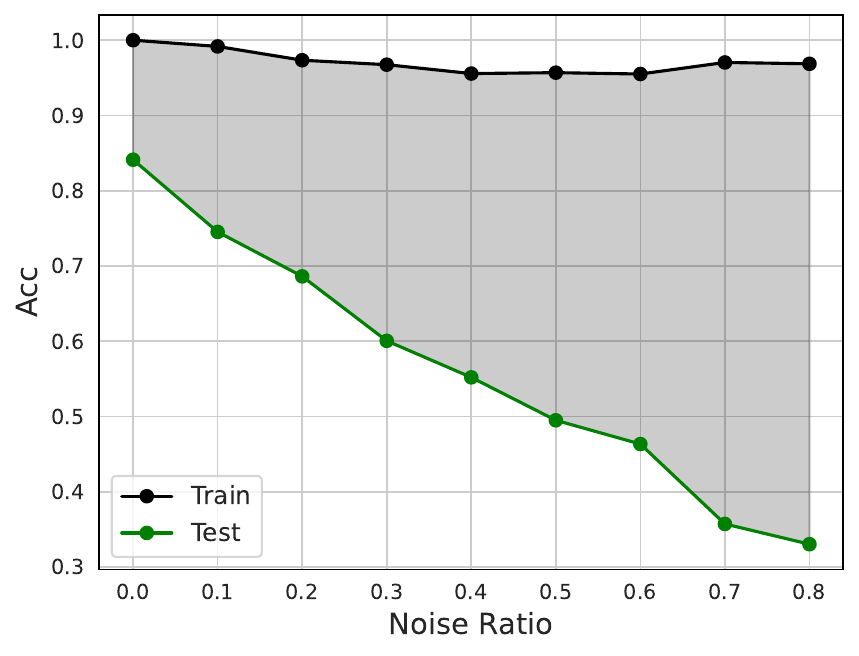}
        \caption{MUTAG}
        \label{fig:sub1}
    \end{subfigure}
    \hspace{0.00\textwidth} % 添加水平间距
    \begin{subfigure}{0.25\textwidth}
        \centering
        \includegraphics[width=\textwidth]{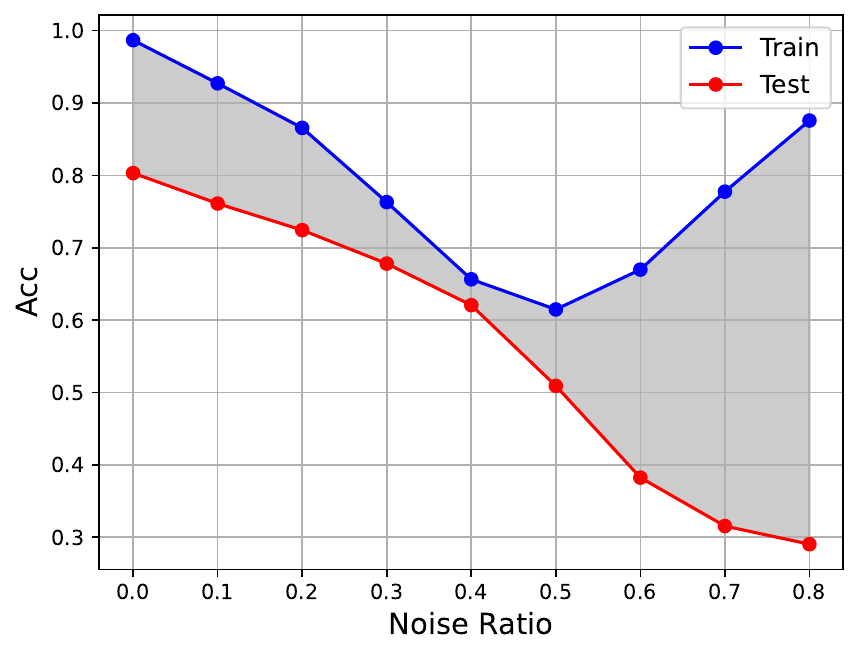}
        \caption{NCI1}
        \label{fig:sub2}
    \end{subfigure}
    \hspace{0.00\textwidth} % 添加水平间距
    \begin{subfigure}{0.45\textwidth}
        \centering
        \includegraphics[width=\textwidth]{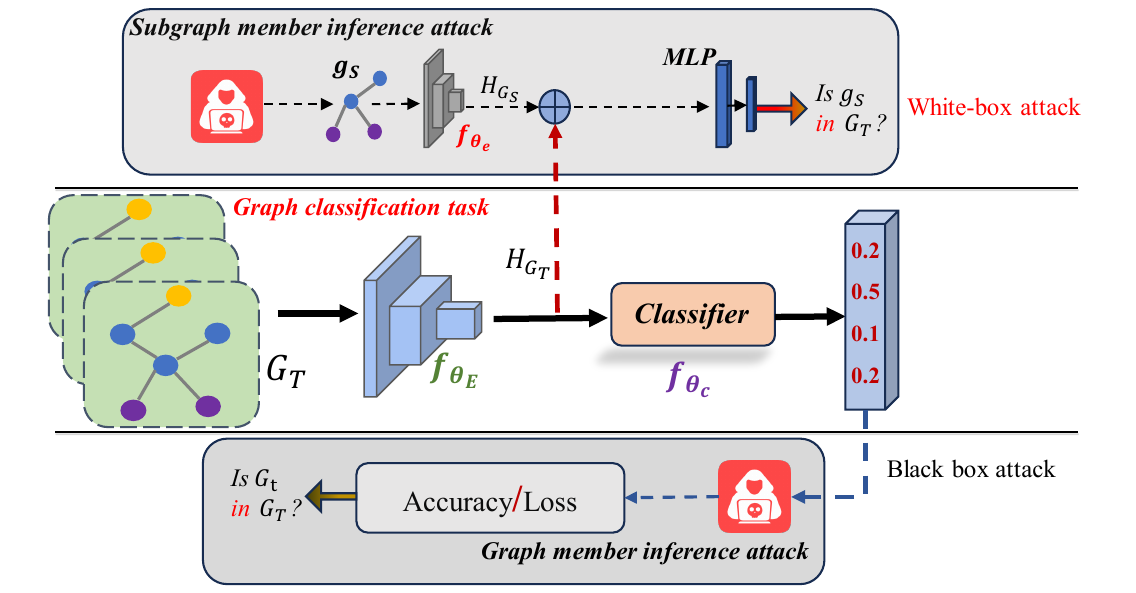}
        \caption{Membership inference attacks in graphs}
        \label{fig:sub4}
    \end{subfigure}
    \caption{(a)-(b) represent the training and testing accuracies of GIN under different noise label rates on the MUTAG and NIC1 datasets, respectively, where the shaded portion indicates the absolute gap between the training and testing accuracies, and the larger the gap indicates that the model overfitting is more serious; (c) is the subgraph membership inference attack based on the embedding information in the white-box attack scenario as well as the graph membership inference attack based on the model's output information (accuracy or loss) in the black-box attack scenario.}
    \label{fig:main_figure}
\end{figure*}

\section{Preliminaries}
\label{se3}
\subsection{Problem definition}
First, we formalize the notation and problem definition. Let $G = (V, E, X, Y)$ be a graph, where $V$, $E$, and $Y$ denote the set of nodes, edges, and the graph label, respectively. We use $x_v$ to define the attribute vector of node $\boldsymbol{v}$, $\mathbf{X} \in R^{|V| \times d}$ represents the node attribute matrix, and $d$ to be the dimension of the node attribute vector. In the task of graph classification with noisy labels, we have graph datasets $\mathcal{G}=\left\{G_1, G_2, \ldots, G_n\right\}$, where some of the graph data labels are not equivalent to their true labels. Our task is to accurately predict the graph labels in the test set by training an encoder $f_{\theta_E}$ and classifier $f_{\theta_C}$ in an end-to-end manner.

\subsection{Research objectives}
In this study, we re-think the impact of noise labels on graph classification from the perspective of data privacy and model utility, thus causing more researchers to study the classification of graphs with noise labels. Therefore, we propose a robust graph neural network method with noisy label graph classification. The main objectives are: (1) Using the existing member inference attack methods to measure the privacy attack capability of graph and subgraph level member inference attacks under noisy label graph classification. (2) The noise label screening method of the class-first feature principal component vector and the noise label correction method guided by the double view are designed. (3) Based on the modified supervised information, supervised graph Contrastive learning is proposed to improve the robustness of graph embedding. (4) Evaluate the effectiveness and privacy performance of our methods.

\textit{It is worth noting that in this article, we are not specifically addressing the problem of member inference attacks on graph classification. On the premise of label learning with noise, we analyze and alleviate the graph privacy problems caused by label damage.}

\subsection{ Message Passing Neural Networks}
We briefly introduce the message-passing neural network for generating graph-level representations. This network aggregates information from the domain along the structure of the graph topology to achieve an update to the target node $\boldsymbol{v}_i$ with the following update formula:

\begin{equation}
\begin{aligned}
\boldsymbol{v}_{N\left(v_i\right)}^{(l)} & =\operatorname{AGGR}^{(l)}\left(\left\{\boldsymbol{v}_i^{(l-1)}: j \in \mathcal{N}(i)\right\}\right) \\
\boldsymbol{v}_i^{(l)} & =\operatorname{COMB}^{(l)}\left(\boldsymbol{v}_i^{(l-1)}, \boldsymbol{v}_{N\left(v_i\right)}^{(l)}\right)
\end{aligned}
\label{eq2}
\end{equation}
where $\boldsymbol{v}_i^{(l)}$ denotes the embedded representation of the node $\boldsymbol{v}_i$ at the $l$-th level. The $\operatorname{AGGR}^{(l)}(\cdot)$ and $\operatorname{COMB}^{(l)}(\cdot)$ denote the aggregation and combination operations at the $l$-th level, respectively. In order to obtain an embedded representation of the whole graph, global pooling operations are introduced to the last layer of all node representations:
\begin{equation}
\boldsymbol{Z}=\operatorname{GOPL}\left(\left\{\boldsymbol{v}_i^{(l)}\right\}_{i=1}^n\right)
\label{eq3}
\end{equation}
where $\mathrm{GOPL}(\cdot)$ denotes the global pooling operation. Max pooling and Mean pooling are commonly used to obtain graph-level representations. In addition, more advanced pooling methods such as Hierarchical Pooling, Differential Pooling, and MinCut Pooling are used to obtain graph-level representations from different perspectives \citep{DBLP:conf/nips/YingY0RHL18,bianchi2020spectral}.

\section{METHODOLOGY}
\label{se4}
In this section, we first analyze how noisy labels enhance the capability of membership inference attacks on graphs or subgraphs in graph classification tasks from the data privacy perspective. Then, we detail each submodule of the RGLC method.

\begin{figure*}[!h]
    \centering
    \begin{subfigure}{0.31\textwidth}
        \centering
        \includegraphics[width=\textwidth]{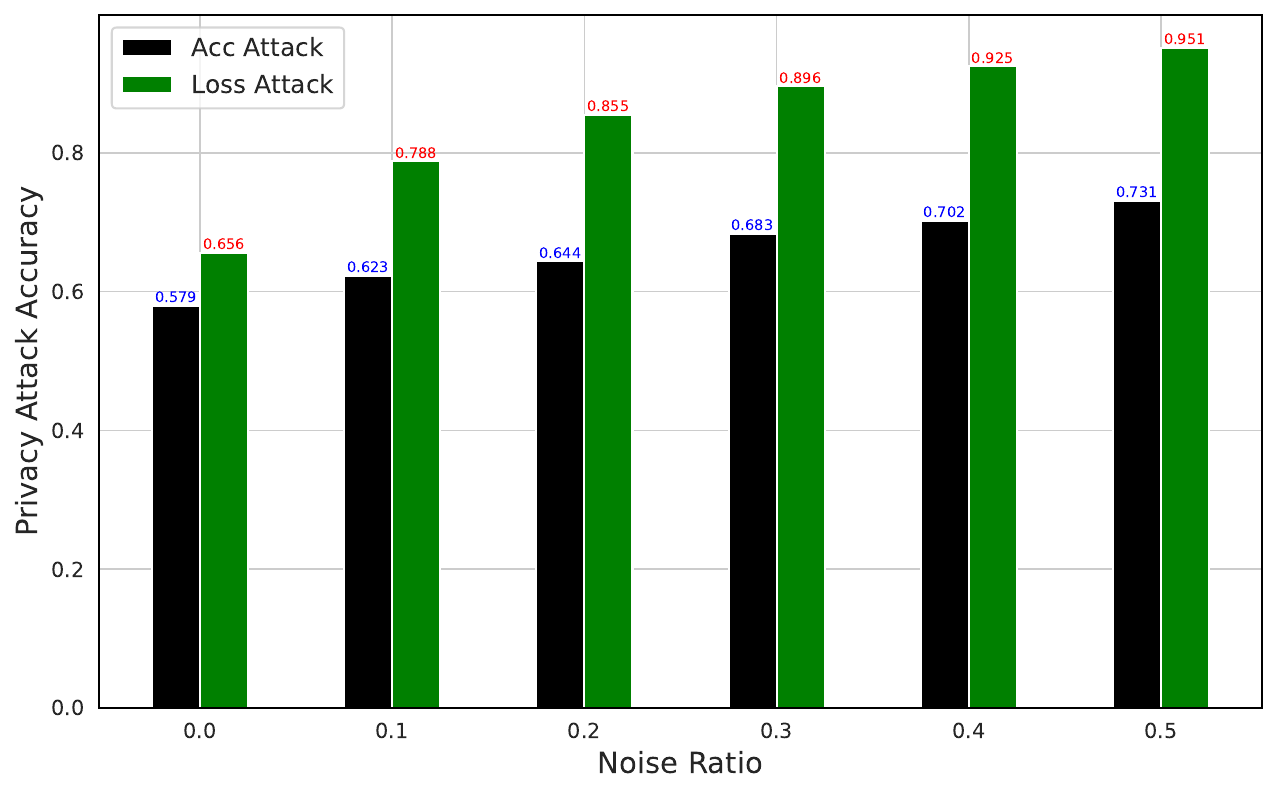}
        \caption{MUTAG}
        \label{fig:sub3}
    \end{subfigure}
    \hspace{0.00\textwidth} % 添加水平间距
    \begin{subfigure}{0.31\textwidth}
        \centering
        \includegraphics[width=\textwidth]{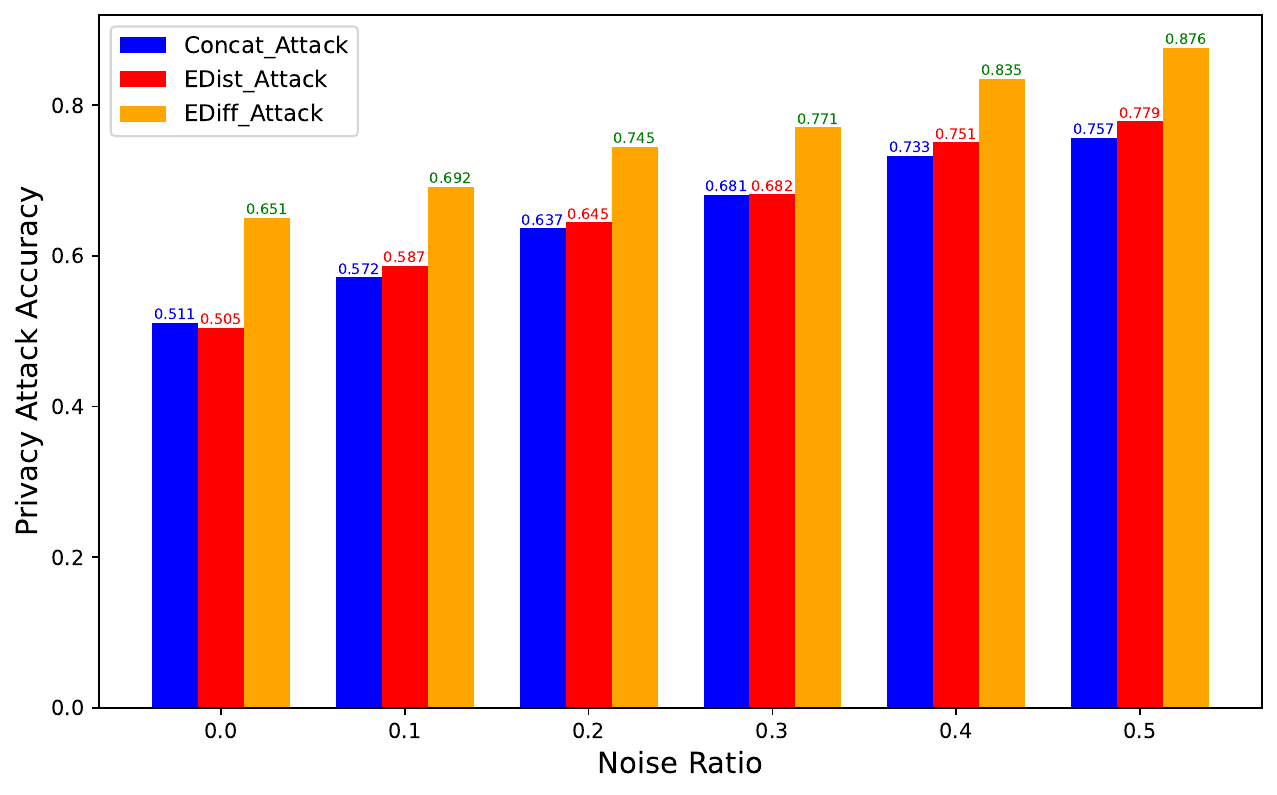}
        \caption{NCI1}
        \label{fig:sub4}
    \end{subfigure}
    \hspace{0.00\textwidth} % 添加水平间距
    \begin{subfigure}{0.327\textwidth}
        \centering
        \includegraphics[width=\textwidth]{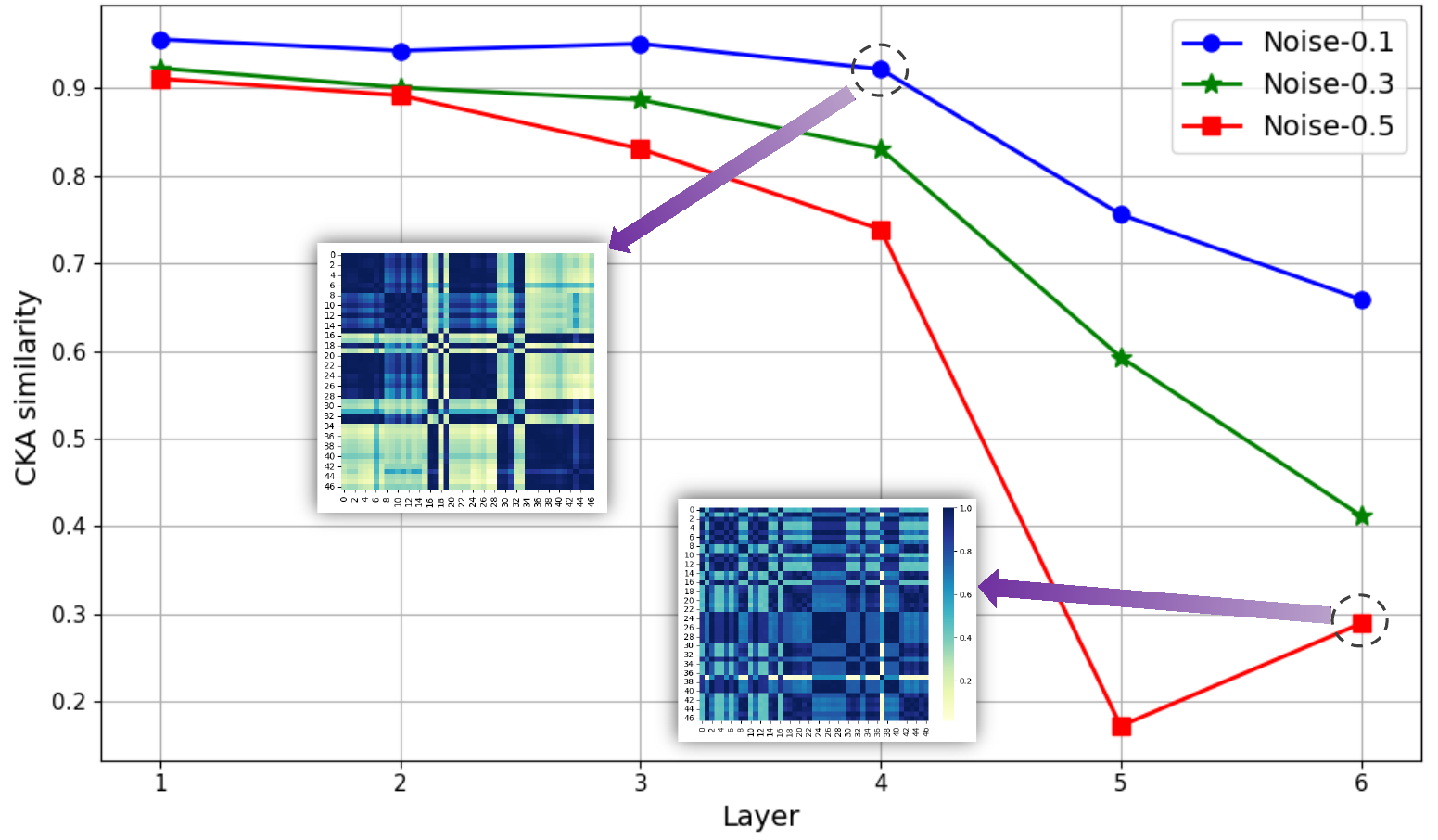}
        \caption{NCI1}
        \label{fig:sub4}
    \end{subfigure}
    \caption{Attack accuracy of graph membership inference attacks or subgraph membership inference attacks at different noise rates. (a): graph membership inference attack using model-based accuracy as well as loss value as metrics in black-box attack scenarios in the MUTAG dataset; (b): subgraph membership inference attack using embedding information as metrics in white-box attack scenarios in the NCI1 dataset; (c):CKA similarity of different layers under clean and noisy graph data.}
    \label{fig:main_figure}
\end{figure*}
\subsection{Analyzing the Impact of Noisy Labels in Graph Classification Tasks from a Data Privacy Perspective}
\label{sub1}

It is well known that noisy labels cause the model to constantly deviate from the original decision boundaries during the training phase, which leads to poor generalization performance during the testing phase \citep{li2022selective}. In the field of machine learning, the overfitting phenomenon has been verified to lead to further amplification of model members' private information. To explore the membership privacy risk of graph neural networks under noisy labels, we adopt the dataset partitioning approach in our previous work \citep{DBLP:conf/uss/LiuWH000CF022,liu2022membership}, where the target training set and the shadow dataset are each half of the original dataset and do not overlap. Fig. 2 (a) and (b) show GIN's training and testing accuracies under different noise rates for the MUTAG and NCI1 datasets, where the shaded area indicates the accuracy gap. We found that in graph classification tasks based on graph neural networks, the noise label produces a serious overfitting phenomenon for smaller graph data (e.g., MUTAG), especially when the noise rate becomes progressively larger, the overfitting phenomenon becomes more and more obvious, which enhances the ability of the inference attack of graph members and seriously threatens the privacy information of the training dataset.
Meanwhile, we also find that in larger graph datasets (e.g., NCI1), when the network structure of the model remains unchanged, both the training accuracy and the testing accuracy decrease when the noise rate is lower than 0.5. In contrast, when the noise rate is higher than 0.5, the training accuracy gradually increases while the testing accuracy decreases. In the experiment, we use the same noise rate setting as the OMG method \citep{yin2023omg}, that is, the noise rate is between 0 and 0.5. Therefore, the conventional quantitative method of member privacy measurement (e.g., cross-entropy, confidence scores) \citep{song2019privacy} will not be able to truly measure the risk of member privacy disclosure in this case.

For this reason, we utilize the method proposed by \cite{zhang2022inference} to measure the subgraph membership inference attack on data in the embedding space and determine whether the subgraph $g_{s}$ is contained in the target training set $G_{T}$ by the information in the embedding space. We follow the attack method in \citep{zhang2022inference} and use the random wandering subgraph sampling method, the pooling method of DiffPool, and a subgraph sampling rate of 0.4 for preliminary experimental validation. Fig. 2 (c) shows the membership attack methods we adopted in two dataset sizes. We used the black-box scenario in the smaller graph dataset to initiate the graph membership inference attack based on the model output information. In comparison, we used the subgraph membership inference attack in the larger graph dataset based on the embedding information in the white-box scenario. Fig. 3 (a) shows the results of the graph membership inference attack using model output precision and loss value as the attack method, from which it can be seen that compared to the attack method based on precision difference, the loss value-based attack is more effective. It can effectively attack the privacy information of the original training data at a lower noise rate. Fig. 3 (b) shows the experimental results of subgraph inference attacks on the NCI1 dataset using the three different embedding aggregation strategies set in \citep{zhang2022inference}. From Fig. 3(b), we know that noisy labels make the model more vulnerable to subgraph membership inference attacks than the original clean labels, and the subgraph inference membership inference attacks are more potent with the gradual increase of the noise rate. We quantified the risk level of graph or subgraph membership inference attacks on the graph classification task under noise labels through the model's output spatial information or embedded spatial information. We found that the noise label brings more serious privacy leakage to the training data in the graph classification task, highlighting the importance of studying the graph classification task with noise labels from the perspective of data confidentiality.

\begin{figure*}[!h]
    \centering
    \begin{subfigure}{1\textwidth}
        \centering
        \includegraphics[width=\textwidth]{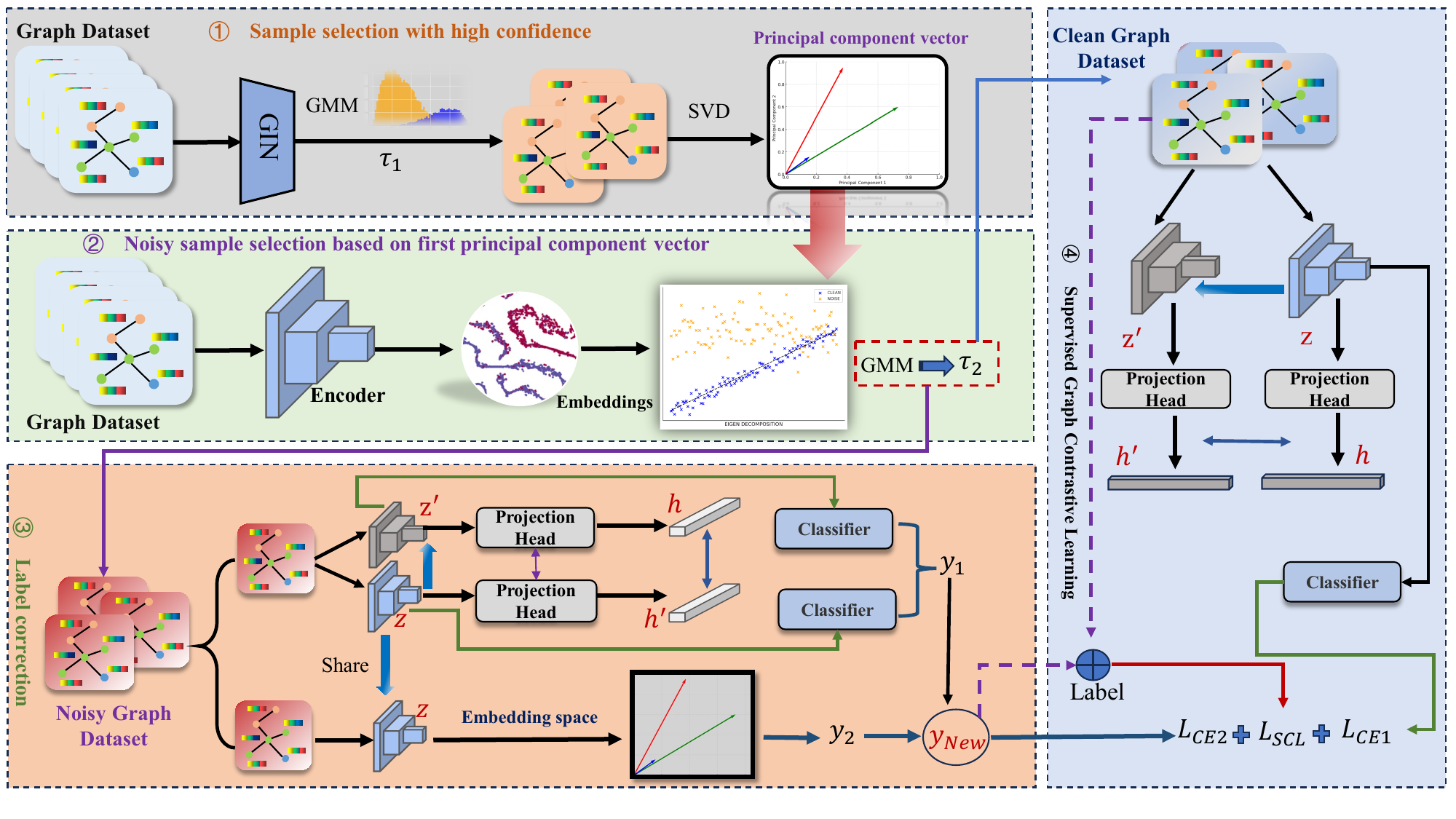}
        \label{fig:sub3}
    \end{subfigure}
    \caption{The framework of the RGLC method is shown as the process of one round. It mainly includes the selecting high confidence samples to obtain the first feature principal component vector of each class, noise sample selection based on the first feature principal component vector of the class, label correction guided by two-space information, and supervised graph contrastive learning.}
    \label{fig:main_figure}
\end{figure*}
Next, we analyze why the model is inaccurate enough to measure member privacy leakage using output spatial information when the noise label rate is less than 0.5 in a large dataset. Intuitively, the model will fit the noise label in the training process so that the model has better training accuracy and poor test accuracy, all of which are premised on the ability of the model's parameters to generalize the current data distribution. For small data sets, it is easier for the model to fit the distribution of its training data, which leads to a severe overfitting phenomenon in the noise label scenario. However, by learning limited data knowledge, the model can achieve a specific generalization ability in a large data set. When we fixed the model's training rounds and parameter settings, we found that the model would not fit the error information of the noise label, resulting in the under-fitting phenomenon under the noise label. This is similar to the use of differential privacy technology in deep learning to reduce the member inference attack ability of the model \citep{ghazi2021deep}, but the cost is the reduction of test accuracy. However, when we adopt the white box attack hypothesis, that is, to get the embedded information of the training data, we can still produce a strong member inference attack capability. The main reason is that the embedded information is generated by the encoder $f_{\theta_E}$, while the output information is generated by the classifier $f_{\theta_C}$, in which softmax function is often used to obtain the class probability distribution. However, due to the class-related properties of the softmax function \citep{yi2023class}, it is sensitive to noise labels, resulting in misclassification of one class and penalizing the distribution of the other classes. Therefore, compared with the larger offset of the output space, the embedded spatial information is less offset in the noisy and clean samples. To verify the deviation of different layers, we use centralized kernel alignment (CKA) \citep{kornblith2019similarity} to measure the representational similarity between the corresponding layers of the clean graph data extracted by the model and the noise graph data, which is calculated as follows:

\begin{equation}
\operatorname{CKA}(Z_1, Z_2)=\frac{\left\|Z_1^T Z_2\right\|_F^2}{\left\|Z_1^T Z\right\|_F^2\left\|Z_2^T Z_2\right\|_F^2}
\end{equation}
where $ Z_1$ and $Z_2$ represent the data representation information of different models, the score of CKA similarity ranges from 0 to 1, and the similarity increases with the value increase. Fig.3 (c) shows the similarity between embedding the original clean graph data and embedding the noise graph data at different noise label rates. We can see that the similarity decreases with the number of layers increase, and the similarity of the output layer is the lowest. At the same time, we use cosine similarity to calculate the similarity between the graph embeddings obtained at the fourth layer of the model when the noise label rate is 0.1 and visualize the similarity matrix. From the results, the model can now distinguish the same category and different categories of graph data. However, when the noise rate is 0.5, the ability of the output layer of the model to distinguish the graph data drops sharply, resulting in poor classification performance. There are also similar phenomena in the field of image learning with noisy labels \citep{tam2023federated,lee2019robust}. The above experiments indicate that in large graph datasets, the embedded space information has more obvious discriminative power in noisy scenarios compared with the output space information. Therefore, subgraph membership inference attacks based on the white-box attack in the graph embedding space can more accurately verify the risk of member privacy leakage in large graph classification datasets with noisy labels. In deep learning image research, studies \citep{tramer2022truth} and \citep{chen2022amplifying} also share similar views. They utilize data poisoning to amplify the privacy leakage risks of original training data members.

\subsection{Robust graph neural networks with noisy label learning}
This paper adopts research directions of noisy sample selection and noise label correction to mitigate the impact of noisy labels in graph classification. In terms of noise sample selection, the first feature principal component vector of each category is constructed to complete the accurate screening of noise samples based on the high confidence samples with small loss screening. Regarding the correction of noisy sample labels, noise label correction under a dual perspective is realized by leveraging the information from the principal component vectors in the embedding space and the information in the model output space under data augmentation. Finally, we introduce supervised graph contrastive learning to enhance the model's representation capability. A framework diagram of our approach is shown in Fig. 4, with specific methodological details as follows.

\subsubsection{Two-stage noise sample selection method}
To realize the robust separation of noisy samples, we construct the corresponding Gram matrix in the embedding space according to the category information, use the method of matrix decomposition (e.g., SVD) to obtain the first principal component vector (corresponding to the largest eigenvalue) representing each category, then calculate the degree of alignment of the embedding of the samples with their category principal component vectors and obtain the alignment results of each sample, and finally realize the clean samples and noisy samples through the unsupervised clustering algorithm GMM. In addition, to reduce the computational load of eigenmatrix decomposition and improve the quality of the first principal component vector, before using the first principal component vector to screen the noise samples, we first obtained some samples with high confidence by setting a larger threshold according to the small loss criterion and then used the samples with high confidence to construct the Gram matrix embedded in the graph according to the category conditions. The specific method is as follows.

\begin{algorithm}[!h]
\caption{\emph{Two-stage noise sample selection method}}
\SetAlgoLined % 开启算法的竖线
\KwIn{Graph dataset $\mathcal{G}$ with noise labels, thresholds $\tau_1$ and $\tau_2$, encoder $f_{\theta_E}$, classifier $f_{\theta_C}$.}
\KwOut{Graph datasets $\mathcal{G}^c$ and $\mathcal{G}^n$.}

The loss value $\mathcal{L}$ for each graph data is calculated by Eq. \ref{eq4} and the probability of each sample belonging to a small loss-mean component $\left\{w_i\right\}_{i=1}^N$ is obtained using the GMM algorithm using Eq. \ref{eq5}\;
Set the threshold $\tau_1$ to obtain high confidence samples  $\mathcal{G}^{\prime}=\left\{G_i \mid G_i \in \mathcal{G}, w_i>\tau_1\right\}$\;
For the graph data $\mathcal{G}^{\prime}$ use the encoder $f_{\theta_E}$ to obtain the graph embedding via Eq. \ref{eq7} and construct the Gram matrix $\left\{\mathcal{M}_{y_i}\right\}_{i=1}^K=Z_{y_i} Z_{y_i}{ }^T$ for the corresponding category, and obtain the first principal component vector $\left\{u_{y_i}\right\}_{i=1}^K$ corresponding to each category via matrix decomposition\;
The embedding of the graph data $\mathcal{G}$ is obtained using the encoder $f_{\theta_E}$, and the squares of the inner product values of the graph embeddings with their corresponding category principal component vectors $s_i=\left\langle u_{y_i}, z_i\right\rangle^2, i=1, \ldots, N$ are computed by Eq. \ref{eq9}\;
The posterior probability that each graph data belongs to a larger mean component is obtained by Eq. \ref{eq10}, and the clean labeled graph data $\mathcal{G}^c$ and the noisy labeled graph data $\mathcal{G}^n$ are filtered out by using a set threshold $\tau_2$\;
\label{algo1}
\end{algorithm}

For the graph dataset $\mathcal{G}$ containing noisy labels, we use the output of the model as well as the given labels to compute the loss values and fit the loss distribution via a GMM, where $w$ is the posterior probability of fitting a low-mean component with small loss, which allows us to provide high-confidence clean samples  $\mathcal{G^\prime}$  for the computation of the eigenprincipal component vectors:
\begin{equation}
\mathcal{L}=\left\{\ell_i\right\}_{i=1}^N=\left\{-Y_i^T \log \left(p\left(G_i ; f_{\theta_e} ,f_{\theta_c}\right)\right)\right\}_{i=1}^N
\label{eq4}
\end{equation}
\begin{equation}
\left\{w_i\right\}_{i=1}^N=G M M\left\{\ell_i\right\}_{i=1}^N=\left\{p\left(g_1 \mid \ell_i\right)\right\}_{i=1}^N
\label{eq5}
\end{equation}
\begin{equation}
\mathcal{G}^{\prime}=\left\{G_i \mid G_i \in \mathcal{G}, w_i>\tau_1\right\}
\label{eq6}
\end{equation}
where $\tau_1$ is the probability threshold that the sample belongs to the small loss-mean component $g_1$, to obtain high-confidence samples and reduce the computational complexity of matrix decomposition, we set $\tau_1=0.7$ in this paper.

Next, we use the encoder $f_{\theta_E}$ to obtain the embedding $Z$ of the clean sample $\mathcal{G^\prime}$ and construct the corresponding Gram matrix according to the category information:
\begin{equation}
Z=\left\{z_i\right\}_{i=1}^M, z_i=\left\{f_{\theta_E}\left(G_i\right) \mid G_i \in \mathcal{G}^{\prime}\right\}
\label{eq7}
\end{equation}
\begin{equation}
\left\{\mathcal{M}_{y_i}\right\}_{i=1}^K=Z_{y_i}Z_{y_i}{ }^T 
\label{eq8}
\end{equation}
where $\mathcal{M}_{y_i}$ denotes the Gram matrix for the $i$-th category, K denotes the number of categories, and M represents the number of clean samples. Then, we apply matrix decomposition to the Gram matrix of each category, obtaining the first principal component vectors $\left\{u_{y_i}\right\}_{i=1}^K$. We calculate the square of the inner product between the graph embeddings and the category's first principal component vectors based on category information. By fitting these calculations with a GMM model, we determine the probability threshold for each graph dataset belonging to the larger mean component $g_2$. Finally, by setting a probability threshold $\tau_2$, we achieve the separation of noisy and clean samples:
\begin{equation}
s_i=\left\langle u_{y_i}, z_i\right\rangle^2, i=1, \ldots, N
\label{eq9}
\end{equation}
\begin{equation}
\left\{w_i\right\}_{i=1}^N=G M M\left\{s_i\right\}_{i=1}^N=\left\{p\left(g_2 \mid s_i\right)\right\}_{i=1}^N
\label{eq10}
\end{equation}
\begin{equation}
\mathcal{G}^c=\left\{G_i \mid G_i \in \mathcal{G}, w_i \geq \tau_2\right\}
\label{eq11}
\end{equation}
\begin{equation}
\mathcal{G}^n=\left\{G_i \mid G_i \in \mathcal{G}, w_i<\tau_2\right\}
\label{eq12}
\end{equation}
where $u_{y_i}$ denotes the principal component vector corresponding to each graph data, $\mathcal{G}^c$ and $\mathcal{G}^n$ denote the clean-labeled and noise-labeled graph data, respectively. Algorithm 1 shows the entire computational process of noise sample selection. Proposition \ref{pro1} shows that the two-stage noise sample selection method can efficiently filter clean and noisy graph data in the graph embedding space, as demonstrated in \ref{app3}.

\begin{proposition}
\textbf{ In graph learning with noisy labels, the first principal component vector obtained through matrix decomposition of high-confidence data can effectively separate clean and noisy data in the graph embedding space.} 
\label{pro1}
\end{proposition}

\subsubsection{Label Correction Mechanism Based on Dual Space Perspective}
In order to fully utilize the knowledge of the noisy labeled graph dataset $\mathcal{G}^n$, we fuse the discriminative information of the embedding space and the output space to correct the label of the noisy graph data. On the embedding space, label correction is performed by measuring the degree of alignment of the graph embedding $Z$ to the corresponding category feature principal component $\left\{u_{y_i}\right\}_{i=1}^K$. In the output space, different from correcting labels by using model prediction output in the past, we construct both original and enhanced view angles of graph data to obtain stable model prediction results for label correction. Ultimately, we adjust the weights of the category probability distribution based on the feature space and the category probability distribution based on the model prediction output to achieve more accurate label correction results.

\textbf{Label correction strategy based on embedding space.} For each graph data $G_i$ in the noisy labeled graph dataset $\mathcal{G}^n$, we perform the label correction by calculating the square of the inner product of the noisy graph embedding and the vector of principal components of each category feature:
\begin{equation}
{y_i}^f=\left[p_{i 1}, p_{i 2}, \ldots, p_{i k}\right], p_{i k}=\frac{\exp \left(\left\langle u_k, z_i\right\rangle^2\right)}{\sum_{j=1}^k \exp \left(\left\langle u_j, z_i\right\rangle^2\right)}
\label{eq13}
\end{equation}
where $p_{ik}$ denotes the probability that the $i$-th noise graph data belongs to the $k$-th category.

\begin{algorithm}[!h]
\caption{\emph{Label Correction Mechanism Based on Dual Space Perspective}}
\SetAlgoLined % 开启算法的竖线
\KwIn{Noisy labeled dataset $\mathcal{G}^n$, vector of category feature principal components $\left\{u_{y_i}\right\}_{i=1}^K$, encoder $f_{\theta_E}$, classifier $f_{\theta_C}$, noise perturbation coefficients $\eta$, and weight balancing coefficients $\lambda$.}
\KwOut{Noisy labeled dataset $\mathcal{G}^n$ corrected labels $y^{new}$.}
\tcc{Embedding of noisy graph datasets obtained through encoder}
\For{$G_i \in \mathcal{G}^n$}{$z_i=f_{\theta_E}\left(G_i\right)$}
\tcc{Noise labels correction in the embedding space}
\For{$z_i \in Z^n$}{$y_i{ }^f=\left[p_{i 1}, p_{i 2}, \ldots, p_{i k}\right], p_{i k}=\frac{\exp \left(\left\langle u_k, z_i\right\rangle^2\right)}{\sum_{j=1}^k \exp \left(\left\langle u_j, z_i\right\rangle^2\right)}$}\
The encoder is perturbed by Eq. \ref{eq15}: $f_{\theta_E}^{\prime}=f_{\theta_E}+\eta \cdot \Delta f_{\theta_E}$\;
\tcc{Obtaining the predicted output probability distribution for the original-enhanced view}
\For{$G_i \in \mathcal{G}^n$}{$y_i^{O_1}=f_{\theta_C}\left(f_{\theta_E}\left(G_i\right)\right), y_i^{O_2}=f_{\theta_C}\left(f_{\theta_E^{\prime}}\left(G_i\right)\right)$\;$y_i{ }^O=\operatorname{Sharpen}\left(\overline{y_i{ }^O}, T\right)$\;}
\tcc{Adjusting label weights for two-space corrections using balancing coefficients $\lambda$}
\For{$G_i \in \mathcal{G}^n$}{$y_i{ }^{new}=(1-\lambda) y_i{ }^f+\lambda \cdot y_i{ }^O$}
\label{algo2}
\end{algorithm}
\textbf{Label correction strategy based on output space.}  Inspired by the semi-supervised learning MixMatch \citep{berthelot2019mixmatch}, we use graph data augmentation technology to achieve label correction of noisy samples in the output space. In addition, we introduce supervised graph contrastive learning in this paper to improve the embedding quality of the model. Therefore, data enhancement in graph contrastive learning (e.g., edge discarding or subgraph sampling) can be used to obtain prediction outputs under different enhanced views, and the prediction average of multiple views can be used for label correction. However, in graph classification with labeled noise learning, the model is prone to fitting noisy labels, and direct use of enhancement operations on the original graph data may fail to ensure the consistency of the spatial semantics of the outputs, which results in the accumulation of erroneous information during the training process. For this reason, we draw on the enhancement method in SimGRACE \citep{xia2022simgrace}, a graph contrastive learning, and utilize the addition of noise to the encoder to construct an enhanced view, in order to avoid the prediction inconsistency caused by direct enhancement operations on the original graph data. The classifier generates the predictive distributions of the original graph and the augmented view, thus realizing the label correction of the noisy graph data in the output space:
\begin{equation}
y_i^{O_1}=f_{\theta_C}\left(f_{\theta_E}\left(G_i\right)\right), y_i^{O_2}=f_{\theta_C}\left(f_{\theta_E}^{\prime}\left(G_i\right)\right)
\label{eq14}
\end{equation}
where $y_i{ }^{O_1}$ and $y_i{ }^{O_2}$ are the model's predicted probability distributions for the original and augmented views, respectively, and $f_{\theta_E}^{\prime}$ is the encoder subjected to the perturbation, which is computed as:
\begin{equation}
f_{\theta_E}^{\prime}=f_{\theta_E}+\eta \cdot \Delta f_{\theta_E} ; \Delta f_{\theta_E} \sim N\left(0, \sigma^2\right)
\label{eq15}
\end{equation}
where $\eta$ denotes the noise perturbation scale. Then, the predicted probability distributions of the two views were averaged and temperature sharpening operations were utilized to obtain labels in the noisy labeled graph data corrected in the output space:
\begin{equation}
y_i{ }^O=\operatorname{Sharpen}\left(\overline{y_i{ }^O}, T\right) ; \overline{y_i{ }^O}=\frac{y_i{ }^{O_1}+y_i{ }^{O_2}}{2}
\label{eq16}
\end{equation}
where $T$ is the sharpening parameter, and in this paper, we set $T$=0.5 by default. Ultimately, we adjust the weights of the corrected labels based on the feature space and the output space by setting the balancing coefficients $\lambda$, and their corrected labels are expressed as:
\begin{equation}
y_i{ }^{new}=(1-\lambda) y_i{ }^f+\lambda \cdot y_i{ }^O
\label{eq17}
\end{equation}
where $y_i{ }^{new}$ denotes the corrected label of the $i$-th graph data $G_i$ in the noisy labeled graph dataset $\mathcal{G}^n$. We use the curriculum learning to update the value of $\lambda$, i.e., $\lambda = e / E * 0.6$, with $e$ and $E$ denoting the current and maximum rounds of model training, respectively. Algorithm 2 shows a label correction mechanism based on a two-space perspective.

\subsubsection{Privacy Preservation Based on Supervised Graph Contrastive Learning}
Based on the noise sample separation and label correction mechanism, we can obtain a clean graph dataset $\mathcal{G}_c$ and a label-corrected graph dataset $\mathcal{G}_r$ in each training round. To attenuate the ability of subgraph inference attacks based on embedding space, we introduce supervised graph contrastive learning, which constructs the corresponding positive-negative sample pairs by category information, thus enhancing the ability of the encoder to learn the graph representation, attenuating the discrepancy between the graph representations of training and prediction phases, and effectively mitigating the privacy leakage of training data.

Specifically, for the graph dataset $\mathcal{G}_{\text {new }}=\left\{\mathcal{G}_c, \mathcal{G}_r\right\}$, we utilize the graph enhancement approach in SimGRACE and utilize the label information to achieve supervised graph contrastive learning. Specifically, we use the label category information of each graph data to bring the feature distances of similar graph data closer and push the feature distances of different classes of graph data farther in the feature space, and the supervised graph contrastive loss is as follows:
\begin{equation}
z_i=f_{\theta_E}\left(G_i\right), z_i{ }^{\prime}=f_{\theta_E}^{\prime}\left(G_i\right)
\end{equation}
\begin{equation}
h_i=f_{\theta_P}\left(z_i\right), h_i{ }^{\prime}=f_{\theta_P}\left(z_i{ }^{\prime}\right)
\end{equation}
\begin{equation}
\ell_i=-\sum_{i^{\prime}=1}^N \mathbf{1}_{i \neq i^{\prime}} .\mathbf{1}_{y_i=y_{i^{\prime}}} \cdot \log \frac{\exp \left(\operatorname{sim}\left(h_i, h_{i^{\prime}}\right) / \tau\right)}{\sum_{i^{\prime}=1, i^{\prime} \neq i}^N \exp \left(\operatorname{sim}\left(h_i, h_{i^{\prime}}\right) / \tau\right)}
\label{eq20}
\end{equation}
where $\ell_i$ is the contrast loss of the $i$-th graph data $G_i$ of the graph dataset $\mathcal{G}_{new}$, $f_{\theta_P}$ is the nonlinear projection head, and $\tau$ is the temperature parameter.

Next, we briefly explain why supervised graph contrastive learning can mitigate subgraph membership inference attacks based on graph embedding information. Firstly, we review the three subgraph member inference attacks based on graph embeddings proposed in \cite{zhang2022inference}: the connection attack (i.e., $\chi=H_{G_T} \| H_{\mathcal{G}_S}$), the per-positional difference attack (i.e., $\chi=H_{G_T}-H_{G_S}$), and the Euclidean distance attack (i.e., $\chi= \left\|H_{G_T}-H_{\mathcal{G}_S}\right\|_2$), where $H_{G_T}$ denotes the embedding of the target graph data and $H_{G_S}$ denotes the embedding of the subgraphs. According to the results in Fig. 3(b), it can be seen that the difference-based attack \textit{EDiff\_Attack} and the Euclidean distance-based attack \textit{EDist\_Attack} are more effective. Therefore, we will analyze these two attacks as the standard next. The essence of the above two attacks is the more significant variability of the graph embeddings of the training graph data and the test graph data extracted by the encoder $f_{\theta_E}$, which leads to the enhanced success rate of the subgraph inference attack. The essence of graph contrastive learning is to optimize the encoder's representational capability by improving the view to have superior \textit{alignment} and \textit{uniformity} \citep{wang2020understanding}, while we further introduce label information to construct supervised graph contrastive learning so that the embedding vectors of graph data in the same category are more similar to each other. The embedding vectors of graph data in different categories can be more uniformly distributed on the feature hypersphere, thus indenting the training and testing graph data embedding variability of the training and test graph data, weakening the ability of inference attack by subgraph members. In addition, the introduction of supervised graph contrastive learning will make the model learn more robust graph embeddings, which will help the model's label correction for noise-labeled graph data, improving the model's ability to fight against noise labels while reducing the ability of output-space-based graph membership inference attacks.

\begin{table*}
  \centering
  \caption{Detailed statistical information on the graph datasets. }
    \begin{tabular}{ccccccccc}
    \toprule
    \textbf{Datasets} & \textbf{MUTAG} & \textbf{NCI1} & \textbf{PTC} & \textbf{COX2} & \textbf{PROTEINS} & \textbf{PROTEINS\_F} & \textbf{IMDB-B} & \textbf{IMDB-M} \\
    \midrule
    Graphs Num & 188   & 4110  & 344   & 467   & 1113  & 1113  & 1000  & 1500 \\ 
    Attribute Dim & 7     & 37     & 18     &1    & 3     & 1     & 0    & 0 \\
    Avg.Nodes & 17.93 & 29.87 & 14.29 & 41.22 & 39.06 & 39.06 & 19.77 & 13 \\
    Avg.Edges & 19.79 & 32.3  & 14.69 & 43.45 & 73.82 & 73.82 & 96.53 & 65.94 \\
    \bottomrule
    \end{tabular}%
  \label{tab:addlabel}%
\end{table*}%

\begin{algorithm}
\caption{\emph{Robust graph neural networks with noisy label learning}}
\SetAlgoLined % 开启算法的竖线
\KwIn{Graph dataset $\mathcal{G}$ with noise labels, thresholds $\tau_1$ and $\tau_2$, encoder $f_{\theta_E}$, classifier $f_{\theta_C}$, projected head $f_{\theta_P}$, batch size $B$, step size of a single round $iter_{max}$, noisy perturbation coefficients $\eta$, weight balancing coefficients $\lambda$, Loss-constrained hyperparameters $\alpha$ and $\beta$, warm-up rounds $T_l$, and maximum training rounds $T_{L}$.}
\KwOut{The final parameters of the model $f_{\theta_E}^{T_L}, f_{\theta_C}^{T_L}$. }
\tcc{Perform warm-up operations on the model}
\For{$t \leftarrow 1,2, \ldots, T_l$}{$f_{\theta_E}^t, f_{\theta_C}^t \leftarrow \sum_{G_i \in \mathcal{G}} l_{C E}\left(G_i, f_{\theta_E}, f_{\theta_c}\right)$}
\For{$t \leftarrow T_{l+1}, T_{l+2}, \ldots, T$}
{Noisy samples are screened using Algorithm 1 to obtain a clean graph dataset $\mathcal{G}_c$ and a noisy graph dataset $\mathcal{G}_n$\; \For{$iter \leftarrow iter_{1}, iter_{2}, \ldots, iter_{max}$}{Sample graph data $\left\{G_i\right\}_{i=1}^B$ with batch size $B$\; \If{$G_i \in \mathcal{G}_{n}$}{ Label correction of noisy graph data using Algorithm 2;} Update the parameters of the network according to the objective function in Eq. \ref{eq21};} }
\label{algo3}
\end{algorithm}
\subsection{Model training}
In deep learning with noisy labels, there is a phenomenon where the model initially fits clean labels at the early stages of training and gradually fits noisy labels later on. Moreover, to ensure the model has some discriminative capability, we perform a few training rounds on the model in batches as a warm-up before the official training begins. Then, we use Algorithm 1 to filter out noise from the original graph dataset with noisy labels $\mathcal{G}$, thereby obtaining a clean graph dataset $\mathcal{G}_c$ and a noisy graph dataset $\mathcal{G}_n$. For a batch of graph data $B$, data belonging to the clean graph dataset $\mathcal{G}_c$ are optimized directly using the cross-entropy loss function. In contrast, data from the noisy graph dataset $\mathcal{G}_n$ undergo label correction using Algorithm 2, obtaining the corrected graph dataset $\mathcal{G}_r$, then optimized using cross-entropy loss. Furthermore, for the graph dataset $\mathcal{G}_{new}$ with corrected labels, we construct a supervised graph contrastive loss using Eq. \ref{eq20}, introducing label information to enforce representation constraints in the embedding space. The overall objective function $L$ of the model can ultimately be expressed as follows:

\begin{equation}\begin{split}
L=\sum_{G_i \in \mathcal{G}_c} \ell_{C E}\left(G_i, f_{\theta_E}, f_{\theta_C}\right)+\alpha \sum_{G_i \in \mathcal{G}_r} \ell_{C E}\left(G_i, f_{\theta_E}, f_{\theta_E}^{\prime}, f_{\theta_C}\right)\\+\beta \sum_{G_i \in \mathcal{G}_{new}} \ell_{S C L}\left(G_i, f_{\theta_E}, f_{\theta_E}^{\prime}, f_{\theta_P}\right)
\label{eq21}
\end{split}\end{equation}
where $\ell_{CE}$ and $\ell_{SCL}$ denote the cross-entropy loss and the contrast loss, respectively, and $\alpha $ and $\beta$ are the loss-constrained hyperparameters. Algorithm 3 describes the training process of the RGLC method.

\section{Experimental Results}
\label{se5}
\subsection{Datasets}
This paper uses the same eight graph classification datasets as OMG \citep{yin2023omg} to verify the utility and privacy of the RGLC method, including MUTAG \citep{debnath1991structure}, NCI1 \citep{wale2008comparison}, PTC \citep{kriege2012subgraph}, COX2 \citep{long2021theoretically}, PROTEINS \citep{borgwardt2005protein}, PROTEINS\_F \citep{borgwardt2005protein}, IMDB-B and IMDB-M \citep{yanardag2015deep}. Table 1 summarizes the statistical data of each dataset. Detailed data are described as follows: 

\textit{Bioinformatics graph datasets:} MUTAG contains 188 compounds, labeled according to whether they have a mutagenic effect on bacteria, in which the node features are one-hot encodings of atom types; NIC1 contains 4100 compounds, labeled according to whether they have a mutagenic effect on cancer cells. The characteristics of growth are data-tagged; PTC contains 344 organic molecules, which are data-tagged according to their carcinogenicity to rodents, where the node features are one-hot encodings of atomic types; COX2 is composed of 467 inhibitors, according to whether they have in vitro Data labeling for anti-human recombinant protein activity; PROTEINS and PROTEINS\_F contain two data sets of 1113 proteins labeled as enzymatic or non-enzymatic, and their node features are one-hot encoding of amino acid types.

\textit{Social network graph datasets:} IMDB-B and IMDB-M are movie collaboration graph data sets labeled according to the type of self-network, in which node features are one-hot encoding of node degrees.

More detailed information and downloads of the above graph data sets can be obtained through this address \footnote{https://chrsmrrs.github.io/datasets/docs/datasets/}. Since the original labels of all graph datasets are clean, we randomly flip the labels in the training dataset to other categories with $p \%$ noise rate according to the label corruption method in OMG.

\begin{table*}[htbp]
  \centering
  \caption{All methods were tested for graph classification accuracy $(\%)$ $(Acc \pm 
 Std)$ at $30 \%$ noise label rate, with best performance using bold font.}
    \begin{tabular}{ccccccccc}
    \toprule
    \textbf{Methods} & \textbf{MUTAG} & \textbf{PTC} & \textbf{COX2} & \textbf{NCI1} & \textbf{IMDB-B} & \textbf{IMDB-M} & \textbf{PROTEINS} & \textbf{PROTEINS\_F} \\
    \midrule
    WL subtree & 73.4±9.3 & 55.3±10.3 & 65.1±8.6 & 64.5±4.3 &  64.5±4.1 & 46.7±4.5 & 65.1±5.9 & 66.0±7.7 \\
    LDP   & 62.2±6.7  & 45.4±16.7 & 59.3±11.1 & 55.8±2.4 & 53.1±8.2 & 36.8±5.0 & 56.4±6.8 &  54.8±5.9 \\
    GIN   & 73.2±10.4 & 56.8±7.4 & 76.2±6.7 & 66.4±3.4 & 64.3±3.7 & 46.7±3.3 & 68.1±6.3 & 70.1±5.0 \\
    GraphCL & 74.2±9.7 &  57.4±7.8 & 75.8±5.4 & 66.8±4.2 & 62.8±5.1 & 43.8±3.8 & 68.7±6.1 & 71.0±4.6 \\
    SimGRACE & 75.5±8.1 & 58.8±8.0 & 76.2±4.9 & 67.5±4.4 & 63.3±5.5 & 42.7±2.3 & 68.9±7.2 & 71.7±4.6 \\
    Co-teaching & 59.0±14.6 &  55.8±6.7 & 58.7±11.1 & 52.3±5.2 & 57.2±4.0 & 43.8±4.8 & 40.2±4.1 &  68.2±8.2 \\
    DivideMix & 65.4±10.9 & 56.5±6.3 & 61.3±8.6 & 59.7±4.9 & 61.2±5.3 & 44.2±3.9 & 51.5±5.1 & 67.3±9.9 \\
    Taylor-CE & 72.1±11.3 & 55.8±8.2 & 75.4±7.8 &  65.9±3.2 & 62.6±4.7 & 47.9±3.9 & 67.5±6.9 & 69.1±5.4 \\
    CDR   & 73.2±12.6 &  54.9±7.9 & 73.4±7.4 & 65.7±2.6 & 64.3±3.4 & 46.7±4.1 & 69.8±3.6 & 64.2±3.7 \\
    RTGNN & 73.8±10.9 & 58.1±6.8 & 76.5±6.6 & 67.7±2.9 & 65.1±4.2 & 46.5±5.2 & 69.6±3.5 & 70.2±3.9 \\
    Sel-CL & 77.0±8.8 & 60.8±6.9 & 77.4±6.1 & 68.3±3.6 & 67.2±4.6 & 46.9±3.9 & 69.9±4.3 & 72.5±4.1 \\
    OMG   & 77.7±9.5  &  61.7±8.6 & 78.8±5.7 & 65.1±3.1  & 67.4±5.5 & 45.1±3.9 & 70.8±4.0 & 72.9±3.8 \\
    \midrule
    RGLC  & \textbf{81.6±7.1} & \textbf{63.0±5.7} & \textbf{79.6±3.5} & \textbf{72.9±3.3} & \textbf{69.4±3.6} & \textbf{49.6±4.1} & \textbf{72.3±3.2} & \textbf{74.2±3.5} \\
    \bottomrule
    \end{tabular}%
  \label{tab:addlabel}%
\end{table*}%
\subsection{Comparing Methods}
We used twelve existing comparison methods to verify the effectiveness of the proposed RGLC method, including graph classification methods: WL subtree \citep{DBLP:journals/jmlr/ShervashidzeSLMB11}, LDP \citep{cai2018simple} and GIN \citep{DBLP:conf/iclr/XuHLJ19}; Graph Contrastive learning methods: GraphCL \citep{you2020graph} and SimGRACE \citep{xia2022simgrace}; Deep learning methods with noise label learning: Co-teaching \citep{han2018co}, DivideMix \citep{DBLP:conf/iclr/LiSH20}, Taylor-CE \citep{feng2021can}, CDR \citep{xia2021robust} and Sel-CL \citep{li2022selective}; Graph neural network methods with noise label learning: RTGNN \citep{qian2023robust} and OMG \citep{yin2023omg}. The specific methods are explained as follows:\\\textbf{Graph classification methods}
 \begin{itemize}
  \item \textit{WL subtree:} The method is based on the Weisfeiler-Lehman test of graph isomorphism for effective feature extraction, and the similarity of the output graphs is used to determine the categories to which they belong.
  \item \textit{LDP:} It computes the degree of each node and its neighborhood and then measures its empirical distribution in the graph.
  \item \textit{GIN:} It effectively uses a message-passing mechanism to learn the feature representation of the nodes in the graph to achieve classification.
\end{itemize}
\textbf{Graph Contrastive learning methods:}
 \begin{itemize}
  \item \textit{GraphCL:} It is a method for unsupervised graph representation learning based on graph enhancement methods such as edge perturbation, node discarding, and attribute masking.
  \item \textit{SimGRACE:} Instead of using conventional graph enhancement methods, the method performs noise perturbation for the encoder's parameters to obtain the supervised signals needed for graph contrastive learning.
\end{itemize}
\textbf{Deep learning methods with noise label learning:}
 \begin{itemize}
  \item \textit{Co-teaching:} It is a method of co-training using two-branch networks, where the training samples of each network come from clean samples screened by the other network based on a small loss criterion for training, and this method requires that the noise label rate is known in advance.
  \item \textit{DivideMix:} The method continues the way of two-branch network co-training, using the small loss criterion to screen out clean samples and re-labeling clean samples, and for noise samples using data enhancement to re-assign pseudo-labels and data enhancement with MixMatch to achieve robust learning. In our experiments, we use graph enhancement methods with subgraph sampling and edge discarding, as well as Mixup operations in the embedding space to accommodate the specificity of the graph data.
  \item \textit{Taylor-CE:} It is a generalized framework for training deep models in the presence of label noise that allows weighting the degree of matching of training labels.
  \item \textit{CDR:} It uses an early stopping technique to reduce the overfitting of the model for noisy labels and make it have better generalization performance.
  \item \textit{Sel-CL:} This method proposes nearest neighbor correction under unsupervised and selective supervised contrastive learning to enhance the model's robustness to noisy labels.
\end{itemize}
\textbf{Graph neural network methods with noise label learning:}
\begin{itemize}
  \item \textit{RTGNN:} This method achieves noise robustness under the node classification task by dynamically selecting confident samples in a two-branch network and introducing self-reinforcement and consistent regularization to supervise and supplement. In this paper, we remove the graph enhancement technique of this method and apply the remaining method to the graph classification task for comparative experiments.
  \item \textit{OMG:} In this method, Mixup enhancement technology is integrated into supervised graph contrastive learning guided by soft labels to learn anti-noise graph representation, and the domain information of samples is used to remove label noise.
\end{itemize}
\subsection{Experimental Setting and Environment}
As with the setup in OMG, we adopt the network structure setup of the encoder and classifier of GIN in \citep{DBLP:conf/iclr/XuHLJ19}, where the encoder is composed of five graph convolutional layers (including the input layer) and one pooling layer, the classifier uses two layers of MLPs, the projection header required in supervised graph contrastive is also used with two layers of MLPs, the embedding dimensions of the hidden layers of all the methods are set to be 64, batch size between $\{32,64,128\}$. The noisy label screening thresholds $\tau_1$ were set to 0.7 and $\tau_2$ to 0.5. The noise perturbation coefficient $\eta$ was taken to be 1.0, and $\alpha$ and $\beta$ were set to be in the $\{0.1,0.2,0.3,0.4,0.5,0.6,0.7,0.8,0.9,1.0\}$ and $\{0.001,0.01,0.1,1.0,10 \}$ in the range of warm-up rounds $T_l$ in $\{10,15,20\}$. The model uses the Adam optimizer with an initial learning rate of 0.01 and a weight decay factor of $5e-4$. We follow the dataset partitioning approach in \citep{DBLP:conf/iclr/XuHLJ19} and report the mean and standard deviation of the ten-fold results. All experiments were performed on an NVIDIA Tesla V100-SXM2-32GB chip with 32GB memory.
\begin{figure*}[!h]
    \centering
    \begin{subfigure}{0.325\textwidth} % 调整子图的宽度
        \centering
        \includegraphics[width=\textwidth]{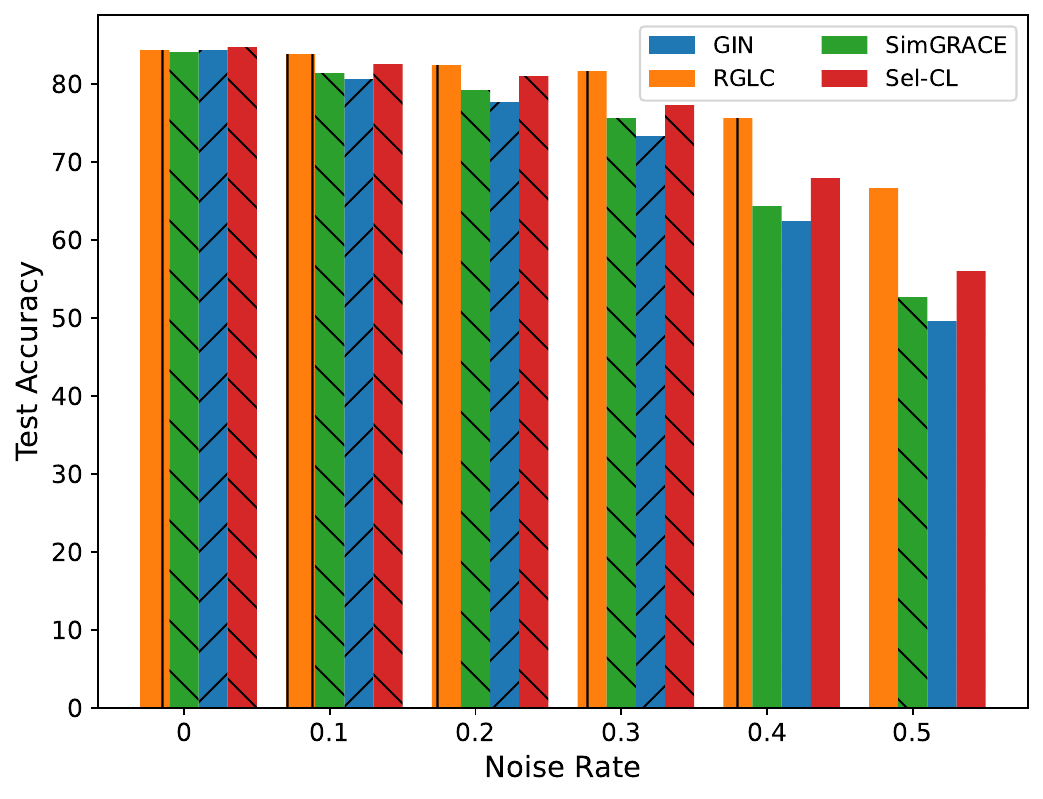}
        \caption{MUTAG}
        \label{fig:sub1}
    \end{subfigure}
    \hspace{0.00\textwidth} % 添加水平间距
    \begin{subfigure}{0.325\textwidth}
        \centering
        \includegraphics[width=\textwidth]{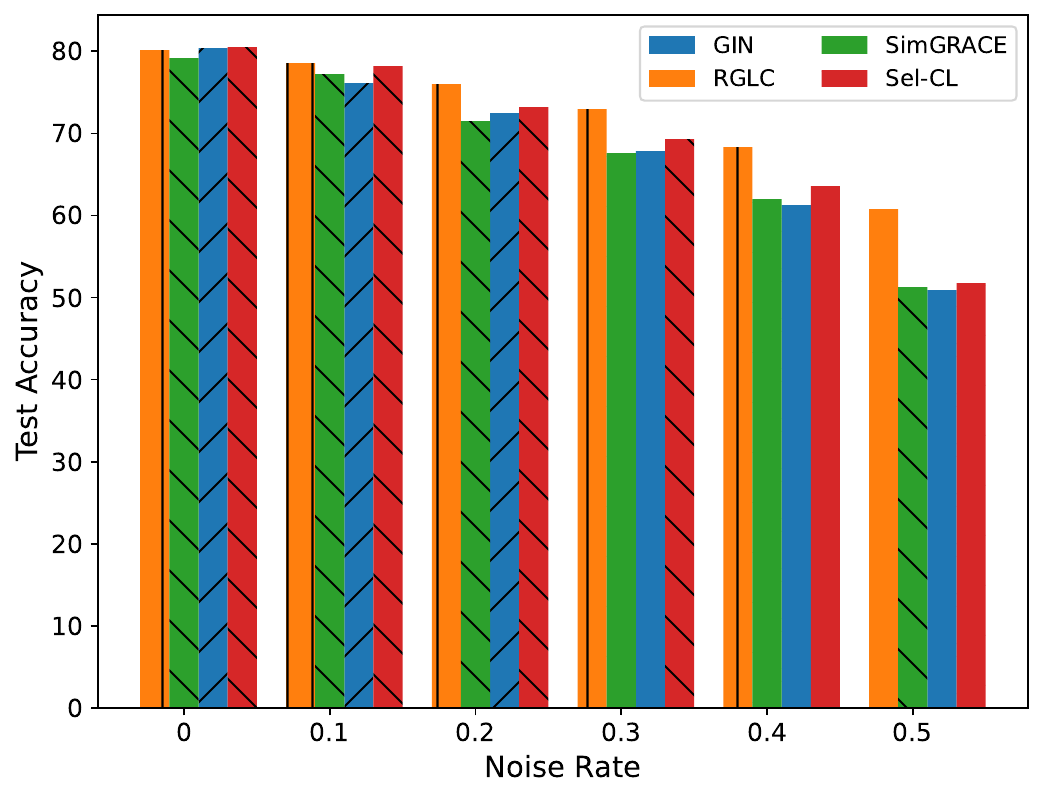}
        \caption{NCI1}
        \label{fig:sub2}
    \end{subfigure}
    \hspace{0.00\textwidth} % 添加水平间距
    \begin{subfigure}{0.325\textwidth}
        \centering
        \includegraphics[width=\textwidth]{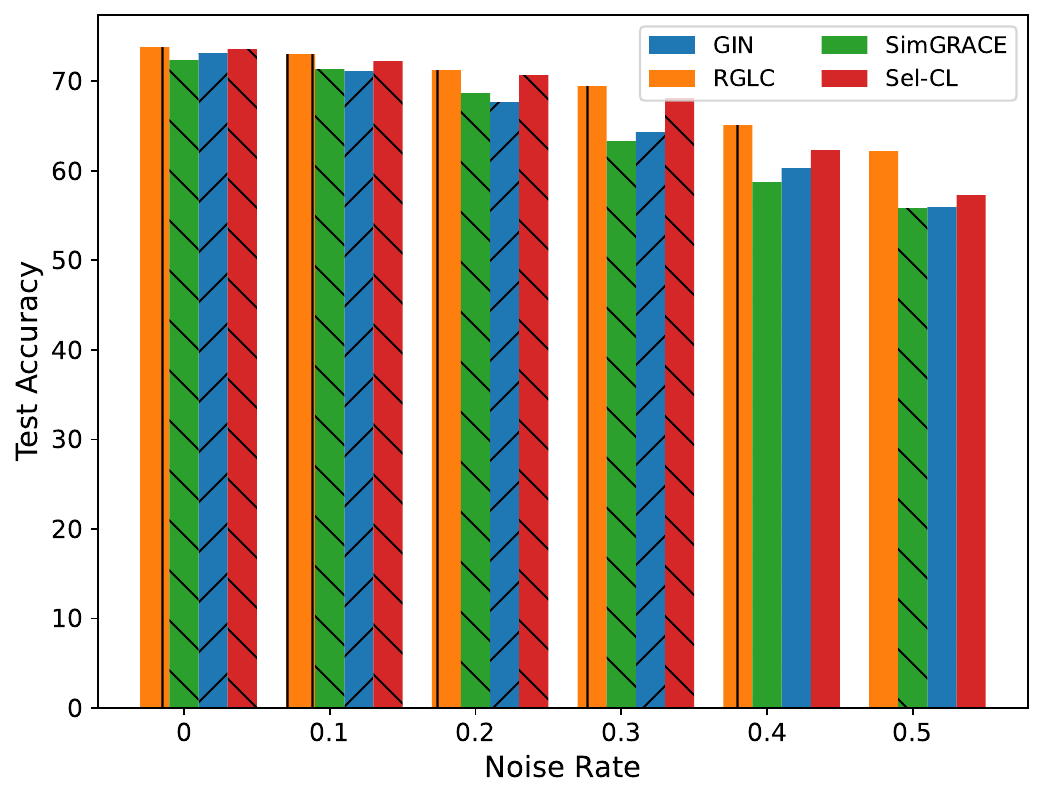}
        \caption{IMDB-B}
        \label{fig:sub4}
    \end{subfigure}
    \caption{Test performance of GIN, SimGRACE, and our RGLC method under different noise label rates for multiple datasets.}
    \label{fig:main_figure}
\end{figure*}
\subsection{Graph classification with noise labels}
We fixed the noise label rate to $30 \%$ for the method comparison, and the experimental results are shown in Table 2, where some of the data are from OMG. From the results in the table, the following findings are made: the three graph classification methods, WL subtree, LDP, and GIN, despite their noise immunity on individual graph datasets, still have poor generalization ability as a whole due to the lack of a corresponding anti-noise mechanism. Unsupervised comparative learning GraphCL and SimGRACE mitigate the effect of noisy labels by introducing robust graph representations, but due to the lack of targeted processing downstream with noisy labels, they do not improve the performance too much compared to the supervised GIN model, and the GraphCL method achieves a test accuracy of $2.9 \%$ lower than that of GIN on the IMDB-M dataset. This is because the graph augmentation approach adopted by GraphCL tends to lead to a lack of consistency of graph semantic information in noisy labeled scenarios, which affects its generalization performance. Compared to the noise-robust methods, our RGLC method achieves optimal test accuracy on all datasets. The performance of both the two-branch network-based Co-teaching and DivideMix methods is lower than that of the GIN method, which indicates that small-loss-based co-training does not filter enough clean samples for graph classification, resulting in the accumulation of the training error, ultimately leading to a poorer generalization performance. For RTGNN and Sel-CL methods, although compared to baselines, some robustness can be improved. However, due to the different data patterns, they can not achieve better generalization performance. Compared with OMG, a graph classification method with noisy labels, our RGLC method achieves the best performance on all datasets, with at most and at least $7.8 \%$ and $0.8 \%$ accuracy gains over OMG, respectively. Our method has a small standard deviation, which suggests that our noisy sample selection method and the label correction mechanism can effectively filter and correct noisy data, thus improving the ability of graph neural networks to resist noisy labels under the graph classification task.
\subsection{Impact at different noise rates}
In order to evaluate the performance of the RGLC method under different noise label rates, we compared the three methods GIN, SimGRACE and Sel-CL on the noise label rates of $\{0,0.1,0.2,0.3,0.4,0.5\}$. From the experimental results in Fig. 5, we can obtain the following information: Although the test performance of the four methods will decrease as the noise label rate increases, our RGLC method can still achieve the best performance, especially when the noise rate is small, It can achieve performance close to the clean label case. In the case of clean labels, our method will hardly reduce the test performance of the model, indicating that our noise sample screening mechanism can accurately calculate the alignment degree of the sample with the first principal component of the category, achieve the separation of clean and noisy samples, and make the model perform better in Noise-free scenes have better generalization performance. At the same time, we can also see that the SimGRACE method can achieve greater performance gains in smaller graph data sets (e.g., MUTAG), while in larger data sets (e.g., NCI1) the performance gains are smaller or even negative. Baseline GIN model. Finally, we find that Sel-CL can achieve performance comparable to our method when the noise rate is small. As the noise rate increases, the Sel-CL method gradually becomes worse due to the increase in erroneous neighbor information.

\subsection{Privacy Protection Evaluation}
\begin{figure*}
    \centering
    \begin{subfigure}{0.33\textwidth} % 调整子图的宽度
        \centering
        \includegraphics[width=\textwidth]{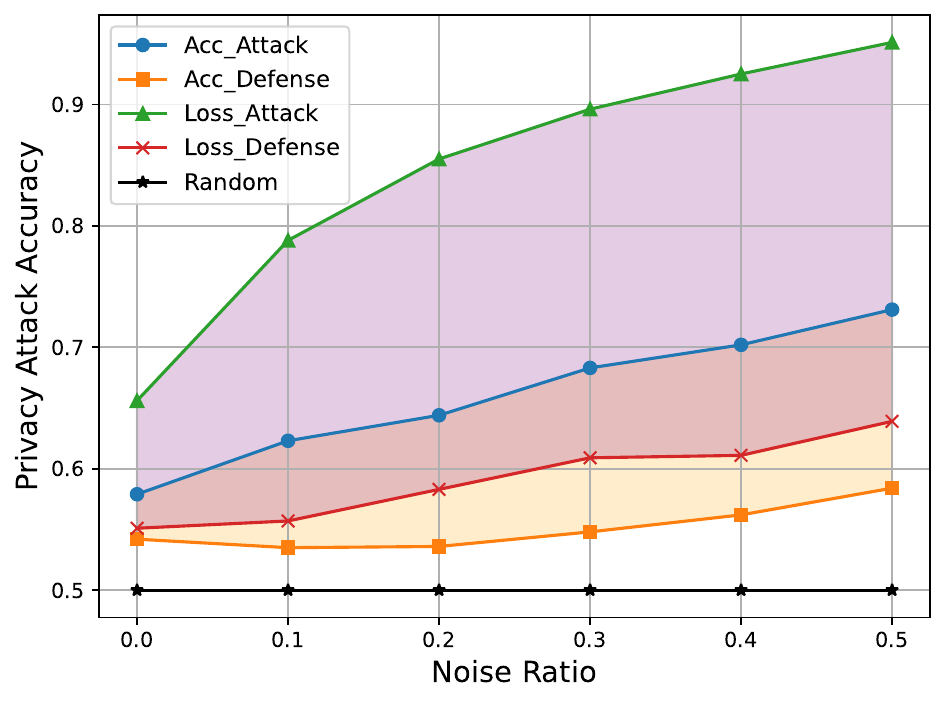}
        \caption{MUTAG}
        \label{fig:sub1}
    \end{subfigure}
    \hspace{0.00\textwidth} % 添加水平间距
    \begin{subfigure}{0.33\textwidth}
        \centering
        \includegraphics[width=\textwidth]{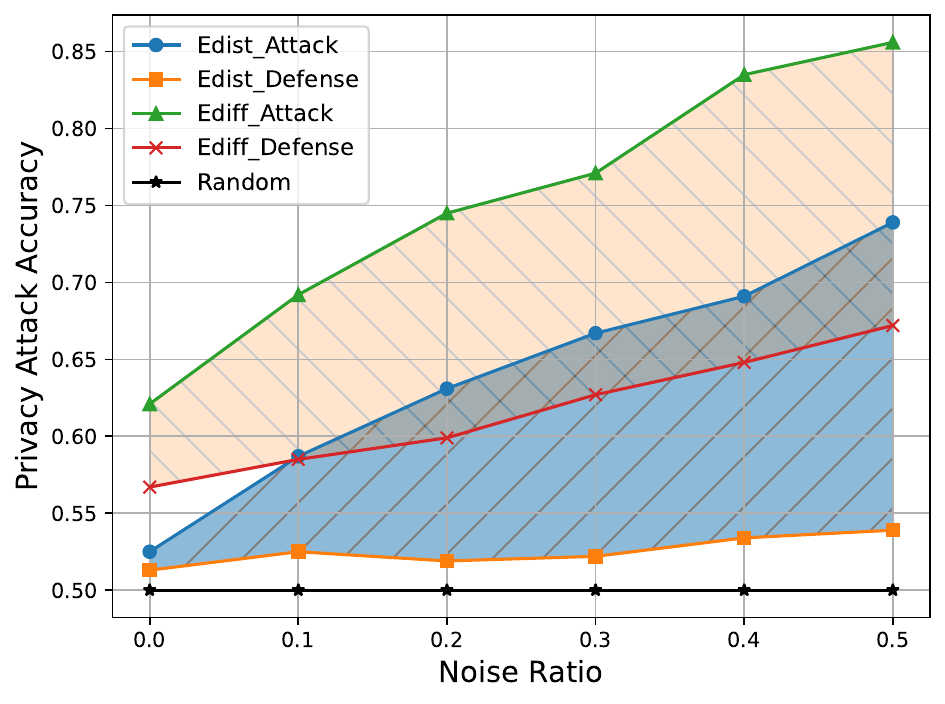}
        \caption{NCI1}
        \label{fig:sub2}
    \end{subfigure}
    \hspace{0.00\textwidth} % 添加水平间距
    \begin{subfigure}{0.31\textwidth}
        \centering
        \includegraphics[width=\textwidth]{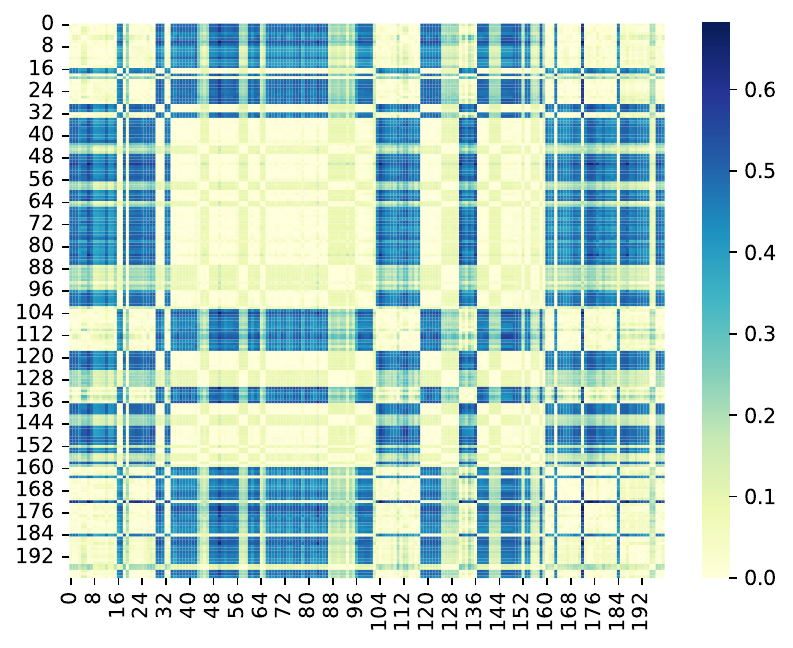}
        \caption{NCI1}
        \label{fig:sub2}
    \end{subfigure}
    \caption{(a)-(b): Validation of our RGLC method on MUTAG and NCI1 datasets to defend against membership inference attacks in learning scenarios with noisy labels; (c):Visualization of the difference matrix of the cosine similarity matrix of the graph embedding.}
    \label{fig:main_figure}
\end{figure*}
We validate the ability of the RGLC method to defend against graph or subgraph membership inference attacks at different noise label rates on the MUTAG and NCI1 datasets, and the corresponding experimental results are illustrated in Fig. 6, where Random denotes a random guessing attack with the default value of 0.5. From the figure, we can see that our RGLC method effectively reduces the membership inference attack under all noise label rates in terms of its ability, especially on the MUTAG dataset, to minimize the cross-entropy loss attack based on model prediction, indicating that our method mitigates the overfitting problem caused by noisy labels, indents the loss gap between training and testing, and thus improves the ability to defend against graph membership inference attacks in black-box scenarios. For the subgraph membership inference attack based on embedding information in the white-box scenario in Fig. 6 (b), our method still reduces its attacking capability. However, it cannot defend Ediff\_Attack well when the noise label rate is high because the attacker has strong attacking knowledge in the scenario of white-box attacking. The Ediff\_Attack utilizes the embedding difference of the corresponding dimension position as the attack input, which provides rich query information for the attack and a noise label rate of 0.5. However, our method improves the ability to fight against the noise, but due to the influence of the noise label still exists, it will still lead to a large discrepancy between the graph embedding of the training data and that of the test data, which will make the scenario of the Ediff\_ Attack difficult to defend. Fig.6 (c) shows the feature similarity of the graph embeddings of the test graph dataset extracted by the encoder trained without noise labels and the encoder trained with noise labels using the cosine similarity at a noise label rate of 0.5, respectively, and performing the subtraction by positional elements to obtain the difference matrix. Ideally, this difference matrix should be almost close to the similarity of 0 value. It can be seen from the results that because the encoder is interfered by noise labels, the generalization of embedding of different categories of test graph data is not good enough, resulting in a certain deviation from the embedding of original graph data. Therefore, it is still affected by Ediff\_Attack. \cite{zhang2022inference} proposes that graph embeddings be noised to protect against privacy leakage, but the addition of noise affects the model's utility and increases the complexity of model training. Therefore, improving the privacy protection capabilities of the model under high noise label rates will be a research point that needs to be explored in the future.
\subsection{Ablation experiments}

\begin{table*}[htbp]
  \centering
  \caption{Quantification of the test accuracy ($\%$) of each sub-module of the RGLC method at a noise label of 30$\%$.}
    \resizebox{\textwidth}{!}{%
    \begin{tabular}{ccccccccc}
    \toprule
    \multicolumn{1}{c|}{\textbf{Methods}} & \textbf{MUTAG} & \textbf{PTC} & \textbf{COX2} & \textbf{NCI1} & \textbf{IMDB-B} & \textbf{IMDB-M} & \textbf{PROTEINS} & \textbf{PROTEINS\_F} \\
    \midrule
    \multicolumn{1}{c|}{GIN} & 73.2±10.4 & 56.8±7.4 & 76.2±6.7 & 66.4±3.4 & 64.3±3.7 & 46.7±3.3 & 68.1±6.3 & 70.1±5.0 \\
    \multicolumn{1}{c|}{OMG} & 77.7±9.5  &  61.7±8.6 & 78.8±5.7 & 65.1±3.1  & 67.4±5.5 & 45.1±3.9 & 70.8±4.0 & 72.9±3.8 \\
    \midrule
    \midrule
    RGLC W/O S1 & 76.1±8.3 & 59.2±5.1 & 76.6±5.8 & 68.6±3.7 & 65.3±5.5 & 46.9±2.6 & 70.6±4.9 & 70.6±5.4 \\
    RGLC W/O S2 & 70.4±11.5 & 53.9±7.6 & 72.4±6.7 & 65.5±5.9 & 61.7±6.1 & 41.2±5.5 & 65.7±4.3 & 67.2±5.5 \\
    RGLC W/O C & 75.2±7.7 & 59.1±5.6 & 75.2±4.9 & 68.4±5.0 & 65.1±3.5 & 47.7±2.5 & 70.2±3.9 & 71.7±4.4 \\
    RGLC W/O C1 & 78.5±7.9 & 59.6±6.6 & 77.1±5.8 & 69.7±3.8 & 66.8±4.5 & 47.9±3.4 & 69.5±3.6 & 72.0±4.7 \\
    RGLC W/O C2 & 79.9±6.5 & 60.4±4.0 & 77.7±4.9 & 70.8±4.1 & 67.7±4.3 & 48.5±2.8 & 71.6±2.8 & 72.9±3.8 \\
    RGLC W/O SCL & 80.3±8.2 & 62.1±5.9 & 78.3±4.2 & 72.0±3.1 & 68.9±3.2 & 49.5±2.4 & 71.4±3.1 & 73.7±3.4 \\
    \midrule
    \rowcolor[rgb]{ .851,  .851,  .851} RGLC  & \textbf{81.6±7.1} & \textbf{63.0±5.7} & \textbf{79.6±3.5} & \textbf{72.9±3.3} & \textbf{69.4±3.6} & \textbf{49.6±2.1} & \textbf{72.3±3.2} & \textbf{74.2±3.5} \\
    \bottomrule
    \end{tabular}%
  \label{tab:addlabel}%
  }
\end{table*}%
In order to quantify the importance of different sub-modules in the RGLC method, we conducted ablation experiments on eight benchmark graph datasets with the noise label rate set to 30$\%$, and the results of the experiments are shown in Table 3. First, we give a brief introduction of each sub-module: \textbf{\textit{RGLC W/O S1}} denotes that the small-loss preparation is not used to screen high confidence samples for the computation of the category feature principal components in the training process; \textbf{\textit{RGLC W/O S2}} denotes that the small loss criterion is directly used to screen clean versus noisy graph data during training, discarding the mechanism of screening noisy samples using the category principal component vectors on the feature space; \textbf{\textit{RGLC W/O C}} denotes that the label correction mechanism of the two-space perspective is not used, and only the screened clean samples are utilized for model training; \textbf{\textit{RGLC W/O C1}} denotes that in the noisy label correction mechanism, the noise is discarded in the feature space label correction; \textbf{\textit{RGLC W/O C2}} denotes discarding the supervised information in the output space label correction in the noisy label correction mechanism; and \textbf{\textit{RGLC W/O SCL}} denotes not employing supervised graph contrastive learning to enhance the quality of graph representation. The specific results are analyzed as follows:
 \begin{itemize}
  \item In the noisy sample screening part, the RGLC method proposed in this paper outperforms RGLC W/O S1 and RGLC W/O S2 on all the datasets tested, and in particular, if the RGLC W/O S2 sub-module is discarded, then our method will achieve the worst generalization performance, showing that the degree of alignment of graph embeddings in the embedding space using the principal components of the category features can be a good way to separate the clean and noisy graph data and improve the performance of the model. At the same time, without high-confidence sample screening, the feature principal components of each category may be biased by noise, leading to a decrease in the model's noise immunity, and this effect may be exacerbated as the noise rate increases.
  \item In the noise label correction part, if the label correction mechanism is directly discarded and the information of the noisy samples is not utilized, the performance of RGLC W/O C on some datasets will not be much improved compared to the GIN model or even the performance is worse. For example, on the COX2 dataset, the performance of RGLC W/O C is still 1$\%$ lower than that of GIN, indicating that the model must obtain sufficient clean samples. In addition, the RGLC method achieves better performance than both label correction mechanisms in all cases, suggesting that both label correction based on the embedding space and label correction in the output space improve the model's resistance to noise and that the label correction mechanism based on the principal component vectors of the category features in the embedding space provides more stable label information in most cases.
  \item  Finally, we can observe that RGLC W/O SCL achieves slightly lower performance than the RGLC method on all the datasets, indicating that graph contrastive learning constructed using label information on top of space label correction helps the model to learn more robust graph representations, which in turn improves the model's ability to screen and correct for noisy labels.
\end{itemize}

\begin{figure*}
    \centering
    \begin{subfigure}{0.24\textwidth} % 调整子图的宽度
        \centering
        \includegraphics[width=\textwidth]{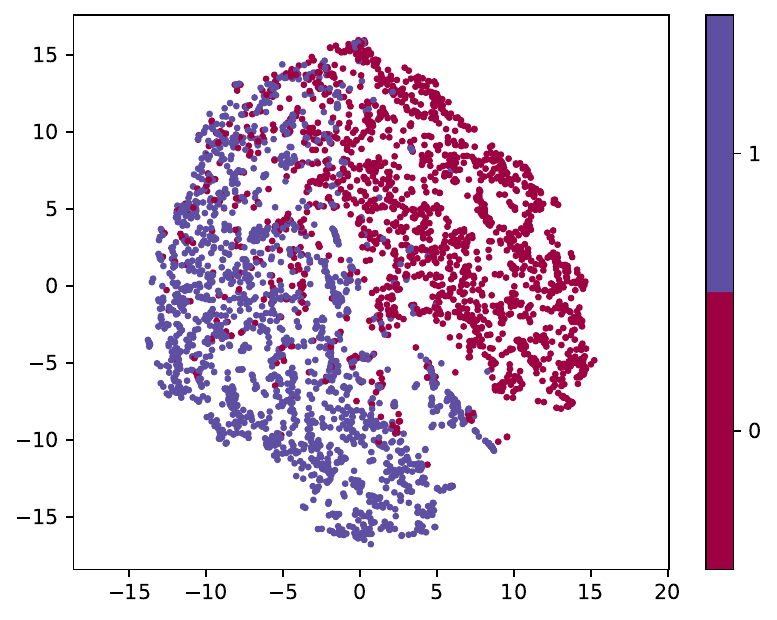}
        \caption{RGLC-E}
        \label{fig:sub1}
    \end{subfigure}
    \hspace{0.00\textwidth} % 添加水平间距
    \begin{subfigure}{0.24\textwidth}
        \centering
        \includegraphics[width=\textwidth]{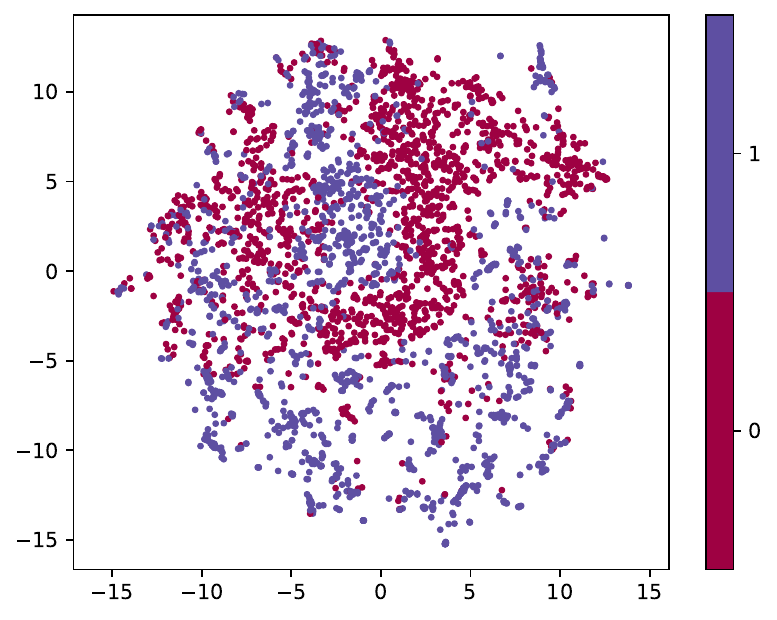}
        \caption{DivideMix-E}
        \label{fig:sub2}
    \end{subfigure}
    \hspace{0.00\textwidth} % 添加水平间距
    \begin{subfigure}{0.24\textwidth}
        \centering
        \includegraphics[width=\textwidth]{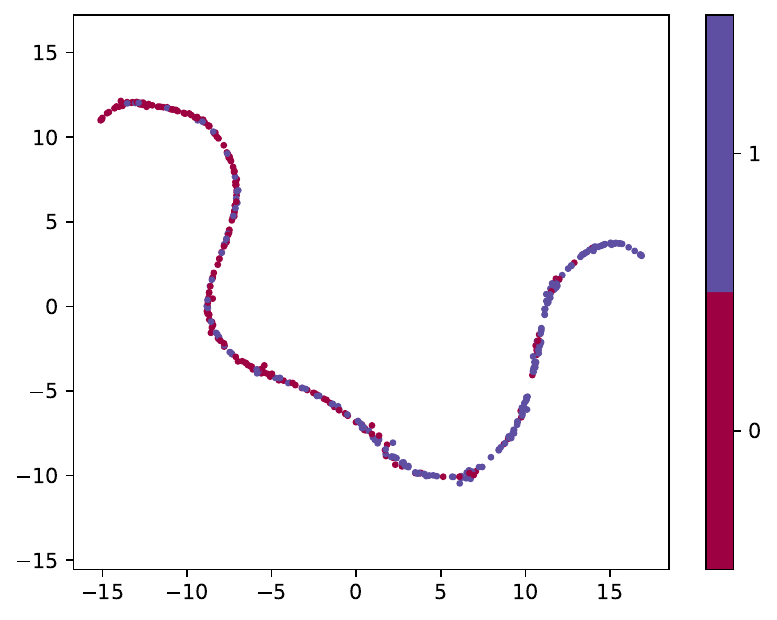}
        \caption{RGLC-O}
        \label{fig:sub4}
    \end{subfigure}
    \hspace{0.00\textwidth} % 添加水平间距
    \begin{subfigure}{0.24\textwidth}
        \centering
        \includegraphics[width=\textwidth]{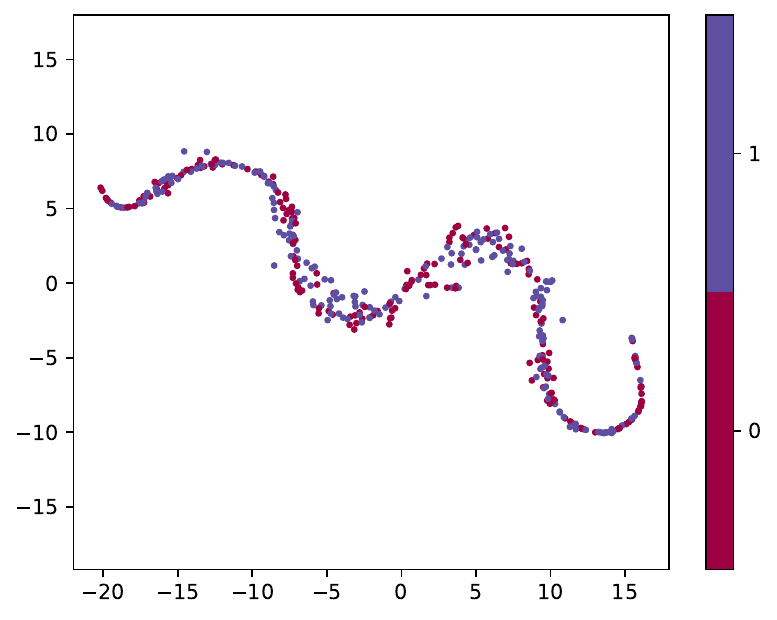}
        \caption{DivideMix-O}
        \label{fig:sub4}
    \end{subfigure}
    \caption{Visualize the graph embedding and prediction distributions of the RGLC and DivideMix methods on the NCI1 dataset with 30$\%$ noise label rate, where RGLC-E and RGLC-O denote the embedding distributions of the RGLC method on the training graph dataset and the prediction distributions on the test graph dataset, respectively.}
    \label{fig:main_figure}
\end{figure*}
\subsection{Visualization and generalizability analyses}
To visualize the effectiveness of our RGLC method, we visualize the graph embedding of the training set as well as the prediction distributions of the test set on the NCI1 dataset with 30$\%$ noisy label rate using the t-SNE \citep{van2008visualizing} technique and the visualization of the graph embedding is shown in Fig. 7(a)-(b), which shows that, compared with the DivideMix method, our RGLC method can embed the spatial obtain clear intra-class consistency and inter-class variability, indicating that supervised graph contrastive learning with the introduction of category information can further improve the graph encoder's capability with the aid of label correction. The visualization of the prediction distributions is shown in Fig. 7(c)-(d), from which it can be seen that our method achieves better discriminative power of category information, indicating that we accurately screen out the noisy samples in the embedding space and utilize the label correction in the dual-space can make the model obtain practical training supervised information in the noise-carrying learning.

In addition, we also visualize the training and testing loss variation trends of our RGLC method and GIN method under the setting of 30$\%$ noisy labels by visualizing our RGLC method and GIN method on COX2 and PTC graph datasets, and it is observable through the experimental results in Fig. 8, that our method can achieve minor loss differentiation, which leads to excellent generalization performance in the testing phase.

\begin{figure*}[!h]
    \centering
    \begin{subfigure}{0.24\textwidth} % 调整子图的宽度
        \centering
        \includegraphics[width=\textwidth]{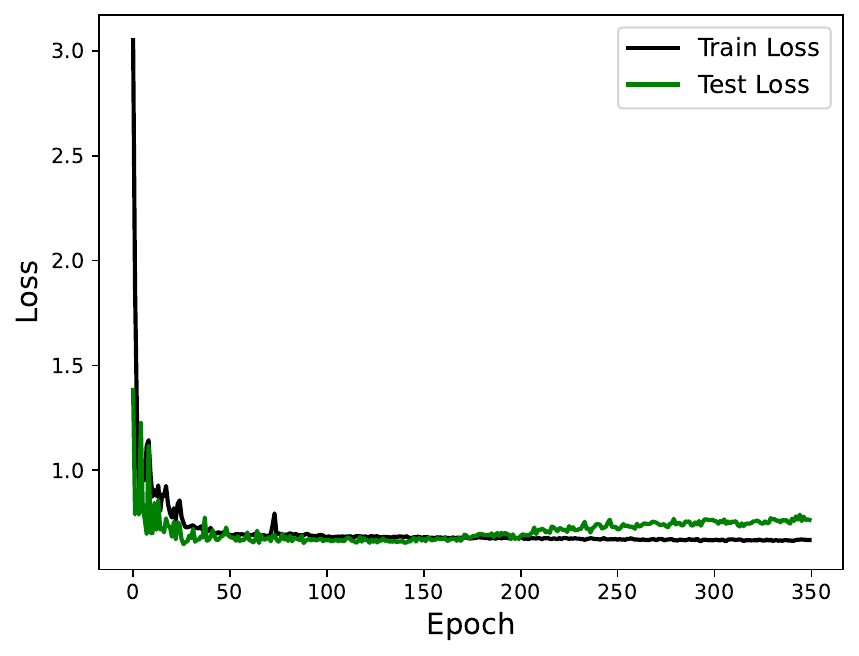}
        \caption{RGLC on COX2}
        \label{fig:sub1}
    \end{subfigure}
    \hspace{0.00\textwidth} % 添加水平间距
    \begin{subfigure}{0.24\textwidth}
        \centering
        \includegraphics[width=\textwidth]{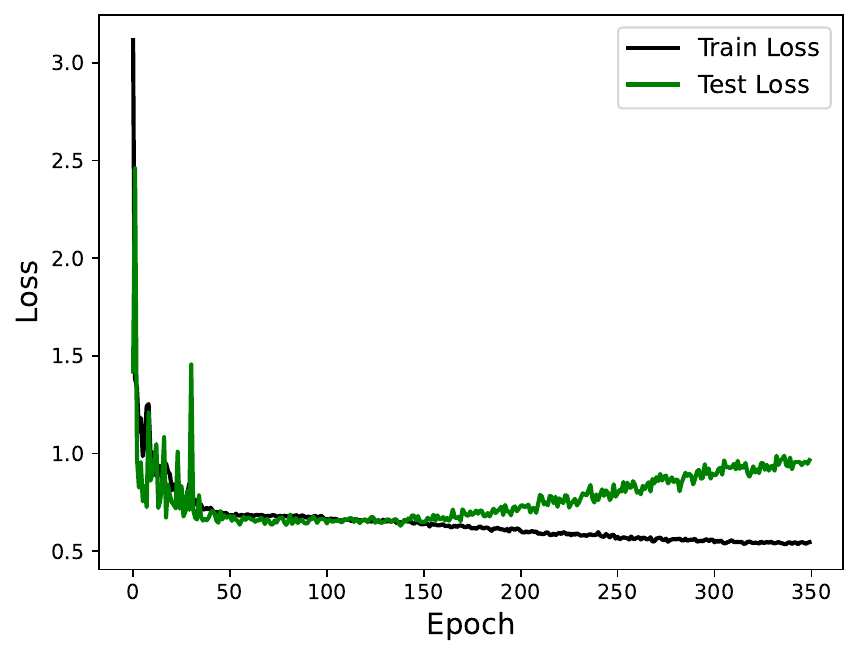}
        \caption{GIN on COX2}
        \label{fig:sub2}
    \end{subfigure}
    \hspace{0.00\textwidth} % 添加水平间距
    \begin{subfigure}{0.24\textwidth}
        \centering
        \includegraphics[width=\textwidth]{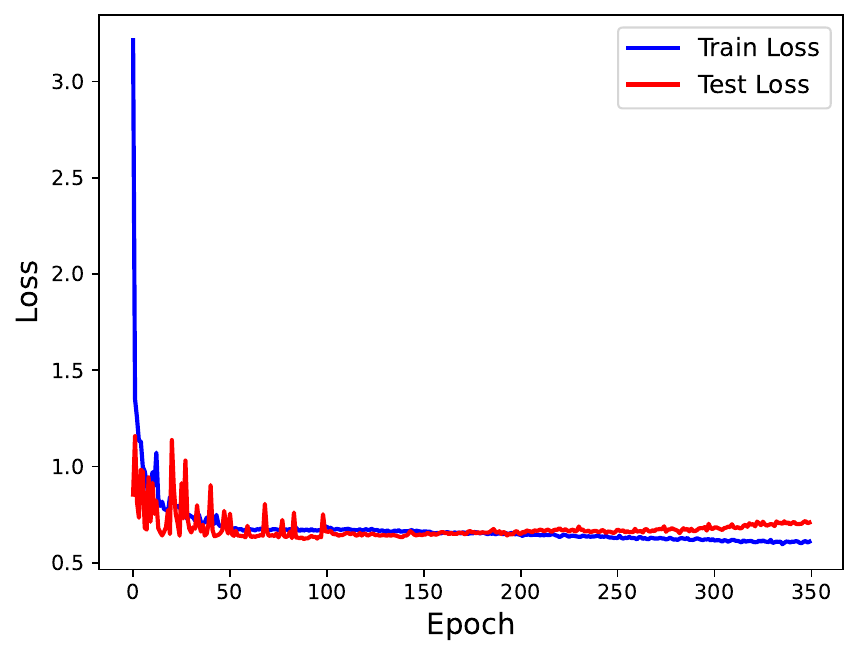}
        \caption{RGLC on PTC}
        \label{fig:sub4}
    \end{subfigure}
    \hspace{0.00\textwidth} % 添加水平间距
    \begin{subfigure}{0.24\textwidth}
        \centering
        \includegraphics[width=\textwidth]{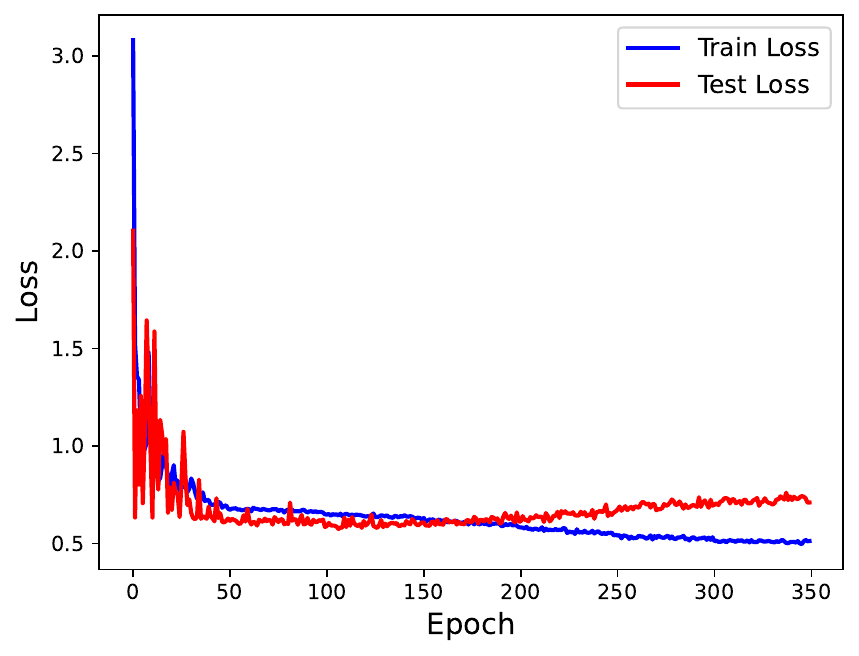}
        \caption{GIN on PTC}
        \label{fig:sub4}
    \end{subfigure}
    \caption{Loss value variation of RGLC and GIN methods on COX2 dataset and PTC dataset with 30$\%$ noise label rate.}
    \label{fig:main_figure}
\end{figure*}

% Table generated by Excel2LaTeX from sheet 'Sheet2'
\begin{table*}[!h]
  \centering
  \caption{Time for the model to train at each epoch (seconds).}
   \resizebox{\textwidth}{!}{%
    \begin{tabular}{cccccccccc}
    \toprule
    \textbf{Method} & \textbf{GIN} & \textbf{Co-teaching} & \textbf{DivideMIX} & \textbf{SimGRACE} & \textbf{Taylor-CE} & \textbf{RTGNN} & \textbf{Sel-CL} & \textbf{OMG} & \textbf{RGLC} \\
    \midrule
    MUTAG & 0.06  & 0.11  & 0.14  & 0.09  & 0.1   & 0.11  & 0.21  & 0.18  & 0.14 \\
    NCI1  & 1.54  & 2.89  & 3.12  & 1.73  & 1.78  & 2.84  & 3.95  & 3.5   & 3.78 \\
    IMDB-B & 0.55  & 0.98  & 1.05  & 0.75  & 0.63  & 1.03  & 1.9   & 1.4   & 1.12 \\
    PROTEINS & 0.59  & 1.22  & 1.3   & 0.69  & 0.85  & 1.15  & 1.64  & 1.34  & 0.96 \\
    \bottomrule
    \end{tabular}%
  \label{tab:addlabel}%
  }
\end{table*}%
Finally, Table 4 showcases the training efficiency of our method compared to other baseline methods, with the results representing the running time of the model for each epoch. The results show that our method increases the time complexity compared to the GIN method, yet our approach outperforms GIN in terms of performance under learning with noisy labels. Furthermore, compared to the OMG method, our method demonstrates shorter running times on most datasets, and, as observed from the results in Table 1, our approach consistently surpasses the OMG method in performance.

\subsection{Parameter sensitivity experiments}
\begin{figure*}[!h]
    \centering
    \begin{subfigure}{0.30\textwidth} % 调整子图的宽度
        \centering
        \includegraphics[width=\textwidth]{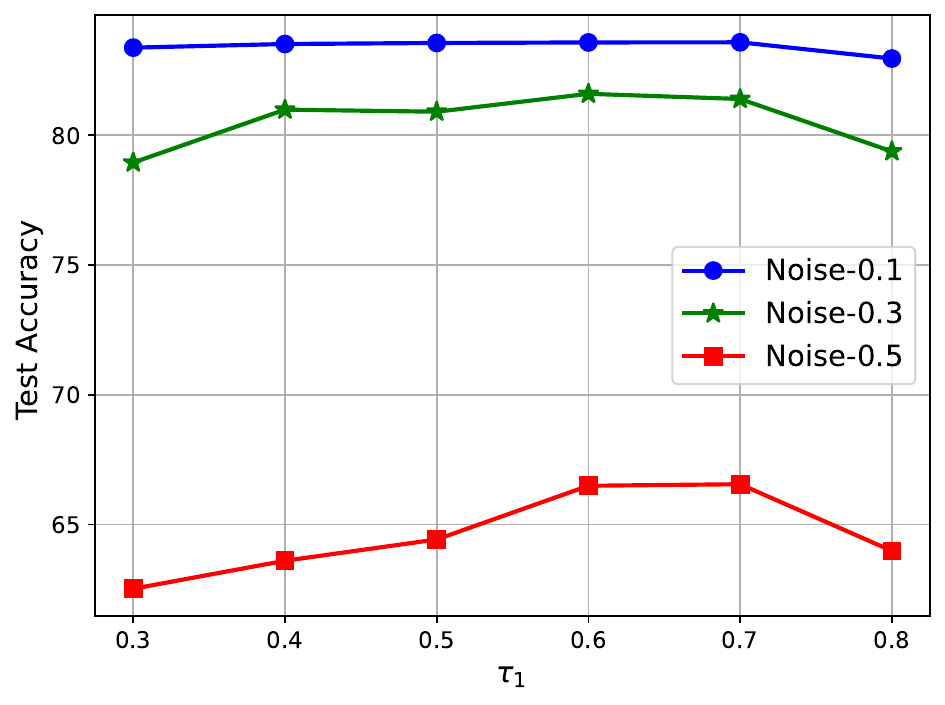}
        \caption{MUTAG}
        \label{fig:sub1}
    \end{subfigure}
    \hspace{0.00\textwidth} % 添加水平间距
    \begin{subfigure}{0.30\textwidth}
        \centering
        \includegraphics[width=\textwidth]{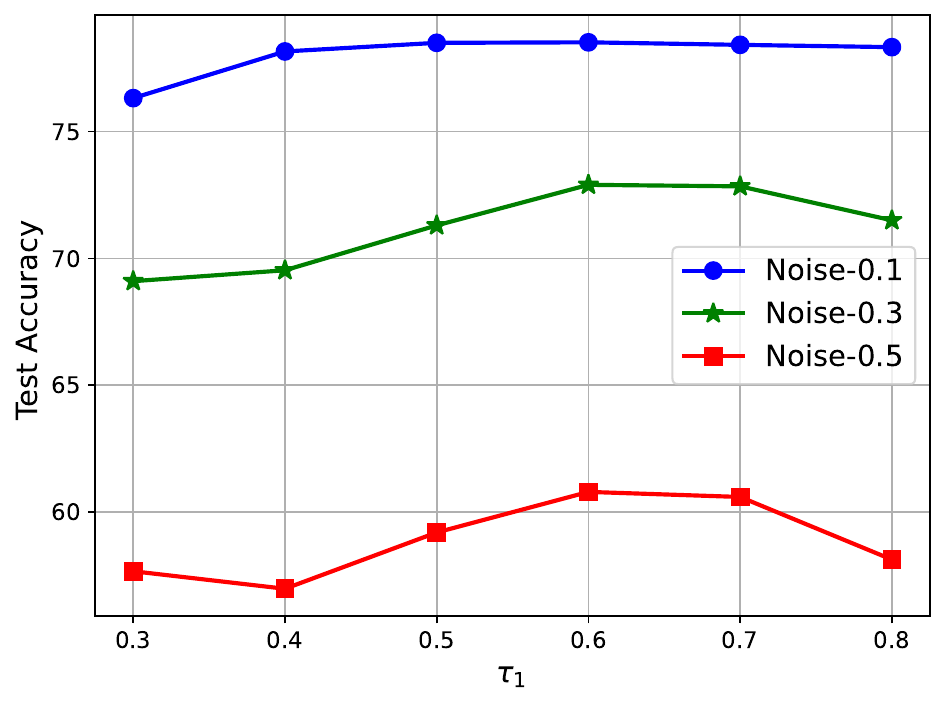}
        \caption{NCI1}
        \label{fig:sub2}
    \end{subfigure}
    \hspace{0.00\textwidth} % 添加水平间距
    \begin{subfigure}{0.30\textwidth}
        \centering
        \includegraphics[width=\textwidth]{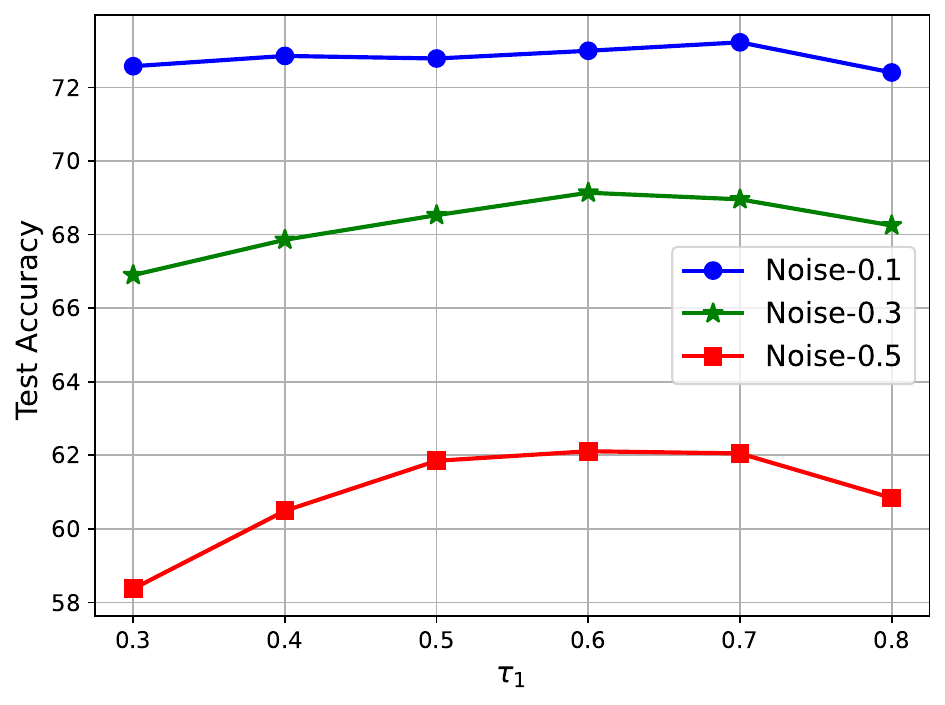}
        \caption{IMDB-B}
        \label{fig:sub2}
    \end{subfigure}
    \caption{Sensitivity of the high-confidence sample selection threshold $\tau_1$ under different noise label rates on the MUTAG, NCI1, and IMDB-B datasets.}
    \label{fig:main_figure}
\end{figure*}
\begin{figure*}[!h]
    \centering
    \begin{subfigure}{0.32\textwidth} % 调整子图的宽度
        \centering
        \includegraphics[width=\textwidth]{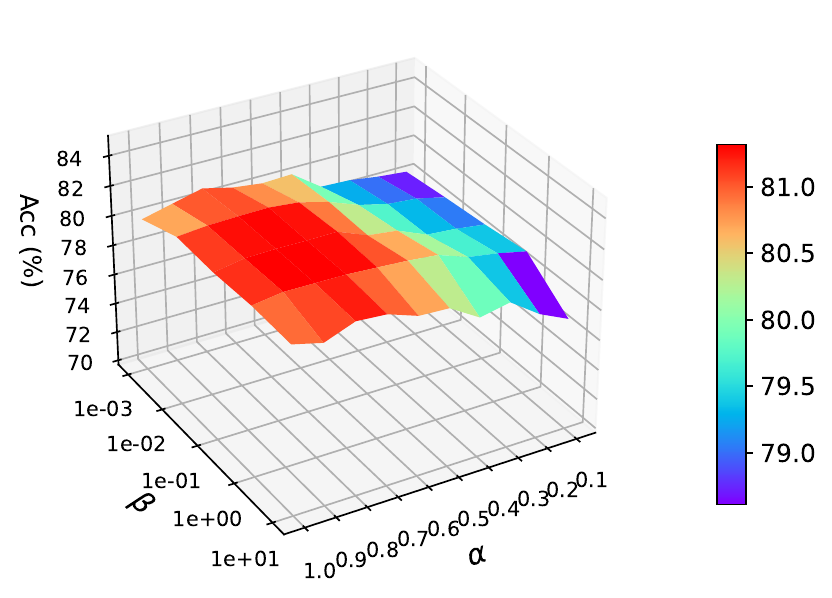}
        \caption{MUTAG}
        \label{fig:sub1}
    \end{subfigure}
    \hspace{0.00\textwidth} % 添加水平间距
    \begin{subfigure}{0.32\textwidth}
        \centering
        \includegraphics[width=\textwidth]{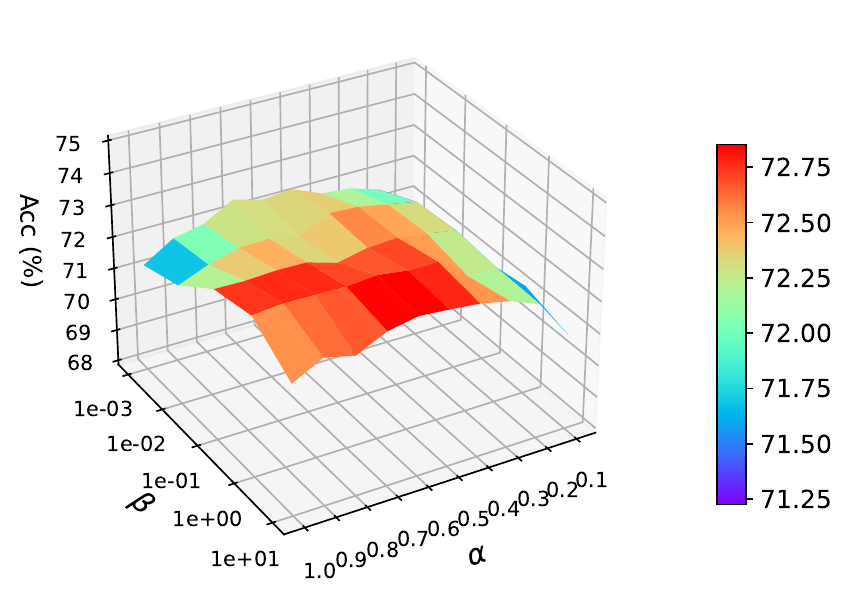}
        \caption{NCI1}
        \label{fig:sub2}
    \end{subfigure}
    \hspace{0.00\textwidth} % 添加水平间距
    \begin{subfigure}{0.32\textwidth}
        \centering
        \includegraphics[width=\textwidth]{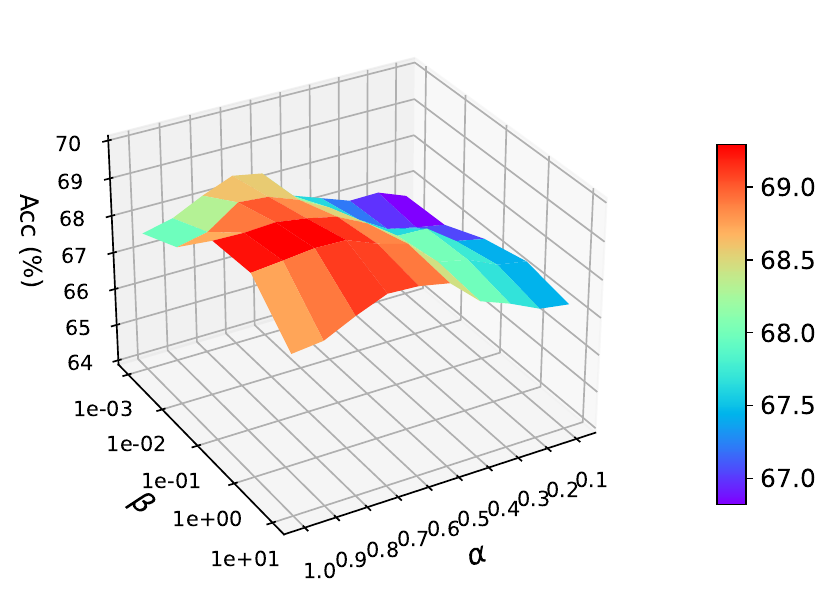}
        \caption{IMDB-B}
        \label{fig:sub2}
    \end{subfigure}
    \caption{Sensitivity of loss-constrained hyperparameters $\alpha$ and $\beta$ at 30$\%$ noise label rate on MUTAG, NCI1 and IMDB-B datasets.}
    \label{fig:main_figure}
\end{figure*}

In this section, we explore the effects of the hyperparameters $\tau_1$, $\alpha$, and $\beta$ on the performance of our RGLC method on the MUTAG, NCI1, and IMDB-B datasets under different noisy label rates, where the hyperparameter $\tau_1$ controls the number of high-confidence samples required for the computation of principal components of the category features, and the hyperparameters $\alpha$ and $\beta$ regulate the effect of corrected noise samples and supervised contrast learning on the model, respectively. $\alpha$ and $\beta$ modulate the impact of corrected noise samples and supervised graph contrastive learning on the model. To thoroughly investigate the sensitivity of the parameters, we set $\tau_1$ in the range of $\{0.3,0.4,0.5,0.6,0.7,0.8\}$ and conduct experiments with three different levels of noise label rates, 10$\%$, 30$\%$ and 50$\%$. The ranges of $\{0.1,0.2,0.3,0.4,0.5,0.6,0.7,0.8,0.9,1.0\}$ and $\{0.001,0.01,0.1,1.0,10\}$ were set up for $\alpha$ and $\beta$, respectively, and the experiments were carried out with a noise label rate of 30$\%$, and the results of the experiments are shown in Fig. 9 and 10. From Fig. 9, we observe that in most cases, as $\tau_1$ increases, the performance of the model shows a tendency to grow and then decrease; too small $\tau_1$ will introduce noise samples to interfere with the calculation of the principal components of categorical features, which makes the final screening of the noisy samples inaccurate, and too large $\tau_1$ will lead to too few clean samples that can be utilized, which also affects the calculation of the principal components of categorical features. In this article, we use the setting of $\tau_1$=0.7 to reduce the complexity of matrix decomposition while maintaining good accuracy. From Fig. 10, we can find that $\alpha$ is mainly in the range of 0.5 to 1.0 for better performance, which suggests that our noise label correction mechanism provides more clean samples for model training, while the optimal range of the hyperparameter $\beta$ for supervised graph contrastive learning is between 0.01 and 1.0. However, multiple combinations of hyperparameters affect the model's performance in the noisy label scenario. Thus, further careful tuning of our parameters is required for optimal performance.

\subsection{Limitations}
Although our RGLC method achieves superior performance on several graph categorization datasets, there are still some limitations that we need to improve in the next step. On the one hand, the embedding space-based principal component noise graph data screening mechanism for class features relies on a more robust principal component computation, and the existing graph noise labels are all class-independent noises, so if we deal with class-dependent noise labels under high noise label rates, our method will be somewhat affected. On the other hand, although supervised graph contrastive learning improves the model's performance under noisy labels, there is still a particular gap between the ideal situation and the defense against subgraph inference attacks in white-box settings under high-noise labels.

\section{Conclusion}
\label{se6}
In this paper, we rethink the impact of noise labels on graph classification from both utility and privacy aspects. It is found that the noise label not only reduces the generalization of the model but also magnifies the inference attack ability of the members of the graph or subgraph. Therefore, we use the small loss criterion and the class's first principal component vector to screen the noise samples accurately. Then, the noise label is modified based on the dual view information of the embedded and output spaces. Finally, we introduce supervised graph contrastive learning to optimize the performance of encoders and further reduce the ability of privacy attacks. Experimental results on real datasets show the proposed method can improve effectiveness and privacy protection under different noise rates. In the future, we will consider enhancing the utility and privacy of our approach in high noise rate class-dependent noise label scenarios.

\section{Acknowledgements}
 This paper was supported by the National Natural Science Foundation of China (No. U21A20474), Guangxi Science and Technology Project (GuikeAA22067070), National Natural Science Foundation of China (No. 62162005), Guangxi Science and technology project (GuikeAD21220114), Innovation Project of Guangxi Graduate Education (YCBZ2023058), Guangxi ``Bagui Scholar'' Teams for Innovation and Research Project, Guangxi Bagui Youth Talent Training Program, Guangxi Talent Highland Project of Big Data Intelligence and Application, Guangxi Collaborative Innovation Center of Multisource Information Integration and Intelligent Processing, and Center for AppliedMathematics of Guangxi (Guangxi Normal University).
% \appendix
\section{Appendix}
\label{app1}
\subsection{Analyze the influence of the amount of data and noise on the first principal component vector}
\label{app2}
\begin{figure*}[h]
\centering
    \begin{subfigure}{0.22\textwidth} % 调整子图的宽度
        \centering
        \includegraphics[width=\textwidth]{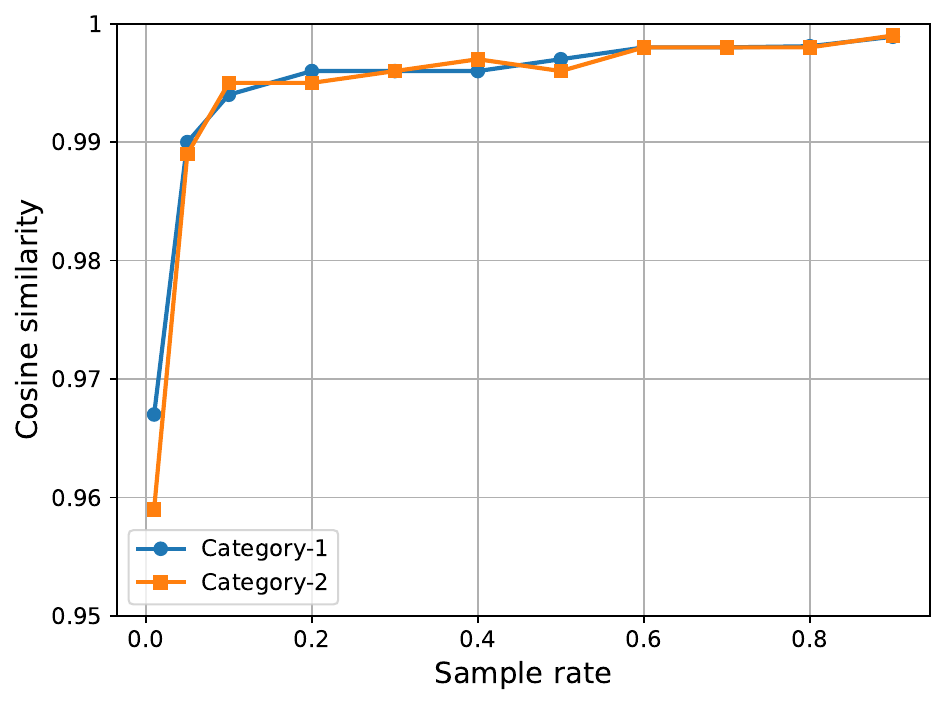}
        \caption{NCI1-0.0}
        \label{fig:sub1}
    \end{subfigure}
    \hspace{0.00\textwidth} % 添加水平间距
    \begin{subfigure}{0.22\textwidth}
        \centering
        \includegraphics[width=\textwidth]{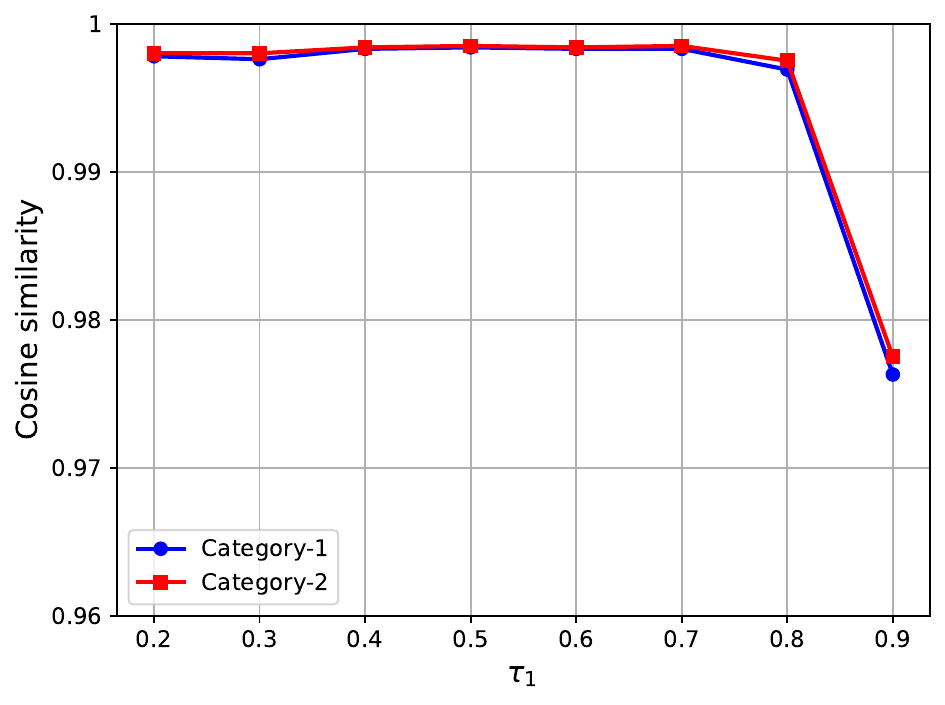}
        \caption{NCI1-0.1}
        \label{fig:sub2}
    \end{subfigure}
    \begin{subfigure}{0.22\textwidth} % 调整子图的宽度
        \centering
        \includegraphics[width=\textwidth]{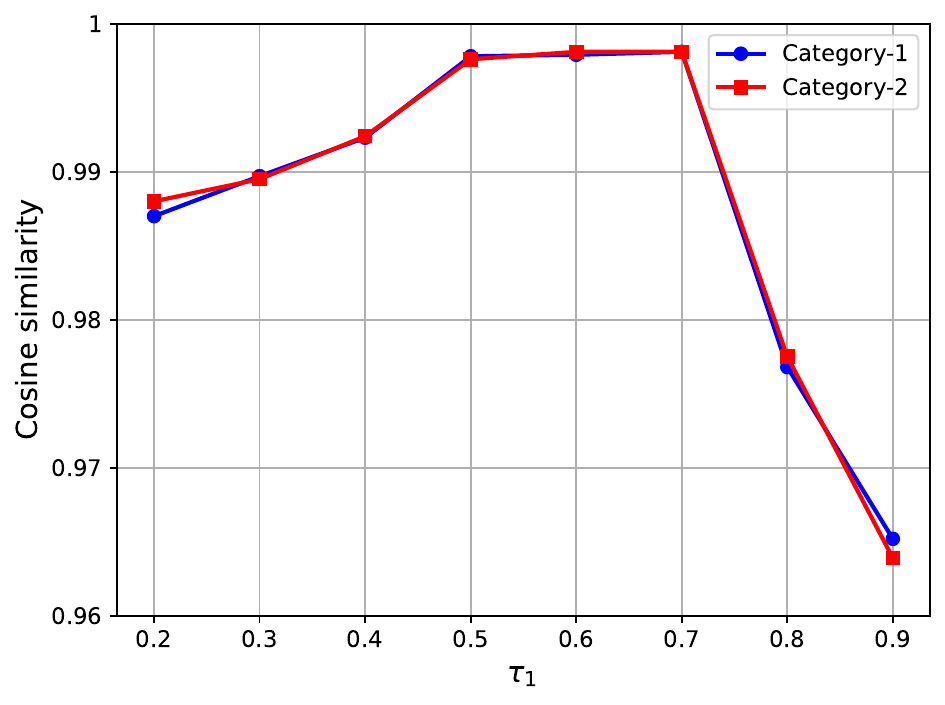}
        \caption{NCI1-0.3}
        \label{fig:sub1}
    \end{subfigure}
    \hspace{0.00\textwidth} % 添加水平间距
    \begin{subfigure}{0.22\textwidth} % 调整子图的宽度
        \centering
        \includegraphics[width=\textwidth]{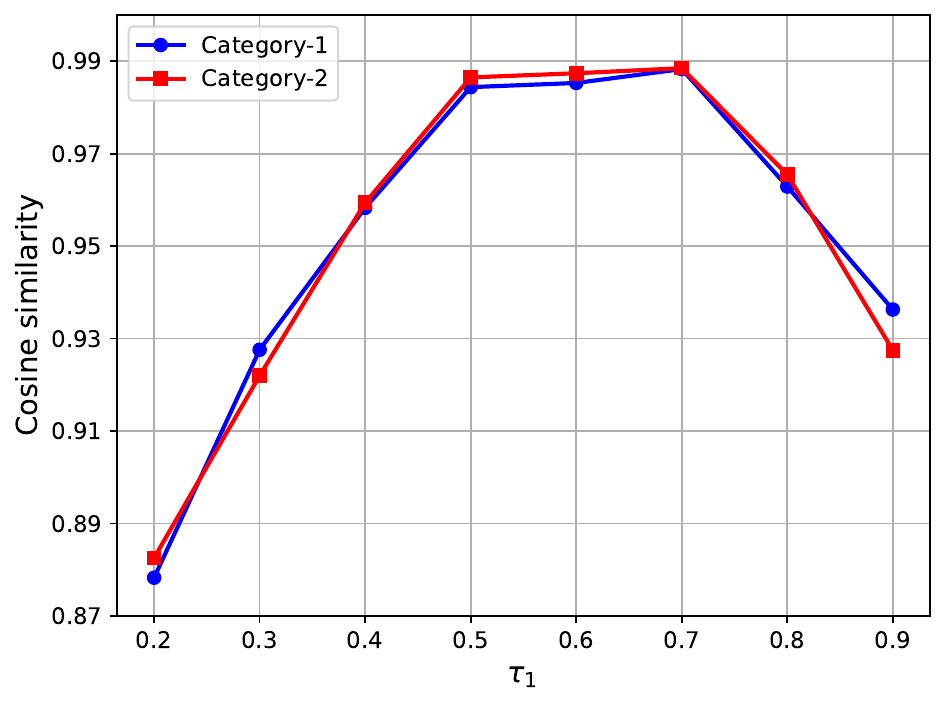}
        \caption{NCI1-0.5}
        \label{fig:sub1}
    \end{subfigure}
    \hspace{0.00\textwidth} % 添加水平间距
    \caption{Cosine similarity of the first principal component vectors of the filtered graph data to the original clean graph data for each category at different noise rates and different data volumes.}
    \label{fig:main_figure}
\end{figure*}
In this part, we validate through experiments that the first principal component vector of each category, under clean label conditions, is not significantly influenced by the volume of graph data. However, it exhibits more significant variability under noisy label conditions when the noise label rate is high. We conduct experiments on the NCI1 graph dataset at noise levels of 0$\%$, 10$\%$, 30$\%$, and 50$\%$, comparing the cosine similarity between principal component vectors obtained from high-confidence graph data selected via the small-loss criterion and those obtained from original clean label graph data. This comparison validates the results under different noise rates and data volumes. We randomly sample clean graph data from the original data, whereas, for noisy labels, we utilize the threshold \(\tau_1\) of the small-loss criterion to acquire the graph data. From Fig. 11(a), it can be seen that under clean labels, the first principal component vector of each category can achieve a cosine similarity of 0.99 with just over 5$\%$ of the data, indicating that not much graph data is needed to approximate the original first feature principal component vector. Similar results have been observed in green machine learning studies designed using applied statistics \citep{chen2020pixelhop}. Fig. 11(b)-(d) shows the similarity between the first principal component vectors calculated from the graph data selected based on the small-loss criterion and the original clean graph data under three different noise label rates. It can be observed that at lower noise rates, the similarity remains relatively stable, whereas at higher noise rates, selecting too little or too much graph data can affect the calculation of the similarity of the first principal component vector, which in turn impacts the effectiveness of distinguishing clean and noisy graph data based on the first feature principal component vector. Therefore, choosing an appropriate threshold $\tau_1$ is particularly important.
\subsection{Proof of Proposition 1}
\label{app3}
\textbf{\textit{Proof:}} Given the noisy label graph dataset $\mathcal{G}=\left\{G_1,  \ldots, G_n\right\}$ where each graph data $G_i$
  corresponds to a label $Y_i$
  and some of these labels are corrupted, we can leverage the observation from noisy label learning that models tend to fit clean labels first before gradually fitting noisy ones. This characteristic allows us to warm up the model and endow it with a certain degree of discrimination. Based on experimental findings in Appendix \ref{app2}, it has been shown that the first principal component vector is less influenced by the amount of data. When using a small-loss criterion to select high-confidence graph data, the obtained first principal component vector is close to that of the original clean graph data. This ensures that the first principal component vector derived from the small-loss criterion is less perturbed.

Then, we perform Singular Value Decomposition (SVD) on the Gram matrix of the high-confidence graph embeddings:
\begin{equation}
M=U \Sigma V^T
\end{equation}
where \(U \in \mathbb{R}^{n \times n}\) and \(V \in \mathbb{R}^{n \times n}\) are orthogonal matrices, \(\Sigma \in \mathbb{R}^{n \times d}\) is a diagonal matrix whose diagonal elements are the singular values of \(M\), arranged in descending order. The first principal component vector \(u_1\) corresponds to the first column of the \(V\) matrix. According to the principles of principal component analysis, this vector captures the direction of greatest variance in the embeddings of that category. The first principal component vector \(u_1\) satisfies the following condition:

\begin{equation}
\operatorname{argmax}_v \frac{1}{|H|} \sum_{g \in H}\left(z_i \cdot u\right)^2 \text { s.t. }\|u\|=1
\end{equation}
where \(|H|\) is the number of high-confidence graph data in each category. Based on the low-rank characteristics in the embedding space for learning with noisy labels described in \citep{wang2024low}, we measure the contribution of each graph data in that direction by filtering the alignment degree of the graph embeddings with the first principal component, namely the square of the inner product values. There exists clean graph data that has a higher degree of alignment with the first principal component vector, that is, \((z_i \cdot u_1)^2 \geq \alpha\), while the noise graph data has a poorer alignment with the first principal component vector, that is, \((z_j \cdot u_1)^2 \leq \beta\), and \(\beta < \alpha\). At the same time, based on the Linear Discriminant Analysis (LDA) assumption \citep{fisher1936use}: the distribution of embeddings follows two Gaussian distributions, each Gaussian distribution is identified as a clean distribution and a noise distribution, then we can effectively separate noise graph embeddings and clean graph embeddings based on the square of the inner product values of graph embeddings with the first principal component. Finally, by fitting the mean and variance of the distribution using a Gaussian mixture model, we can effectively filter noise graph data in the embedding space.
%% If you have bibdatabase file and want bibtex to generate the
%% bibitems, please use
%%
 \bibliographystyle{elsarticle-harv} 
 \bibliography{ref.bib}

\begin{thebibliography}{69}
\expandafter\ifx\csname natexlab\endcsname\relax\def\natexlab#1{#1}\fi
\providecommand{\url}[1]{\texttt{#1}}
\providecommand{\href}[2]{#2}
\providecommand{\path}[1]{#1}
\providecommand{\DOIprefix}{doi:}
\providecommand{\ArXivprefix}{arXiv:}
\providecommand{\URLprefix}{URL: }
\providecommand{\Pubmedprefix}{pmid:}
\providecommand{\doi}[1]{\href{http://dx.doi.org/#1}{\path{#1}}}
\providecommand{\Pubmed}[1]{\href{pmid:#1}{\path{#1}}}
\providecommand{\bibinfo}[2]{#2}
\ifx\xfnm\relax \def\xfnm[#1]{\unskip,\space#1}\fi
%Type = Inproceedings
\bibitem[{Berthelot et~al.(2019)Berthelot, Carlini, Goodfellow, Oliver, Papernot and Raffel}]{berthelot2019mixmatch}
\bibinfo{author}{Berthelot, D.}, \bibinfo{author}{Carlini, N.}, \bibinfo{author}{Goodfellow, I.}, \bibinfo{author}{Oliver, A.}, \bibinfo{author}{Papernot, N.}, \bibinfo{author}{Raffel, C.}, \bibinfo{year}{2019}.
\newblock \bibinfo{title}{Mixmatch: a holistic approach to semi-supervised learning}, in: \bibinfo{booktitle}{Proceedings of the 33rd International Conference on Neural Information Processing Systems}, pp. \bibinfo{pages}{5049--5059}.
%Type = Inproceedings
\bibitem[{Bianchi et~al.(2020)Bianchi, Grattarola and Alippi}]{bianchi2020spectral}
\bibinfo{author}{Bianchi, F.M.}, \bibinfo{author}{Grattarola, D.}, \bibinfo{author}{Alippi, C.}, \bibinfo{year}{2020}.
\newblock \bibinfo{title}{Spectral clustering with graph neural networks for graph pooling}, in: \bibinfo{booktitle}{Proceedings of the 37th International Conference on Machine Learning}, pp. \bibinfo{pages}{874--883}.
%Type = Article
\bibitem[{Borgwardt et~al.(2005)Borgwardt, Ong, Sch{\"o}nauer, Vishwanathan, Smola and Kriegel}]{borgwardt2005protein}
\bibinfo{author}{Borgwardt, K.M.}, \bibinfo{author}{Ong, C.S.}, \bibinfo{author}{Sch{\"o}nauer, S.}, \bibinfo{author}{Vishwanathan, S.}, \bibinfo{author}{Smola, A.J.}, \bibinfo{author}{Kriegel, H.P.}, \bibinfo{year}{2005}.
\newblock \bibinfo{title}{Protein function prediction via graph kernels}.
\newblock \bibinfo{journal}{Bioinformatics} \bibinfo{volume}{21}, \bibinfo{pages}{47--56}.
%Type = Article
\bibitem[{Cai and Wang(2018)}]{cai2018simple}
\bibinfo{author}{Cai, C.}, \bibinfo{author}{Wang, Y.}, \bibinfo{year}{2018}.
\newblock \bibinfo{title}{A simple yet effective baseline for non-attributed graph classification}.
\newblock \bibinfo{journal}{arXiv preprint arXiv:1811.03508} .
%Type = Article
\bibitem[{Chen et~al.(2023a)Chen, Zhang, Huang, Su, Lin, Xiao, Xia and Liu}]{chen2023erase}
\bibinfo{author}{Chen, L.H.}, \bibinfo{author}{Zhang, Y.}, \bibinfo{author}{Huang, T.}, \bibinfo{author}{Su, L.}, \bibinfo{author}{Lin, Z.}, \bibinfo{author}{Xiao, X.}, \bibinfo{author}{Xia, X.}, \bibinfo{author}{Liu, T.}, \bibinfo{year}{2023}a.
\newblock \bibinfo{title}{Erase: Error-resilient representation learning on graphs for label noise tolerance}.
\newblock \bibinfo{journal}{arXiv preprint arXiv:2312.08852} .
%Type = Inproceedings
\bibitem[{Chen et~al.(2023b)Chen, Fu, Li, Shi, Shen and Zhu}]{chen2023co}
\bibinfo{author}{Chen, X.}, \bibinfo{author}{Fu, W.}, \bibinfo{author}{Li, T.}, \bibinfo{author}{Shi, X.}, \bibinfo{author}{Shen, H.}, \bibinfo{author}{Zhu, X.}, \bibinfo{year}{2023}b.
\newblock \bibinfo{title}{Co-assistant networks for label correction}, in: \bibinfo{booktitle}{International Conference on Medical Image Computing and Computer-Assisted Intervention}, pp. \bibinfo{pages}{159--168}.
%Type = Article
\bibitem[{Chen and Kuo(2020)}]{chen2020pixelhop}
\bibinfo{author}{Chen, Y.}, \bibinfo{author}{Kuo, C.C.J.}, \bibinfo{year}{2020}.
\newblock \bibinfo{title}{Pixelhop: A successive subspace learning (ssl) method for object recognition}.
\newblock \bibinfo{journal}{Journal of Visual Communication and Image Representation} \bibinfo{volume}{70}, \bibinfo{pages}{102749}.
%Type = Inproceedings
\bibitem[{Chen et~al.(2022)Chen, Shen, Shen, Wang and Zhang}]{chen2022amplifying}
\bibinfo{author}{Chen, Y.}, \bibinfo{author}{Shen, C.}, \bibinfo{author}{Shen, Y.}, \bibinfo{author}{Wang, C.}, \bibinfo{author}{Zhang, Y.}, \bibinfo{year}{2022}.
\newblock \bibinfo{title}{Amplifying membership exposure via data poisoning}, in: \bibinfo{booktitle}{Proceedings of 36th Conference on Neural Information Processing Systems}, pp. \bibinfo{pages}{29830--29844}.
%Type = Inproceedings
\bibitem[{Dai et~al.(2021)Dai, Aggarwal and Wang}]{dai2021nrgnn}
\bibinfo{author}{Dai, E.}, \bibinfo{author}{Aggarwal, C.}, \bibinfo{author}{Wang, S.}, \bibinfo{year}{2021}.
\newblock \bibinfo{title}{Nrgnn: Learning a label noise resistant graph neural network on sparsely and noisily labeled graphs}, in: \bibinfo{booktitle}{Proceedings of the 27th ACM SIGKDD conference on knowledge discovery \& data mining}, pp. \bibinfo{pages}{227--236}.
%Type = Article
\bibitem[{Debnath et~al.(1991)Debnath, Lopez~de Compadre, Debnath, Shusterman and Hansch}]{debnath1991structure}
\bibinfo{author}{Debnath, A.K.}, \bibinfo{author}{Lopez~de Compadre, R.L.}, \bibinfo{author}{Debnath, G.}, \bibinfo{author}{Shusterman, A.J.}, \bibinfo{author}{Hansch, C.}, \bibinfo{year}{1991}.
\newblock \bibinfo{title}{Structure-activity relationship of mutagenic aromatic and heteroaromatic nitro compounds. correlation with molecular orbital energies and hydrophobicity}.
\newblock \bibinfo{journal}{Journal of medicinal chemistry} \bibinfo{volume}{34}, \bibinfo{pages}{786--797}.
%Type = Inproceedings
\bibitem[{Di et~al.(2021)Di, Yao, Zhang and Chen}]{DBLP:conf/icde/DiYZC21}
\bibinfo{author}{Di, S.}, \bibinfo{author}{Yao, Q.}, \bibinfo{author}{Zhang, Y.}, \bibinfo{author}{Chen, L.}, \bibinfo{year}{2021}.
\newblock \bibinfo{title}{Efficient relation-aware scoring function search for knowledge graph embedding}, in: \bibinfo{booktitle}{37th {IEEE} International Conference on Data Engineering}, pp. \bibinfo{pages}{1104--1115}.
%Type = Inproceedings
\bibitem[{Englesson and Azizpour(2021)}]{DBLP:conf/nips/EnglessonA21}
\bibinfo{author}{Englesson, E.}, \bibinfo{author}{Azizpour, H.}, \bibinfo{year}{2021}.
\newblock \bibinfo{title}{Generalized jensen-shannon divergence loss for learning with noisy labels}, in: \bibinfo{booktitle}{Advances in Neural Information Processing Systems 34: Annual Conference on Neural Information Processing Systems 2021}, pp. \bibinfo{pages}{30284--30297}.
%Type = Inproceedings
\bibitem[{Feng et~al.(2021)Feng, Shu, Lin, Lv, Li and An}]{feng2021can}
\bibinfo{author}{Feng, L.}, \bibinfo{author}{Shu, S.}, \bibinfo{author}{Lin, Z.}, \bibinfo{author}{Lv, F.}, \bibinfo{author}{Li, L.}, \bibinfo{author}{An, B.}, \bibinfo{year}{2021}.
\newblock \bibinfo{title}{Can cross entropy loss be robust to label noise?}, in: \bibinfo{booktitle}{Proceedings of the twenty-ninth international conference on international joint conferences on artificial intelligence}, pp. \bibinfo{pages}{2206--2212}.
%Type = Article
\bibitem[{Fisher(1936)}]{fisher1936use}
\bibinfo{author}{Fisher, R.A.}, \bibinfo{year}{1936}.
\newblock \bibinfo{title}{The use of multiple measurements in taxonomic problems}.
\newblock \bibinfo{journal}{Annals of eugenics} \bibinfo{volume}{7}, \bibinfo{pages}{179--188}.
%Type = Article
\bibitem[{Fontanesi et~al.(2023)Fontanesi, Micheli, Milazzo and Podda}]{fontanesi2023exploiting}
\bibinfo{author}{Fontanesi, M.}, \bibinfo{author}{Micheli, A.}, \bibinfo{author}{Milazzo, P.}, \bibinfo{author}{Podda, M.}, \bibinfo{year}{2023}.
\newblock \bibinfo{title}{Exploiting the structure of biochemical pathways to investigate dynamical properties with neural networks for graphs}.
\newblock \bibinfo{journal}{Bioinformatics} \bibinfo{volume}{39}, \bibinfo{pages}{btad678}.
%Type = Inproceedings
\bibitem[{Ghazi et~al.(2021)Ghazi, Golowich, Kumar, Manurangsi and Zhang}]{ghazi2021deep}
\bibinfo{author}{Ghazi, B.}, \bibinfo{author}{Golowich, N.}, \bibinfo{author}{Kumar, R.}, \bibinfo{author}{Manurangsi, P.}, \bibinfo{author}{Zhang, C.}, \bibinfo{year}{2021}.
\newblock \bibinfo{title}{Deep learning with label differential privacy}, in: \bibinfo{booktitle}{Advances in Neural Information Processing Systems 34: Annual Conference on Neural Information Processing Systems 2021}, pp. \bibinfo{pages}{27131--27145}.
%Type = Inproceedings
\bibitem[{Han et~al.(2018)Han, Yao, Yu, Niu, Xu, Hu, Tsang and Sugiyama}]{han2018co}
\bibinfo{author}{Han, B.}, \bibinfo{author}{Yao, Q.}, \bibinfo{author}{Yu, X.}, \bibinfo{author}{Niu, G.}, \bibinfo{author}{Xu, M.}, \bibinfo{author}{Hu, W.}, \bibinfo{author}{Tsang, I.W.}, \bibinfo{author}{Sugiyama, M.}, \bibinfo{year}{2018}.
\newblock \bibinfo{title}{Co-teaching: robust training of deep neural networks with extremely noisy labels}, in: \bibinfo{booktitle}{Proceedings of the 32nd International Conference on Neural Information Processing Systems}, pp. \bibinfo{pages}{8536--8546}.
%Type = Article
\bibitem[{Hu et~al.(2023)Hu, Li, Lin, Peng, Zhang, Zhang and Dong}]{hu2023defending}
\bibinfo{author}{Hu, L.}, \bibinfo{author}{Li, J.}, \bibinfo{author}{Lin, G.}, \bibinfo{author}{Peng, S.}, \bibinfo{author}{Zhang, Z.}, \bibinfo{author}{Zhang, Y.}, \bibinfo{author}{Dong, C.}, \bibinfo{year}{2023}.
\newblock \bibinfo{title}{Defending against membership inference attacks with high utility by gan}.
\newblock \bibinfo{journal}{IEEE Transactions on Dependable and Secure Computing} \bibinfo{volume}{20}, \bibinfo{pages}{2144--2157}.
%Type = Inproceedings
\bibitem[{Ju et~al.(2022)Ju, Luo, Qu, Wang, Chen, Deng, Hua and Zhang}]{DBLP:conf/ijcai/JuLQWCD0Z22}
\bibinfo{author}{Ju, W.}, \bibinfo{author}{Luo, X.}, \bibinfo{author}{Qu, M.}, \bibinfo{author}{Wang, Y.}, \bibinfo{author}{Chen, C.}, \bibinfo{author}{Deng, M.}, \bibinfo{author}{Hua, X.}, \bibinfo{author}{Zhang, M.}, \bibinfo{year}{2022}.
\newblock \bibinfo{title}{{TGNN:} {A} joint semi-supervised framework for graph-level classification}, in: \bibinfo{booktitle}{Proceedings of the Thirty-First International Joint Conference on Artificial Intelligence}, pp. \bibinfo{pages}{2122--2128}.
%Type = Inproceedings
\bibitem[{Karim et~al.(2022)Karim, Rizve, Rahnavard, Mian and Shah}]{karim2022unicon}
\bibinfo{author}{Karim, N.}, \bibinfo{author}{Rizve, M.N.}, \bibinfo{author}{Rahnavard, N.}, \bibinfo{author}{Mian, A.}, \bibinfo{author}{Shah, M.}, \bibinfo{year}{2022}.
\newblock \bibinfo{title}{Unicon: Combating label noise through uniform selection and contrastive learning}, in: \bibinfo{booktitle}{Proceedings of the IEEE/CVF Conference on Computer Vision and Pattern Recognition}, pp. \bibinfo{pages}{9676--9686}.
%Type = Inproceedings
\bibitem[{Kornblith et~al.(2019)Kornblith, Norouzi, Lee and Hinton}]{kornblith2019similarity}
\bibinfo{author}{Kornblith, S.}, \bibinfo{author}{Norouzi, M.}, \bibinfo{author}{Lee, H.}, \bibinfo{author}{Hinton, G.}, \bibinfo{year}{2019}.
\newblock \bibinfo{title}{Similarity of neural network representations revisited}, in: \bibinfo{booktitle}{International conference on machine learning}, pp. \bibinfo{pages}{3519--3529}.
%Type = Inproceedings
\bibitem[{Kriege and Mutzel(2012)}]{kriege2012subgraph}
\bibinfo{author}{Kriege, N.}, \bibinfo{author}{Mutzel, P.}, \bibinfo{year}{2012}.
\newblock \bibinfo{title}{Subgraph matching kernels for attributed graphs}, in: \bibinfo{booktitle}{Proceedings of the 29th International Coference on International Conference on Machine Learning}, pp. \bibinfo{pages}{291--298}.
%Type = Inproceedings
\bibitem[{Lee et~al.(2019)Lee, Yun, Lee, Lee, Li and Shin}]{lee2019robust}
\bibinfo{author}{Lee, K.}, \bibinfo{author}{Yun, S.}, \bibinfo{author}{Lee, K.}, \bibinfo{author}{Lee, H.}, \bibinfo{author}{Li, B.}, \bibinfo{author}{Shin, J.}, \bibinfo{year}{2019}.
\newblock \bibinfo{title}{Robust inference via generative classifiers for handling noisy labels}, in: \bibinfo{booktitle}{International conference on machine learning}, pp. \bibinfo{pages}{3763--3772}.
%Type = Inproceedings
\bibitem[{Li et~al.(2020)Li, Socher and Hoi}]{DBLP:conf/iclr/LiSH20}
\bibinfo{author}{Li, J.}, \bibinfo{author}{Socher, R.}, \bibinfo{author}{Hoi, S.C.H.}, \bibinfo{year}{2020}.
\newblock \bibinfo{title}{Dividemix: Learning with noisy labels as semi-supervised learning}, in: \bibinfo{booktitle}{8th International Conference on Learning Representations}.
%Type = Inproceedings
\bibitem[{Li et~al.(2022)Li, Xia, Ge and Liu}]{li2022selective}
\bibinfo{author}{Li, S.}, \bibinfo{author}{Xia, X.}, \bibinfo{author}{Ge, S.}, \bibinfo{author}{Liu, T.}, \bibinfo{year}{2022}.
\newblock \bibinfo{title}{Selective-supervised contrastive learning with noisy labels}, in: \bibinfo{booktitle}{Proceedings of the IEEE/CVF conference on computer vision and pattern recognition}, pp. \bibinfo{pages}{316--325}.
%Type = Article
\bibitem[{Li et~al.(2024)Li, Li, Li, Qian and Wang}]{LI2024106113}
\bibinfo{author}{Li, X.}, \bibinfo{author}{Li, Q.}, \bibinfo{author}{Li, D.}, \bibinfo{author}{Qian, H.}, \bibinfo{author}{Wang, J.}, \bibinfo{year}{2024}.
\newblock \bibinfo{title}{Contrastive learning of graphs under label noise}.
\newblock \bibinfo{journal}{Neural Networks} \bibinfo{volume}{172}, \bibinfo{pages}{106113}.
%Type = Inproceedings
\bibitem[{Li et~al.(2023)Li, Han, Shan and Chen}]{li2023disc}
\bibinfo{author}{Li, Y.}, \bibinfo{author}{Han, H.}, \bibinfo{author}{Shan, S.}, \bibinfo{author}{Chen, X.}, \bibinfo{year}{2023}.
\newblock \bibinfo{title}{Disc: Learning from noisy labels via dynamic instance-specific selection and correction}, in: \bibinfo{booktitle}{Proceedings of the IEEE/CVF Conference on Computer Vision and Pattern Recognition}, pp. \bibinfo{pages}{24070--24079}.
%Type = Inproceedings
\bibitem[{Liang et~al.(2018)Liang, Li and Srikant}]{DBLP:conf/iclr/LiangLS18}
\bibinfo{author}{Liang, S.}, \bibinfo{author}{Li, Y.}, \bibinfo{author}{Srikant, R.}, \bibinfo{year}{2018}.
\newblock \bibinfo{title}{Enhancing the reliability of out-of-distribution image detection in neural networks}, in: \bibinfo{booktitle}{6th International Conference on Learning Representations}.
%Type = Inproceedings
\bibitem[{Liu et~al.(2022a)Liu, Wen, He, Salem, Zhang, Backes, Cristofaro, Fritz and Zhang}]{DBLP:conf/uss/LiuWH000CF022}
\bibinfo{author}{Liu, Y.}, \bibinfo{author}{Wen, R.}, \bibinfo{author}{He, X.}, \bibinfo{author}{Salem, A.}, \bibinfo{author}{Zhang, Z.}, \bibinfo{author}{Backes, M.}, \bibinfo{author}{Cristofaro, E.D.}, \bibinfo{author}{Fritz, M.}, \bibinfo{author}{Zhang, Y.}, \bibinfo{year}{2022}a.
\newblock \bibinfo{title}{Ml-doctor: Holistic risk assessment of inference attacks against machine learning models}, in: \bibinfo{booktitle}{31st {USENIX} Security Symposium}, pp. \bibinfo{pages}{4525--4542}.
%Type = Inproceedings
\bibitem[{Liu et~al.(2022b)Liu, Zhao, Backes and Zhang}]{liu2022membership}
\bibinfo{author}{Liu, Y.}, \bibinfo{author}{Zhao, Z.}, \bibinfo{author}{Backes, M.}, \bibinfo{author}{Zhang, Y.}, \bibinfo{year}{2022}b.
\newblock \bibinfo{title}{Membership inference attacks by exploiting loss trajectory}, in: \bibinfo{booktitle}{Proceedings of the 2022 ACM SIGSAC Conference on Computer and Communications Security}, pp. \bibinfo{pages}{2085--2098}.
%Type = Inproceedings
\bibitem[{Long et~al.(2021)Long, Jin, Wu and Song}]{long2021theoretically}
\bibinfo{author}{Long, Q.}, \bibinfo{author}{Jin, Y.}, \bibinfo{author}{Wu, Y.}, \bibinfo{author}{Song, G.}, \bibinfo{year}{2021}.
\newblock \bibinfo{title}{Theoretically improving graph neural networks via anonymous walk graph kernels}, in: \bibinfo{booktitle}{Proceedings of the Web Conference 2021}, pp. \bibinfo{pages}{1204--1214}.
%Type = Article
\bibitem[{Lu et~al.(2024)Lu, Liu, Wen, Zhou, Zhang and Zhang}]{lu2024noise}
\bibinfo{author}{Lu, Z.}, \bibinfo{author}{Liu, Y.}, \bibinfo{author}{Wen, G.}, \bibinfo{author}{Zhou, B.}, \bibinfo{author}{Zhang, W.}, \bibinfo{author}{Zhang, J.}, \bibinfo{year}{2024}.
\newblock \bibinfo{title}{Noise-resistant graph neural networks with manifold consistency and label consistency}.
\newblock \bibinfo{journal}{Expert Systems with Applications} \bibinfo{volume}{245}, \bibinfo{pages}{123120}.
%Type = Inproceedings
\bibitem[{Luo et~al.(2022)Luo, Ju, Qu and Chen}]{DBLP:conf/icde/LuoJQCDHZ22}
\bibinfo{author}{Luo, X.}, \bibinfo{author}{Ju, W.}, \bibinfo{author}{Qu, M.}, \bibinfo{author}{Chen, C.}, \bibinfo{year}{2022}.
\newblock \bibinfo{title}{Dualgraph: Improving semi-supervised graph classification via dual contrastive learning}, in: \bibinfo{booktitle}{38th {IEEE} International Conference on Data Engineering}, pp. \bibinfo{pages}{699--712}.
%Type = Article
\bibitem[{Luo et~al.(2024)Luo, Zhao, Qin, Ju and Zhang}]{luo2024towards}
\bibinfo{author}{Luo, X.}, \bibinfo{author}{Zhao, Y.}, \bibinfo{author}{Qin, Y.}, \bibinfo{author}{Ju, W.}, \bibinfo{author}{Zhang, M.}, \bibinfo{year}{2024}.
\newblock \bibinfo{title}{Towards semi-supervised universal graph classification}.
\newblock \bibinfo{journal}{IEEE Transactions on Knowledge and Data Engineering} \bibinfo{volume}{36}, \bibinfo{pages}{416--428}.
%Type = Article
\bibitem[{van~der Maaten and Hinton(2008)}]{van2008visualizing}
\bibinfo{author}{van~der Maaten, L.}, \bibinfo{author}{Hinton, G.}, \bibinfo{year}{2008}.
\newblock \bibinfo{title}{Visualizing data using t-sne}.
\newblock \bibinfo{journal}{Journal of Machine Learning Research} \bibinfo{volume}{9}, \bibinfo{pages}{2579--2605}.
%Type = Inproceedings
\bibitem[{Mo et~al.(2022)Mo, Peng, Xu, Shi and Zhu}]{mo2022simple}
\bibinfo{author}{Mo, Y.}, \bibinfo{author}{Peng, L.}, \bibinfo{author}{Xu, J.}, \bibinfo{author}{Shi, X.}, \bibinfo{author}{Zhu, X.}, \bibinfo{year}{2022}.
\newblock \bibinfo{title}{Simple unsupervised graph representation learning}, in: \bibinfo{booktitle}{Proceedings of the AAAI Conference on Artificial Intelligence}, pp. \bibinfo{pages}{7797--7805}.
%Type = Inproceedings
\bibitem[{Ortego et~al.(2021)Ortego, Arazo, Albert, O'Connor and McGuinness}]{ortego2021multi}
\bibinfo{author}{Ortego, D.}, \bibinfo{author}{Arazo, E.}, \bibinfo{author}{Albert, P.}, \bibinfo{author}{O'Connor, N.E.}, \bibinfo{author}{McGuinness, K.}, \bibinfo{year}{2021}.
\newblock \bibinfo{title}{Multi-objective interpolation training for robustness to label noise}, in: \bibinfo{booktitle}{Proceedings of the IEEE/CVF Conference on Computer Vision and Pattern Recognition}, pp. \bibinfo{pages}{6606--6615}.
%Type = Inproceedings
\bibitem[{Qian et~al.(2023)Qian, Ying, Hu, Zhou, Chen, Chen and Wu}]{qian2023robust}
\bibinfo{author}{Qian, S.}, \bibinfo{author}{Ying, H.}, \bibinfo{author}{Hu, R.}, \bibinfo{author}{Zhou, J.}, \bibinfo{author}{Chen, J.}, \bibinfo{author}{Chen, D.Z.}, \bibinfo{author}{Wu, J.}, \bibinfo{year}{2023}.
\newblock \bibinfo{title}{Robust training of graph neural networks via noise governance}, in: \bibinfo{booktitle}{Proceedings of the Sixteenth ACM International Conference on Web Search and Data Mining}, pp. \bibinfo{pages}{607--615}.
%Type = Article
\bibitem[{Rahmani et~al.(2023)Rahmani, Baghbani, Bouguila and Patterson}]{rahmani2023graph}
\bibinfo{author}{Rahmani, S.}, \bibinfo{author}{Baghbani, A.}, \bibinfo{author}{Bouguila, N.}, \bibinfo{author}{Patterson, Z.}, \bibinfo{year}{2023}.
\newblock \bibinfo{title}{Graph neural networks for intelligent transportation systems: A survey}.
\newblock \bibinfo{journal}{IEEE Transactions on Intelligent Transportation Systems} \bibinfo{volume}{24}, \bibinfo{pages}{8846--8885}.
%Type = Inproceedings
\bibitem[{Salem et~al.(2019)Salem, Zhang, Humbert, Berrang, Fritz and Backes}]{salem2019ml}
\bibinfo{author}{Salem, A.}, \bibinfo{author}{Zhang, Y.}, \bibinfo{author}{Humbert, M.}, \bibinfo{author}{Berrang, P.}, \bibinfo{author}{Fritz, M.}, \bibinfo{author}{Backes, M.}, \bibinfo{year}{2019}.
\newblock \bibinfo{title}{Ml-leaks: Model and data independent membership inference attacks and defenses on machine learning models}, in: \bibinfo{booktitle}{Network and Distributed Systems Security (NDSS) Symposium 2019}.
%Type = Inproceedings
\bibitem[{Schweimer et~al.(2022)Schweimer, Gfrerer, Lugstein, Pape, Velimsky, Els{\"a}sser and Geiger}]{schweimer2022generating}
\bibinfo{author}{Schweimer, C.}, \bibinfo{author}{Gfrerer, C.}, \bibinfo{author}{Lugstein, F.}, \bibinfo{author}{Pape, D.}, \bibinfo{author}{Velimsky, J.A.}, \bibinfo{author}{Els{\"a}sser, R.}, \bibinfo{author}{Geiger, B.C.}, \bibinfo{year}{2022}.
\newblock \bibinfo{title}{Generating simple directed social network graphs for information spreading}, in: \bibinfo{booktitle}{Proceedings of the ACM Web Conference 2022}, pp. \bibinfo{pages}{1475--1485}.
%Type = Article
\bibitem[{Shervashidze et~al.(2011)Shervashidze, Schweitzer and van Leeuwen}]{DBLP:journals/jmlr/ShervashidzeSLMB11}
\bibinfo{author}{Shervashidze, N.}, \bibinfo{author}{Schweitzer, P.}, \bibinfo{author}{van Leeuwen, E.J.}, \bibinfo{year}{2011}.
\newblock \bibinfo{title}{Weisfeiler-lehman graph kernels}.
\newblock \bibinfo{journal}{Journal of Machine Learning Research} \bibinfo{volume}{12}, \bibinfo{pages}{2539--2561}.
%Type = Inproceedings
\bibitem[{Shokri et~al.(2017)Shokri, Stronati, Song and Shmatikov}]{shokri2017membership}
\bibinfo{author}{Shokri, R.}, \bibinfo{author}{Stronati, M.}, \bibinfo{author}{Song, C.}, \bibinfo{author}{Shmatikov, V.}, \bibinfo{year}{2017}.
\newblock \bibinfo{title}{Membership inference attacks against machine learning models}, in: \bibinfo{booktitle}{2017 IEEE symposium on security and privacy (SP)}, pp. \bibinfo{pages}{3--18}.
%Type = Inproceedings
\bibitem[{Song et~al.(2019)Song, Shokri and Mittal}]{song2019privacy}
\bibinfo{author}{Song, L.}, \bibinfo{author}{Shokri, R.}, \bibinfo{author}{Mittal, P.}, \bibinfo{year}{2019}.
\newblock \bibinfo{title}{Privacy risks of securing machine learning models against adversarial examples}, in: \bibinfo{booktitle}{Proceedings of the 2019 ACM SIGSAC Conference on Computer and Communications Security}, pp. \bibinfo{pages}{241--257}.
%Type = Article
\bibitem[{Tam et~al.(2023)Tam, Li, Han, Xu and Fu}]{tam2023federated}
\bibinfo{author}{Tam, K.}, \bibinfo{author}{Li, L.}, \bibinfo{author}{Han, B.}, \bibinfo{author}{Xu, C.}, \bibinfo{author}{Fu, H.}, \bibinfo{year}{2023}.
\newblock \bibinfo{title}{Federated noisy client learning}.
\newblock \bibinfo{journal}{IEEE Transactions on Neural Networks and Learning Systems} .
%Type = Inproceedings
\bibitem[{Togninalli et~al.(2019)Togninalli, Ghisu, Llinares{-}L{\'{o}}pez, Rieck and Borgwardt}]{DBLP:conf/nips/TogninalliGLRB19}
\bibinfo{author}{Togninalli, M.}, \bibinfo{author}{Ghisu, M.E.}, \bibinfo{author}{Llinares{-}L{\'{o}}pez, F.}, \bibinfo{author}{Rieck, B.}, \bibinfo{author}{Borgwardt, K.M.}, \bibinfo{year}{2019}.
\newblock \bibinfo{title}{Wasserstein weisfeiler-lehman graph kernels}, in: \bibinfo{booktitle}{Advances in Neural Information Processing Systems 32: Annual Conference on Neural Information Processing Systems 2019}, pp. \bibinfo{pages}{6436--6446}.
%Type = Inproceedings
\bibitem[{Tram{\`e}r et~al.(2022)Tram{\`e}r, Shokri, San~Joaquin, Le, Jagielski, Hong and Carlini}]{tramer2022truth}
\bibinfo{author}{Tram{\`e}r, F.}, \bibinfo{author}{Shokri, R.}, \bibinfo{author}{San~Joaquin, A.}, \bibinfo{author}{Le, H.}, \bibinfo{author}{Jagielski, M.}, \bibinfo{author}{Hong, S.}, \bibinfo{author}{Carlini, N.}, \bibinfo{year}{2022}.
\newblock \bibinfo{title}{Truth serum: Poisoning machine learning models to reveal their secrets}, in: \bibinfo{booktitle}{Proceedings of the 2022 ACM SIGSAC Conference on Computer and Communications Security}, pp. \bibinfo{pages}{2779--2792}.
%Type = Article
\bibitem[{Wale et~al.(2008)Wale, Watson and Karypis}]{wale2008comparison}
\bibinfo{author}{Wale, N.}, \bibinfo{author}{Watson, I.A.}, \bibinfo{author}{Karypis, G.}, \bibinfo{year}{2008}.
\newblock \bibinfo{title}{Comparison of descriptor spaces for chemical compound retrieval and classification}.
\newblock \bibinfo{journal}{Knowledge and Information Systems} \bibinfo{volume}{14}, \bibinfo{pages}{347--375}.
%Type = Inproceedings
\bibitem[{Wang and Isola(2020)}]{wang2020understanding}
\bibinfo{author}{Wang, T.}, \bibinfo{author}{Isola, P.}, \bibinfo{year}{2020}.
\newblock \bibinfo{title}{Understanding contrastive representation learning through alignment and uniformity on the hypersphere}, in: \bibinfo{booktitle}{Proceedings of the 37th International Conference on Machine Learning}, pp. \bibinfo{pages}{9929--9939}.
%Type = Article
\bibitem[{Wang and Yang(2024)}]{wang2024low}
\bibinfo{author}{Wang, Y.}, \bibinfo{author}{Yang, Y.}, \bibinfo{year}{2024}.
\newblock \bibinfo{title}{Low-rank graph contrastive learning for node classification}.
\newblock \bibinfo{journal}{arXiv preprint arXiv:2402.09600} .
%Type = Article
\bibitem[{Wu et~al.(2022)Wu, Zhan, Zhang, Luo and Tang}]{wu2022mtgcn}
\bibinfo{author}{Wu, Z.}, \bibinfo{author}{Zhan, M.}, \bibinfo{author}{Zhang, H.}, \bibinfo{author}{Luo, Q.}, \bibinfo{author}{Tang, K.}, \bibinfo{year}{2022}.
\newblock \bibinfo{title}{Mtgcn: A multi-task approach for node classification and link prediction in graph data}.
\newblock \bibinfo{journal}{Information Processing \& Management} \bibinfo{volume}{59}, \bibinfo{pages}{102902}.
%Type = Article
\bibitem[{Xia et~al.(2024)Xia, Lin, Xu, Tan, Wu, Li and Li}]{xia2024gnn}
\bibinfo{author}{Xia, J.}, \bibinfo{author}{Lin, H.}, \bibinfo{author}{Xu, Y.}, \bibinfo{author}{Tan, C.}, \bibinfo{author}{Wu, L.}, \bibinfo{author}{Li, S.}, \bibinfo{author}{Li, S.Z.}, \bibinfo{year}{2024}.
\newblock \bibinfo{title}{Gnn cleaner: Label cleaner for graph structured data}.
\newblock \bibinfo{journal}{IEEE Transactions on Knowledge \& Data Engineering} \bibinfo{volume}{36}, \bibinfo{pages}{640--651}.
%Type = Inproceedings
\bibitem[{Xia et~al.(2022)Xia, Wu, Chen, Hu and Li}]{xia2022simgrace}
\bibinfo{author}{Xia, J.}, \bibinfo{author}{Wu, L.}, \bibinfo{author}{Chen, J.}, \bibinfo{author}{Hu, B.}, \bibinfo{author}{Li, S.Z.}, \bibinfo{year}{2022}.
\newblock \bibinfo{title}{Simgrace: A simple framework for graph contrastive learning without data augmentation}, in: \bibinfo{booktitle}{Proceedings of the ACM Web Conference 2022}, pp. \bibinfo{pages}{1070--1079}.
%Type = Inproceedings
\bibitem[{Xia et~al.(2021)Xia, Liu, Han, Gong, Wang, Ge and Chang}]{xia2021robust}
\bibinfo{author}{Xia, X.}, \bibinfo{author}{Liu, T.}, \bibinfo{author}{Han, B.}, \bibinfo{author}{Gong, C.}, \bibinfo{author}{Wang, N.}, \bibinfo{author}{Ge, Z.}, \bibinfo{author}{Chang, Y.}, \bibinfo{year}{2021}.
\newblock \bibinfo{title}{Robust early-learning: hindering the memorization of noisy labels}, in: \bibinfo{booktitle}{International Conference on Learning Representations 2021}, pp. \bibinfo{pages}{1--15}.
%Type = Article
\bibitem[{Xie et~al.(2023)Xie, Kannan and Kuo}]{xie2023label}
\bibinfo{author}{Xie, T.}, \bibinfo{author}{Kannan, R.}, \bibinfo{author}{Kuo, C.J.}, \bibinfo{year}{2023}.
\newblock \bibinfo{title}{Label efficient regularization and propagation for graph node classification}.
\newblock \bibinfo{journal}{IEEE transactions on pattern analysis and machine intelligence} \bibinfo{volume}{45}, \bibinfo{pages}{14856--14871}.
%Type = Inproceedings
\bibitem[{Xu et~al.(2019)Xu, Hu and Leskovec}]{DBLP:conf/iclr/XuHLJ19}
\bibinfo{author}{Xu, K.}, \bibinfo{author}{Hu, W.}, \bibinfo{author}{Leskovec, J.}, \bibinfo{year}{2019}.
\newblock \bibinfo{title}{How powerful are graph neural networks?}, in: \bibinfo{booktitle}{7th International Conference on Learning Representations}.
%Type = Inproceedings
\bibitem[{Yanardag and Vishwanathan(2015)}]{yanardag2015deep}
\bibinfo{author}{Yanardag, P.}, \bibinfo{author}{Vishwanathan, S.}, \bibinfo{year}{2015}.
\newblock \bibinfo{title}{Deep graph kernels}, in: \bibinfo{booktitle}{Proceedings of the 21th ACM SIGKDD international conference on knowledge discovery and data mining}, pp. \bibinfo{pages}{1365--1374}.
%Type = Article
\bibitem[{Ye et~al.(2022)Ye, Chen, Han and Liao}]{ye2022robust}
\bibinfo{author}{Ye, S.}, \bibinfo{author}{Chen, D.}, \bibinfo{author}{Han, S.}, \bibinfo{author}{Liao, J.}, \bibinfo{year}{2022}.
\newblock \bibinfo{title}{Robust point cloud segmentation with noisy annotations}.
\newblock \bibinfo{journal}{IEEE Transactions on Pattern Analysis and Machine Intelligence} \bibinfo{volume}{45}, \bibinfo{pages}{7696--7710}.
%Type = Inproceedings
\bibitem[{Yi et~al.(2023)Yi, Guan, Huang and Lu}]{yi2023class}
\bibinfo{author}{Yi, R.}, \bibinfo{author}{Guan, D.}, \bibinfo{author}{Huang, Y.}, \bibinfo{author}{Lu, S.}, \bibinfo{year}{2023}.
\newblock \bibinfo{title}{Class-independent regularization for learning with noisy labels}, in: \bibinfo{booktitle}{Proceedings of the AAAI Conference on Artificial Intelligence}, pp. \bibinfo{pages}{3276--3284}.
%Type = Article
\bibitem[{Yin et~al.(2023)Yin, Shen, Wang, Luo, Luo and Tao}]{yin2023omg}
\bibinfo{author}{Yin, N.}, \bibinfo{author}{Shen, L.}, \bibinfo{author}{Wang, M.}, \bibinfo{author}{Luo, X.}, \bibinfo{author}{Luo, Z.}, \bibinfo{author}{Tao, D.}, \bibinfo{year}{2023}.
\newblock \bibinfo{title}{Omg: Towards effective graph classification against label noise}.
\newblock \bibinfo{journal}{IEEE Transactions on Knowledge and Data Engineering} \bibinfo{volume}{35}, \bibinfo{pages}{12873--12886}.
%Type = Inproceedings
\bibitem[{Ying et~al.(2018)Ying, You, Morris, Ren, Hamilton and Leskovec}]{DBLP:conf/nips/YingY0RHL18}
\bibinfo{author}{Ying, Z.}, \bibinfo{author}{You, J.}, \bibinfo{author}{Morris, C.}, \bibinfo{author}{Ren, X.}, \bibinfo{author}{Hamilton, W.L.}, \bibinfo{author}{Leskovec, J.}, \bibinfo{year}{2018}.
\newblock \bibinfo{title}{Hierarchical graph representation learning with differentiable pooling}, in: \bibinfo{booktitle}{Advances in Neural Information Processing Systems}, pp. \bibinfo{pages}{4805--4815}.
%Type = Inproceedings
\bibitem[{You et~al.(2020)You, Chen, Sui, Chen, Wang and Shen}]{you2020graph}
\bibinfo{author}{You, Y.}, \bibinfo{author}{Chen, T.}, \bibinfo{author}{Sui, Y.}, \bibinfo{author}{Chen, T.}, \bibinfo{author}{Wang, Z.}, \bibinfo{author}{Shen, Y.}, \bibinfo{year}{2020}.
\newblock \bibinfo{title}{Graph contrastive learning with augmentations}, in: \bibinfo{booktitle}{Proceedings of the 34th International Conference on Neural Information Processing Systems}, pp. \bibinfo{pages}{5812--5823}.
%Type = Inproceedings
\bibitem[{Yuan et~al.(2023a)Yuan, Luo, Qin, Mao, Ju and Zhang}]{yuan2023alex}
\bibinfo{author}{Yuan, J.}, \bibinfo{author}{Luo, X.}, \bibinfo{author}{Qin, Y.}, \bibinfo{author}{Mao, Z.}, \bibinfo{author}{Ju, W.}, \bibinfo{author}{Zhang, M.}, \bibinfo{year}{2023}a.
\newblock \bibinfo{title}{Alex: Towards effective graph transfer learning with noisy labels}, in: \bibinfo{booktitle}{Proceedings of the 31st ACM international conference on multimedia}, pp. \bibinfo{pages}{3647--3656}.
%Type = Inproceedings
\bibitem[{Yuan et~al.(2023b)Yuan, Luo, Qin, Zhao, Ju and Zhang}]{yuan2023learning}
\bibinfo{author}{Yuan, J.}, \bibinfo{author}{Luo, X.}, \bibinfo{author}{Qin, Y.}, \bibinfo{author}{Zhao, Y.}, \bibinfo{author}{Ju, W.}, \bibinfo{author}{Zhang, M.}, \bibinfo{year}{2023}b.
\newblock \bibinfo{title}{Learning on graphs under label noise}, in: \bibinfo{booktitle}{ICASSP 2023-2023 IEEE International Conference on Acoustics, Speech and Signal Processing}, pp. \bibinfo{pages}{1--5}.
%Type = Article
\bibitem[{Zhang et~al.(2024)Zhang, Cheng, Yuan and Zhang}]{zhang2024learning}
\bibinfo{author}{Zhang, G.}, \bibinfo{author}{Cheng, D.}, \bibinfo{author}{Yuan, G.}, \bibinfo{author}{Zhang, S.}, \bibinfo{year}{2024}.
\newblock \bibinfo{title}{Learning fair representations via rebalancing graph structure}.
\newblock \bibinfo{journal}{Information Processing \& Management} \bibinfo{volume}{61}, \bibinfo{pages}{103570}.
%Type = Inproceedings
\bibitem[{Zhang et~al.(2022)Zhang, Chen, Backes, Shen and Zhang}]{zhang2022inference}
\bibinfo{author}{Zhang, Z.}, \bibinfo{author}{Chen, M.}, \bibinfo{author}{Backes, M.}, \bibinfo{author}{Shen, Y.}, \bibinfo{author}{Zhang, Y.}, \bibinfo{year}{2022}.
\newblock \bibinfo{title}{Inference attacks against graph neural networks}, in: \bibinfo{booktitle}{Proceedings of the 31th USENIX Security Symposium}, pp. \bibinfo{pages}{1--18}.
%Type = Inproceedings
\bibitem[{Zhang et~al.(2023)Zhang, Chen, Fang, Li, Chen, Lin and Li}]{zhang2023rankmatch}
\bibinfo{author}{Zhang, Z.}, \bibinfo{author}{Chen, W.}, \bibinfo{author}{Fang, C.}, \bibinfo{author}{Li, Z.}, \bibinfo{author}{Chen, L.}, \bibinfo{author}{Lin, L.}, \bibinfo{author}{Li, G.}, \bibinfo{year}{2023}.
\newblock \bibinfo{title}{Rankmatch: Fostering confidence and consistency in learning with noisy labels}, in: \bibinfo{booktitle}{Proceedings of the IEEE/CVF International Conference on Computer Vision}, pp. \bibinfo{pages}{1644--1654}.
%Type = Article
\bibitem[{Zhou et~al.(2024)Zhou, Ye and Cao}]{zhou2024node}
\bibinfo{author}{Zhou, T.}, \bibinfo{author}{Ye, H.}, \bibinfo{author}{Cao, F.}, \bibinfo{year}{2024}.
\newblock \bibinfo{title}{Node-personalized multi-graph convolutional networks for recommendation}.
\newblock \bibinfo{journal}{Neural Networks} , \bibinfo{pages}{106169}.
%Type = Inproceedings
\bibitem[{Zhou et~al.(2021)Zhou, Liu, Wang, Zhai, Jiang and Ji}]{zhou2021learning}
\bibinfo{author}{Zhou, X.}, \bibinfo{author}{Liu, X.}, \bibinfo{author}{Wang, C.}, \bibinfo{author}{Zhai, D.}, \bibinfo{author}{Jiang, J.}, \bibinfo{author}{Ji, X.}, \bibinfo{year}{2021}.
\newblock \bibinfo{title}{Learning with noisy labels via sparse regularization}, in: \bibinfo{booktitle}{Proceedings of the IEEE/CVF international conference on computer vision}, pp. \bibinfo{pages}{72--81}.

\end{thebibliography}

%% else use the following coding to input the bibitems directly in the
%% TeX file.

% \begin{thebibliography}{00}

% %% \bibitem[Author(year)]{label}
% %% Text of bibliographic item

% \bibitem[ ()]{}

% \end{thebibliography}
\end{document}